# Intelligent Building Control Systems for Thermal Comfort and Energy-Efficiency: A Systematic Review of Artificial Intelligence-Assisted Techniques

Ghezlane Halhoul Merabet [a, i,*], Mohamed Essaaidi [a], Mohamed Ben Haddou [b], Basheer Qolomany [c], Junaid Qadir [d], Muhammad Anan [e], Ala Al-Fuqaha [f, g], Mohamed Riduan Abid [h], Driss Benhaddou [i]

[a] *Smart Systems Laboratory (SSL), ENSIAS, Mohammed V University in Rabat, 713, Morocco*

[b] *MENTIS Consulting SA, 13, rue de Congrès, 1000 Brussels, Belgium*

[c] *Department of Cyber Systems, College of Business and Technology, University of Nebraska at Kearney (UNK), Kearney, NE 68849, USA*

[d] *Information Technology University, Lahore 54000, Pakistan*

[e] *Software Engineering Department, Alfaisal University-Riyadh, Saudi Arabia*

[f] *Information and Computing Technology (ICT) Division, College of Science and Engineering (CSE), Hamad Bin Khalifa University, Doha – Qatar*

[g] *Department of Computer Science, Western Michigan University, Kalamazoo, MI 49008, USA*

[h] *School of Science and Engineering, Alakhawayn University in Ifrane, 1005, Morocco*

[i] *Department of Engineering Technology, University of Houston, TX 77204, USA*

**Abstract –** Building operations represent a significant percentage of the total primary energy consumed in most countries due to the proliferation of Heating, Ventilation and Air-Conditioning (HVAC) installations in response to the growing demand for improved thermal comfort. Reducing the associated energy consumption while maintaining comfortable conditions in buildings are conflicting objectives and represent a typical optimization problem that requires intelligent system design. Over the last decade, different methodologies based on the Artificial Intelligence (AI) techniques have been deployed to find the sweet spot between energy use in HVAC systems and suitable indoor comfort levels to the occupants. This paper performs a comprehensive and an in-depth systematic review of AI-based techniques used for building control systems by assessing the outputs of these techniques, and their implementations in the reviewed works, as well as investigating their abilities to improve the energy-efficiency, while maintaining thermal comfort conditions. This enables a holistic view of (1) the complexities of delivering thermal comfort to users inside buildings in an energy-efficient way, and (2) the associated bibliographic material to assist researchers and experts in the field in tackling such a challenge. Among the 20 AI tools developed for both energy consumption and comfort control, functions such as identification and recognition patterns, optimization, predictive control. Based on the findings of this work, the application of AI technology in building control is a promising area of research and still an ongoing, i.e., the performance of AI-based control is not yet completely satisfactory. This is mainly due in part to the fact that these algorithms usually need a large amount of high-quality real-world data, which is lacking in the building or, more precisely, the energy sector. Based on the current study, from 1993 to 2020, the application of AI techniques and personalized comfort models has enabled energy savings on average between 21.81 and 44.36 %, and comfort improvement on average between 21.67 and 85.77 %. Finally, this paper discusses the challenges faced in the use of AI for energy productivity and comfort improvement, and opens main future directions in relation with AI-based building control systems for human comfort and energy-efficiency management.

---

[*] Corresponding author.
   E-mail address: ghezlane.merabet@um5s.net.ma (Ghezlane Halhoul Merabet).

| | | | |
|---|---|---|---|
| **Nomenclature** | | ICT | Information and Communication Technologies |
| | | IEA | International Energy Agency |
| | | IEEMS | Indoor Environment Energy Management System |
| ABM | Agent-based Model | IEQ | Indoor Environmental Quality |
| ACMV | Air-Conditioning and Mechanical Ventilation | IHMS | Intelligent Heat Management System |
| AI | Artificial Intelligence | IoT | Internet of Things |
| ANFIS | Adaptive Neuro Fuzzy Inference System | KBS | Knowledge-Based System |
| ANN | Artificial Neural Networks | kNN | k-Nearest Neighbor |
| ARX | Autoregressive Exogenous | LBMPC | Learning-Based Model Predictive Control |
| ASHRAE | American Society of Heating, Refrigerating, and Air-Conditioning Engineers | LR | Logistic regression |
| BCM | Bayesian Comfort Model | LRLC | Linear Reinforcement Learning Controller |
| BEMS | Building Energy Management System | LSTM | Long Short-Term Memory |
| BN | Bayesian Network | MACES | Multi-Agent Comfort and Energy System |
| BRITE | Berkeley Retrofitted and Inexpensive HVAC Testbed | MAS | Multi-Agent Systems |
| CA | Context-Awareness | MBPC | Model-Based Predictive Control |
| CAC | Comfort Air-Conditioning | MISO | Multi-Input, Single-Output |
| CFD | Computational Fluid Dynamics | ML | Machine Learning |
| CHP | Combined Heat and Power | MLR | Multivariate Linear Regression |
| CI | Computational Intelligence | MOABC | Multi-Objective Artificial Bee Colony |
| CIBSE | Chartered Institution of Building Services Engineers | MOGA | multi-objective genetic algorithm |
| CL | Cooling Load | MOPSO | Multi-Objective Particle Swarm Optimization |
| CTR | Comfort Time Ratio | MRA | Multiple Regression Analysis |
| DAI | Distributed Artificial Intelligence | MSE | Mean Squared Error |
| DCC | Demand-driven Cooling Control | NARX | Nonlinear Autoregressive Exogenous |
| DID | Degree of Individual Dissatisfaction | NFQ | Neural Fitted Q-iteration |
| DL | Load Demand | NIST | National Institute of Standards and Technology |
| DNN | Deep Neural Networks | NSGA | Nondominated Sorting Genetic Algorithm |
| DRL | Reinforcement Learning | OMG | Occupant Mobile Gateway |
| DT | Decision Tree | OSHA | Occupational Safety and Health Administration |
| EACRA | Energy Aware Context Recognition Algorithm | PAR | Peak to Average Ratio |
| EDA | Epistemic-Deontic-Axiologic | PID | Proportional-Integral-Derivative |
| eJAL | Extended Joint Action Learning | PMV | Predicted Mean Vote |
| FACT | Fuzzy Adaptive Comfort Temperature | PPD | Predicted Percentage of Dissatisfied |
| FCM | Fuzzy Cognitive Map | PPV | Predicted Personal Vote |
| FDM | Fused Deposition Modeling | PRISMA | Preferred Reporting Items for Systematic Reviews and Meta-Analysis |
| FIS | Fuzzy Inference System | PSO | Particle Swarm Optimization |
| FLC | Fuzzy Logic Control | RBF | Rule Base Function |
| FRB | Fuzzy Rule Base | RBF | Radial Basis Function |
| GA | Genetic Algorithm | RF | Random Forest |
| gARTMAP | Gaussian Adaptive Resonance Theory Map | RL | Reinforcement Learning |
| HABIT | Human and Building Interaction Toolkit | RMSE | Root-Mean-Square Deviation |
| HIYW | Have-It-Your-Way | RNN | Recurrent Neural Network |
| HL | Heating Load | SBSA | State-Based Sensitivity Analysis |
| HMM | Hidden Markov Model | SQP | Sequential Quadratic Programming |
| HVAC | Heating, Ventilation, and Air-Conditioning | SVM | Support Vector Machine |
| HVAC&R | Heating, Ventilation and Air-Conditioning & Refrigeration | TPI | Thermal Perception Index |
| iBEMS | Intelligent Building Energy Management Systems | VAV | Variable Air Volume |
| ICCS | Intelligent Comfort Control System | WSN | Wireless Sensor Networks |

# 1. Introduction

At a time when energy use and depletion of natural resources have become a perennial debate around the world, questions about the scale of cost growth and the peak demand for electricity have risen exponentially. Proportionally, climate change, increased frequency and the severity of warmer periods throughout the year have made a considerable part of the global population dependent on artificial conditioning, which further drives peak demand during the summer [1–3]. Figure 1 shows the penetration of HVAC systems in households in some countries around the world. HVAC penetration will impact peak power demand which is extreme in countries such as Japan, the United States, and Korea. On the other hand, countries with low HVAC penetration are expected to have their peak electricity load increase by around 45% by 2050, according to the International Energy Agency (IEA) [4]. This significant consumption raises concerns about the management and energy-efficiency of these devices, given the thermal comfort of the buildings' occupants. Research in this area focuses on balancing the two main objectives in question: thermal comfort and energy conservation.

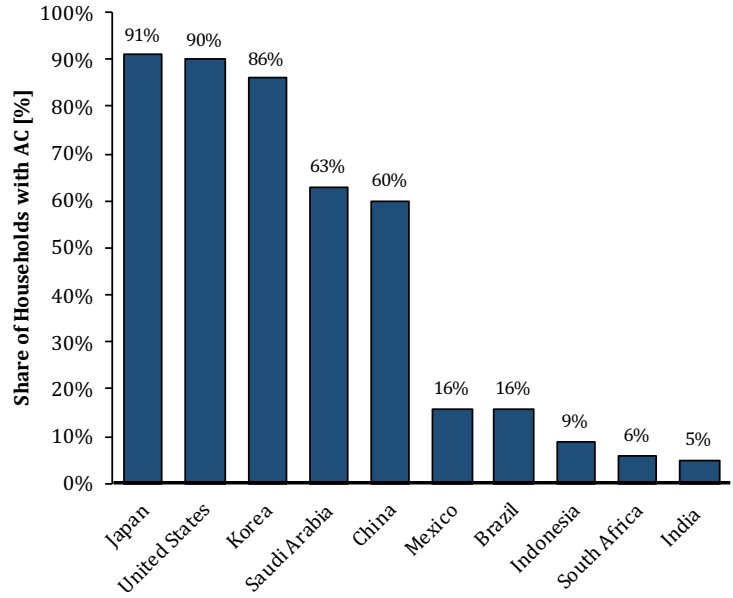

**Figure 1.** Percentage of households equipped with air-conditioners in selected countries, 2018 (Source: adopted from IEA [4]).

Various factors such as thermal comfort, visual comfort, acoustic comfort, and Indoor Air Quality (IAQ) have a significant combined effect on the life quality and comfort conditions perceived by the occupants who may spend between 60% and 90% of their time indoors [5]. Understanding the concept of thermal comfort and finding ways to predict whether a given situation represents comfort or discomfort has been the subject of the study for more than 50 years. Studies in climatic chambers have made it possible to standardize these factors using an analytical approach. In this approach, thermal comfort is assessed by the Predicted Mean Vote (PMV) index, which predicts the average thermal sensation vote on a standardized scale for a large group of individuals. Subsequently, the American Society of Heating, Refrigerating, and Air-Conditioning Engineers (ASHRAE) established a thermal comfort code (cf. Table 1). Then, the PMV index was adopted by ISO 7730 standard, which recommends to maintain the PMV at level 0, with a tolerance of 0.5 as the best thermal comfort level.

**Table 1.** The Seven-point thermal sensation scale (Source: Adopted from ISO 7730-2005).

| SENSATION | Cold | Cool | Slightly Cool | Neutral | Slightly Warm | Warm | Hot |
|---|---|---|---|---|---|---|---|
| VOTE | -3 | -2 | -1 | 0 | +1 | +2 | +3 |

The first so-called *analytical* approach has been identified. It focuses on the physical aspect by defining equations that characterize the degree of comfort that satisfies the occupant, such as the heat balance equation (cf. Equation (1)), which expresses the heat exchange between human body and environment. This approach brings a reductive vision of comfort based on physical and physiological mechanisms. Subsequently, in situ experimental studies have shown a discrepancy between reality and the predictions of current standards that restrict comfort conditions within narrow, and defined intervals independently of the subjectivity of human behavior.

$$M - W = E + R_{HE} + K + C + S \qquad (1)$$

Whereas, $M$ is the metabolic rate ($W/m^2$); $W$ is the mechanical work done by the body ($W/m^2$); $E$ is the evaporative heat gain or loss ($W/m^2$); $R_{HE}$ is the radiant heat exchange ($W/m^2$); $K$ is the conductive heat transfer ($W/ºC.m^2$); $C$ is the convective heat transfer ($W/m^2$); $S$ is the net heat storage ($W/m^2$).

These in-situ studies have made it possible to develop a new approach, the *adaptive* approach, which is based on the fact that the human-being is never passive towards a given thermal environment. He receives thermal information from different sensors located on the body, and reacts accordingly to find his comfort temperature. According to Humphreys M. and Nicol F. [6], the adaptive comfort is defined as "*If a change occurs as to produce discomfort, people react in ways which tend to restore their comfort*". Adaptive methods do not predict a comfort response, but rather the almost constant conditions under which people are likely to be in a comfort situation.

Additionally, visual comfort is a subjective outcome related to the quantity, distribution and quality of light. Achieving a comfortable visual sensation in a building promotes the well-being of its occupants. On the other hand, weak or too strong lighting, poorly distributed in space or whose light spectrum is poorly adjusted to the eye's sensitivity causes fatigue or a feeling of discomfort and reduced visual performance. Whereas, indoor air quality is an important factor in occupant comfort. It can be measured by the carbon dioxide ($CO_2$) concentration in a building. The concentration of $CO_2$ comes from the presence of residents in the building and various other sources of pollution (NOx, Volatile Organic Compounds (VOCs), respirable particles, among others). Ventilation is an effective way to control indoor air quality in buildings. Although in many studies of comfort and energy performance of buildings, different comforts are coupled, in this work we are focusing on thermal comfort.

Moreover, in order to adjust the indoor climate conditions, many technological solutions have been proposed. However, achieving a comfortable and energy efficient environment, the occupant is supposed to become an *expert* of these technologies that can challenge his daily habits. Given the complexity of these technologies, the user could choose the solution of the smart buildings equipped with sensors to adjust everything (temperature, ventilation, opening/closing windows) to promote energy savings and comfort. In this context, research works have been oriented at more advanced building control systems based on Artificial Intelligence (AI), taking into account several aspects such as the user preference over time [7].

Initial efforts to apply AI for building control began in the 1990's. Intelligent controllers have been optimized using evolutionary algorithms designed to control smart building subsystems. Synergy between neural networks, fuzzy logic and evolutionary algorithms, or more broadly Computational Intelligence (CI) techniques, has been applied to buildings. To overcome non-linear functions of thermal comfort indices, time delay, and system uncertainty, certain advanced control algorithms have also integrated adaptive fuzzy control for optimal comfort control. In this context, a direct neural network controller, using a *back-propagation* algorithm, was developed and successfully deployed in Japanese air-conditioning installations and electrical fans for commercial applications. For example, a system incorporating two neural networks has been integrated into an air-conditioner in order to ensure that the equipment is adapted to customer preferences [8].

The literature review has revealed that various published works have reviewed optimization and advanced comfort management controls. However, none of them have brought all the material related to the AI-based methods for both thermal comfort and energy consumption control in built-environments, while including individual interactions into the comfort-energy control loop. Following, we describe and compare the most relevant previous survey papers with our work in Table 2.

An interesting appraisal of advanced energy control and comfort management schemes in sustainable buildings by focusing on agent-based control is provided by Dounis A. I. and Caraiscos C. [9], while Shaikh P. H. et al. [10] analyzed intelligent control and optimization methods for smart sustainable buildings. Similarly, Evins R. [11] reviewed research works on the application of computational optimization to address issues in sustainable building design. Behrooz F. et al. [12] investigated control techniques for heating, ventilation and air conditioning and refrigeration (HVAC & R) control by insisting on a Fuzzy Cognitive Maps (FCM) as an intelligent method, whereas Cheng C.-C. and Lee D. [13] covered the techniques of artificial intelligence to optimize HVAC control. Royapoor M. et al. [14] discussed the industrial perspectives of building control techniques, while Qolomany B. et al. [15] discussed the role of machine learning techniques and big-data analytics in smart building services. Ngarambe J. et al. [16] reviewed the AI-based techniques for thermal comfort and energy usage in buildings with an emphasis on thermal comfort predictive models and their applicability for energy control purposes.

**Table 2.** Comparison of relevant previous survey papers.

| REF. OF THE WORK | YEAR | PURPOSE FROM THE WORK | LIMITATIONS OF THE WORK |
|---|---|---|---|
| **Dounis A. I. and Caraiscos C. [9]** | 2009 | Review and categorize the advanced control schemes of energy and comfort control in indoor environments with a focus on agent-based control systems. | Does not review the intelligent control systems in the framework of artificial intelligence, it does not consider the human factor in energy and thermal comfort control issues. |
| **Evins R. [11]** | 2013 | Review the applicability of computational (multi-objective) optimization in sustainable building design problems. | The work covers the optimization functions of the Computational Intelligence (CI) applied to sustainable building rather than focusing on the application of AI in building control. |
| **Shaikh P. H. et al. [10]** | 2014 | Review the intelligent control systems for energy and comfort conditions control with occupant interaction in Smart Energy Buildings (SEBs). | Does not focus specifically on the applicability of the AI in energy saving and thermal comfort control. |
| **Behrooz F. et al. [12]** | 2018 | Investigate different methods for HVAC systems control with a focus on the Fuzzy Cognitive Map (FCM) as an intelligent method to overcome the system constraints. | Does not classify the research works according to AI techniques, also does not include the human preferences/sensitivity about thermal environment. |
| **Royapoor M. et al. [14]** | 2018 | Investigate the building climate and plant control strategies with surveying their industry perspectives in real world adoption. | Does not focus specifically on AI methods and their applications in building control. |
| **Cheng C.-C. And Lee D. [13]** | 2019 | Cover the role of AI technology to improve the efficiency of HVAC systems with discussing the AI-assisted techniques accuracy in data prediction. | The work considers only energy saving of HVAC systems without including thermal comfort (and hence the human factor) into the loop. |
| **Qolomany B. et al. [15]** | 2019 | Review the role of ML and big-data analytics in smart building structure and utilities. | The work focuses mainly on ML and big-data analytics in intelligent building control, rather than classifying research according to AI/ML techniques for energy-comfort controls. |
| **Ngarambe J. et al. [16]** | 2020 | Review the usage of the advanced AI-based methods in energy control while sustaining thermal comfort in indoor spaces with a focus on the comfort predictive model based on ML and their deployment for energy saving. | The work focuses only AI/ML use in residential buildings, and it does not focus on the individual comfort modeling |

To the best of our knowledge, this paper is the first to cover the role of the personal interaction of occupants using AI-assisted tools to achieve energy efficient thermal comfort optimization in buildings. Our main contributions, through this work, are:

- Investigating and exploring the potential of AI/ML-assisted tools in improving the performance of building control systems, in terms of human comfort and energy efficiency.
- Identifying various research challenges and directions on how AI may be used to improve energy productivity while meeting comfort requirements.
- Identifying the different components of an AI-assisted building control system, their functionalities, and how they perform to ensure better comfort conditions while improving energy efficiency.

The remainder of this paper first describes the search results analysis framework (Section 2), and then discusses AI-based techniques used to control energy consumption while maintaining comfortable conditions in buildings along with a discussion on the methods adopted for thermal comfort assessment. Section 3 presents a theoretical analysis of the popular AI techniques in which case studies on the application of these methods to different parts of the building control system as well as inputs and outputs of the models are examined. Section 4 discusses the main outcomes and findings from the review. Section 5 provides a discussion of key insights, synthesizes further research directions and identifies related challenges. Finally, the paper concludes with a summary description of our work in Section 6.

## 2. Systematic Review Methodology

Through this review, we systematically investigated how Artificial Intelligence (AI) can be used to improve thermal comfort and energy efficiency requirements in buildings. The main terms including "thermal comfort", "artificial intelligence", "big data", "occupant", and "heating, ventilation and air-conditioning systems" were used to perform articles' search on databases and search engines: ACM Digital Library [17], Scopus [18], Google Scholar [19], IEEE Xplore On Line (IEOL) [20], Web of Science [21], and Science Direct-On-Line (SDOL) [22], apart from other sources (i.e., manual search). These databases were selected for being repositories of pertinent and relevant scientific publications over the period between 1992 and 2020 in the field being studied. The review process followed the criteria established by the Preferred Reporting Items for Systematic Reviews and Meta-Analysis (PRISMA) guidelines, designed to guide systematic review and meta-analysis studies [23,24].

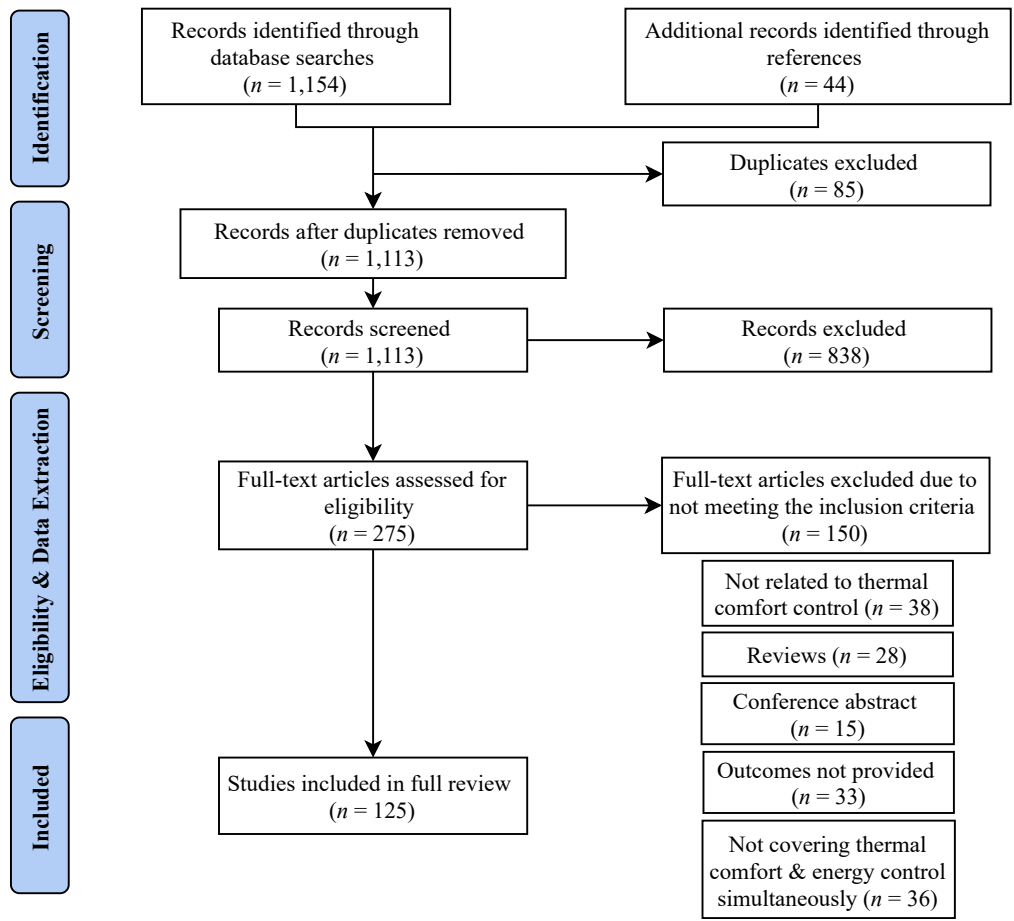

**Figure 2.** Flowchart of the articles' selection process.

Initially, 1,198 articles were gathered from the mentioned resources. Following the application of filters according to the search protocol illustrated in Figure 2, a total of 125 studies satisfied the inclusion criteria, which include (1) studies carried out in indoor conditions; (2) research work incorporating groundbreaking AI-based techniques and their use for HVAC and thermal comfort control; and (3) research work reporting the system performance following the application of AI control tools. The reviewed studies were arranged chronologically to present a literature analysis of innovative AI techniques, and their usability, over time, to improve the performance of energy savings and thermal comfort in buildings. This allows to establish an added value to the intelligence for building control systems. Moreover, the main consideration was given to the diversity amongst the occupants' preferences in buildings, methods to measure or infer thermal comfort from the occupants, indoor control systems (space control devices, load control, building component, occupant interactions). In addition, information involving study cases, AI techniques deployed, the model and/or methods adopted to infer the thermal comfort, input(s) and controlled parameter(s) as well as the enhanced performances of thermal comfort and energy savings, are summarized in the next section (cf. Tables 3 to 9).

## 3. AI Development for Thermal Comfort and Energy-Efficiency in Buildings

AI is a field of expertise that offers decision support and control models based on real facts and empirical and theoretical knowledge. In this sense, one of the main objectives of AI is related to the development of systems capable of solving problems that only the capacity of human beings reasoning allows due to their ability to learn and make decisions correctly (i.e., intelligence). AI has the challenge of developing problem-solving systems that can be converted into mathematical models and programs for use in computers or controllers. In this section, the principles of the different AI techniques used to design building controllers in the reviewed works will be defined first, we will highlight in particular the most used and well-known tools. In the second part of this section, we will introduce the optimization functions, as a fundamental key in the building control system component.

### *3.1. Developed AI Control Techniques*

Numerous AI-based solutions were developed for controlling energy and thermal comfort inside buildings. Accordingly, in this paper, we performed a thorough analysis of all existing works on the use of AI in building control, which are recorded separately in Tables 3-9. Based on this analysis, approximately 20 AI techniques were employed to control both thermal comfort and energy consumption. Among these techniques, the most well-know are **Artificial Neural Networks (ANNs)**, including **Recurrent Neural Networks (RNNs)**, **Deep Neural Networks (DNNs)** and **Feedforward-ANNs**, are among the most well-known tools (cf. Table 3). First introduced by McCulloch and Pitts in 1943, ANNs are now widely used methods for obtaining efficient results in many domains, in supervised or unsupervised classifications. Although it dates back to the 1950s (then called a perceptron and composed of a single neuron) [25,26], ANN was later developed with the introduction of new types of ANN [27], and new learning methods [28,29]. Deep learning has continued to be refined thereafter [30,31], yet has above all revealed its potential through the provision of powerful computational tools (such as graphics processors) to leverage the potential of ANN.

An artificial neuron (or formal neuron) is inspired by a biological neuron to which it gives mathematical inspiration as shown in Figure 3.

In a formal neuron, we observe:
- **Inputs** ($X = x_1, ..., x_n$) to which are associated weights ($W = \omega_1, ..., \omega_n$) relating to the importance of the information conveyed;
- A **bias** ($b$) constituting the weight of a constant input allowing to add flexibility to the network by acting on the position of the decision boundary;
- An **activation function** () applied to inputs and bias, such as:
  - Sigmoid: $\varphi(Z) = 1/1 + e^{-Z}$
  - Hyperbolic tangent ($tanh$): $\varphi(Z) = (1 - e^{-Z})/(1 + e^{-Z})$

o  ReLu (Rectified linear unit): $\varphi(Z) = (0, Z)$
   o  Identity: $\varphi(Z) = Z$
● An **output** ($y$) that can be used as the input of other neurons, such as:

$$y = \varphi(W.X + b) \tag{2}$$

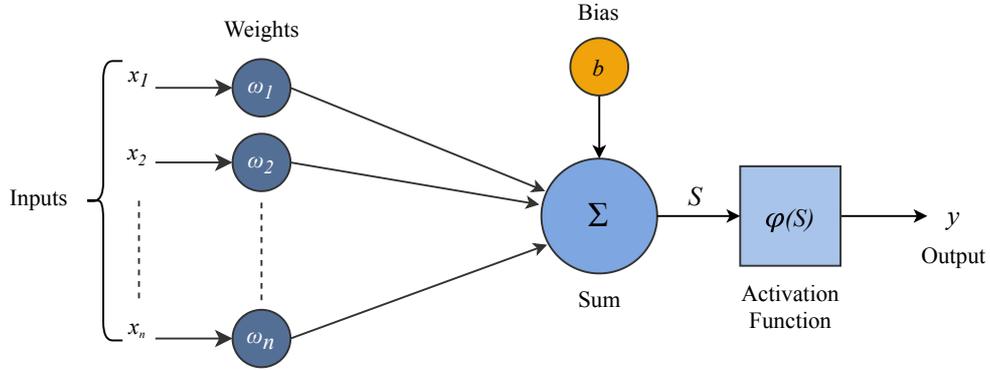

**Figure 3.** Mathematical model of a biological neuron.

An ANN is then the association of several neurons grouped in layers linked by weighted connections.

The ANN architecture determines the way neurons are ordered and connected within the same network. In general, the ANN consists of several successive layers of neurons: the inputs, the hidden layers (which are not accessible outside the network) up to the output layer(s). The depth of the ANN is estimated by the number of hidden layers.

In the field of thermal building, ANNs are used to solve various problems. In fact, direct neural network controllers were applied for thermal comfort monitoring [32] and HVAC control systems [33–35]. Such controllers are simple and do not need a building recognition model, unlike indirect neural network controllers. Figure 4 shows the configuration of a neural network controller, which is dual-layer, Multi-Input, Single-Output (MISO) [32]. There are two inputs and one output for this controller: $e$ is an error between the PMV setting and the feedback value, $\dot{e}$ is a differential error, and u is a building control signal.

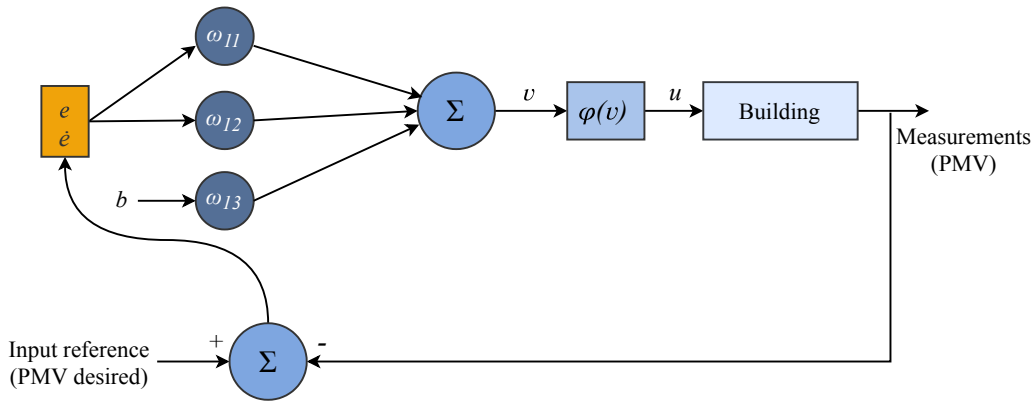

**Figure 4.** A direct neural network controller (Source: Adopted from [36]).

The equations describing this controller are as follow [36]:

$$v = w_{11}e + w_{12}\dot{e} + w_{13}b \tag{3}$$

$$u = \varphi(v) = 1/1 + \exp(-v^2) \tag{4}$$

$$\Delta w_{ij} = -\eta \frac{\partial E}{\partial w_{ij}} = -\eta \frac{\partial E}{\partial PMV} \frac{\partial PMV}{\partial u} \frac{\partial u}{\partial w_{ij}} = \pm \eta^* \frac{\partial E}{\partial PMV} \frac{\partial u}{\partial w_{ij}} \tag{5}$$

Whereas $v$ in the input of the output layer of neural network; $\omega_{11}$ and $\omega_{12}$ are the synaptic weights; $\omega_{13}$ is the synaptic weight of the fixed input $b = 1$; $\varphi(v)$ is the activation function (unipolar sigmoid function); $u$ is the output of the output layer; and $\eta^*$ is the learning rate parameter.

Learning an ANN is, in essence, the adjustment of these weight coefficients to optimize the cost functions. The weight of interconnections between neurons is based on the gradient descent algorithm. Initially, this algorithm sets random values to the weights of the network, obtains the two input signals of the controller, and calculates the output. Afterwards, the algorithm adjusts both the weights and the output signal.

Neural networks can also be used to generate an adaptive regulation model, as is the case in Ref. [37]. Based on the PMV index, this work proposes a control system that shows the ability of neural networks based on a backpropagation algorithm to modify the environmental conditions of an enclosed indoor space, considering not only the indoor climate variables, but also the index of activity level and clothing of the occupants of the space. As a limitation, it has been noticed that the network takes a long time to learn, and this makes the system, even before the variations in user activity level, and the various clothing indices during the day, also has a slow response to keep the PMV index at zero. Moreover, as HVAC systems are dynamic and non-linear, it is also common to use controller based on dynamic neural networks, such as in Ref. [38], where a dynamic neural network based on the idea of Nonlinear Autoregressive Exogenous (NARX) is used to model and control an HVAC system.

In addition to neural network models, **Fuzzy Logic Control** (**FLC**) is an appropriate tool for imitating the behavior of building users and developing linguistic descriptions of the thermal comfort sensation or preferences that approach the PMV and adaptive models, which is not easily interpreted by the control system (cf. Table 4). Unlike conventional control methods, FLC is more widely used in poorly specified, complicated procedures that can be managed by a professional human agent without a deep understanding of their underlying mechanisms. The basic idea behind FLCs is to integrate the "expert knowledge" of a human agent into the regulation of a mechanism whose input-output association is defined as a collection of fuzzy control rules (e.g., IF-THEN), which involve linguistic variables rather than a complex dynamic model. The use of linguistic variables, fuzzy rules and rough sets reasoning offers a way of integrating human expert experience into the design of the controller. The typical FLC architecture is shown in Figure 5, comprising four main elements: a **Fuzzifier**, a **Fuzzy Rule Base** (**FRB**), an **Inference Engine**, and a **Defuzzifier**.

a) **Fuzzification**, is the operation that consists of assigning a degree of membership to each fuzzy subset for each physical input. In other words, it is the operation that allows the transition from numerical (physical) to symbolic (fuzzy) variables.

b) **The knowledge base**, includes knowledge of the field of application and the objectives of the planned control. It includes:
- The basis of fuzzy rules for storing empirical knowledge of how the process works by experts in the field.
- The rule base is a set of linguistic expressions structured around expert knowledge and represented in the form of rules, such as: **IF <condition> THEN < consequence>**.

c) **Defuzzification:** this is the inverse fuzzification operation, which consists of converting the fuzzy number $B$ into the numerical quantity $y_0$.

d) **Fuzzy inference rules:** the inference engine is the core of the FLC, and has the ability to simulate human decision-making by performing rough reasoning in order to achieve the desired control strategy.

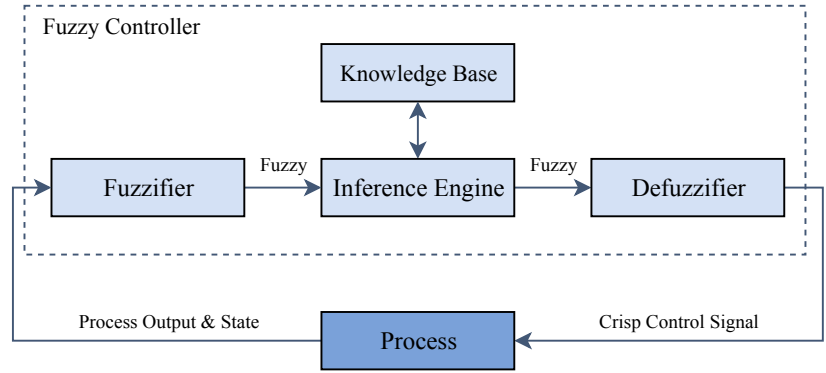

**Figure 5.** Schematic of fuzzy logic controller.

In the context of building control systems, the application of FL control methods for HVAC systems is efficient as this technique is well suited for non-linear systems [39]. These methods are able to uniformly approximate a nonlinear function to any degree of accuracy and also provide fast operation. In Ref. [40], the use of Fuzzy-PID, Fuzzy-PD and adaptive Fuzzy-PD methods to control thermal comfort and indoor air quality is described. One of the main objectives of this work was to reduce energy consumption. The lowest values were obtained with the adaptive Fuzzy-PD controller. Moreover, Bernard T. and Kuntze H-B. [41] proposed a fuzzy logic supervisory system, which allows the monitoring of the thermal environment inside a building where the customer could follow a compromise solution (through a weighting factor) between efficiency and comfort. In the same direction Hamdi M. and Lachiver G. [42] developed a concept of comfort conditions based on human sensitivity, without maintaining a constant internal temperature, but rather a constant indoor comfort. The results showed that it was possible to combine the comfort of the occupants while saving resources. Whereas in Ref. [43], a Fuzzy Adaptive Comfort Temperature (FACT) was introduced with a grey predictor enabling to predict comfort temperature as a linear function of outdoor information. The fuzzy rules were applied to determine the power needed in unpredictable situations.

**Table 3.** Summary of the works focusing on intelligent management of thermal comfort and energy in buildings using Artificial Neural Networks (ANN).

| YEAR | STUDY CASE | UNDERLYING AI/ML TOOLS | AI APPLICATION SCENARIO | THERMAL COMFORT METHOD | OPTIMIZATION OBJECTIVE | OUTCOMES & KEY RESULTS | REF. |
|---|---|---|---|---|---|---|---|
| 2005 | NN-based control development for individual thermal comfort optimization, and energy saving by combining a thermal space model for VAV&HVAC application. | Direct neural network | Optimized setting | PMV (Fanger's model) | Comfort parameters (PMV, HVAC, Temperature, Humidity), Energy/Load | The controller showed high comfort level (by maintaining the comfort zone between -0.5 and +0.5) while conserving energy. But some limitations remain in practice. | [32] |
| 2008 | Intelligent comfort control system (ICCS) design by combining the human learning and minimum energy consumption strategies for HVAC system application. | Deep neural networks | Optimized setting | PMV (Fanger's model) | Comfort parameters (PMV, Temperature, Humidity, VAV), Energy/Load | More energy saving and higher comfort level (by applying VAV control), compared to conventional temperature controller by maintaining the PMV within the comfort zone | [36] |
| 2009 | Developing an inferential sensor based on the adaptive neuro-fuzzy modeling to estimate the average temperature in space heating systems. | Adaptive neuro-fuzzy model | Adaptive neuro fuzzy inference system | Average air temperature estimation (based on To, QSQL, and Fire) | Comfort parameters (Temperature, Hot/Cold water), Energy/Load | The average air temperature estimated by ANFIS control model is very close to experimental results, with a highest possible RMSE = 0.5782°C. | [44] |
| 2009 | Predicting fan speed based on ANFIS for energy saving purpose in HVAC system | Adaptive neuro-fuzzy model | Predictive control | Desired temperature by controlling the damper | Comfort parameters (Temperature), Energy/Load | Simulation results showed that the ANFIS model is more effective and can be used as an alternative for HVAC control systems. | [45] |
| 2010 | Multi-objective optimization methodology used to optimize thermal comfort and energy consumption in a residential building | ANN combined with NSGA-II | Optimized setting | PMV (Fanger's model) | Comfort parameters (PMV, Heating/Cooling, Humidity, Temperature), Energy/Load | Optimization results showed significant improvement in thermal comfort (average PMV<4%), more saving in total energy (relative error<1%) and reduction in simulation time compared to conventional optimization methods. | [46] |
| 2011 | AI-based thermal control of a typical residential building in USA | Adaptive neuro-fuzzy model | Adaptive neuro fuzzy inference system (ANFIS) | Defined comfort ranges | Comfort parameters (Temperature, Hot/cold water), Energy/load | ANFIS could save 0.3% more energy than ANN. Both methods satisfied thermal comfort (~98% in winter/100% in summer), and reduction in Std. dev. of air temperature from setpoint temperature (under 0.3°C). | [47] |
| 2014 | Dynamic and automatic fuzzy controller for indoor for indoor thermal comfort requirements | Neural network-based ARX | Predictive control | Defined Temperature ranges (based on personal thermal preferences) | Comfort parameters (Temperature, Humidity), Energy/Load | The proposed control system allowed for efficient use of energy and brought the room temperature to the maximum value of personal comfort. | [48] |
| 2014 | Radiator-based heating system optimization to maintain indoor thermal comfort and minimize energy consumption for residential building | Random neural network (RNN) | Predictive control & optimized setting | PMV-based setpoint[2] $(PMV = a \cdot t + b \cdot p_v - c)$ | Comfort parameters (PMV, Temperature, Hot/cold water), Energy/load | The proposed model accuracy is of MSE=38.87% for PSO less than GA; MSE=21.19% for PSO less than SQP. RNN with GA allowed to maintain comfortable comfort conditions with the minimum energy consumption (400.6 MWH), compared to the MPC model. | [49] |
| 2014 | Control logic for thermally comfortable and energy-efficient environments in buildings with double skin envelopes | Rule-base & ANN-based control | Predictive & adaptive control | Comfort range (built from the cavity and indoor temperature conditions) | Comfort parameters (Temperature, heating/cooling), Energy/load | ANN-based logic showed significant results in reducing over/undershoots out of the comfort range. Simplest rule-base control logic use allowed to save cooling energy. | [50] |
| 2015 | Developing and testing an NN-based smart controller for maintaining a comfortable environment, and saving energy using a single zone test chamber | Recurrent neural networks (RNN) | Predictive control & optimized setting | User recommendations; PMV-based setpoints (Fanger's model) | Comfort parameters (PMV, Temperature, $CO_2$ Concentration/Air quality/ heating/cooling), Energy load | The proposed controller has learned the human preferences with an accuracy of 94.87% for heating, 98.39% for cooling and 99.27% for ventilation. The occupancy estimation using RNN is about 83.08%. | [51] |
| 2015 | Predictive-based controller development for multizone HVAC systems management in non-residential buildings. | Low-order ANN-based model | Predictive control & optimized setting | PMV (Fanger's model) | Comfort parameters (PMV, Temperature, Heating/ cooling), Time efficiency, Energy/load | The proposed strategy could optimize operation time of HVAC subsystems, reducing energy consumption and improving thermal comfort for cooling/heating modes. | [52] |
| 2015 | AI-theory-based optimal control for improving the indoor temperature conditions and heating energy efficiency of the building with double-skin. | ANN & ANN coupled with FLC | AI-based optimal control | Defined comfort temperature range | Comfort parameters (Temperature, Heating/ cooling), Energy/load | FLC, ANFIS-1 inputs and ANFIS-2 input models significantly increased the comfortable condition period by 2.92%, 2.61% and 2.73% respectively (compared to the rule-based algorithm). | [53] |
| 2015 | Automatic air-conditioning control development for indoor thermal comfort based on PMV and energy saving. | Adaptive neuro-fuzzy based model | Predictive control based on Inverse-PMV mode | Inverse-PMV model (based on desired PMV and measured variables) | Comfort parameters (PMV, Humidity, Temperature), Energy/load | The proposed control method performed better than conventional method by effectively maintaining the PMV within a range ±0.5 and up to 30% of energy saving. | [54] |
| 2016 | ANN-based algorithms development for optimal application of the setback moment during heating season. | ANN-based model | Predictive control & optimized setting | Defined setpoint temperature for occupied periods | Comfort parameters (Temperature), Energy/load | The optimized ANN model showed a promising prediction accuracy ($R^2$ up to 99.99%). ANN-based algorithms are much better in thermal comfort improvement (97.73% by Algorithm (1)); energy saving (14.04% by Algorithm (2)), compared to the conventional algorithm. | [55] |
| 2016 | ANN-based control algorithm development for improving thermal comfort and building energy efficiency of accommodation buildings during the cooling season. | ANN-based algorithms | Predictive & adaptive controls | Fixed setpoint/setback temperatures for occupied/ unoccupied periods | Comfort parameters (Temperature), Energy/load | ANN models gave accurate prediction results with acceptable error for comfort and energy improvement: 1st model: Average difference = 17.07%/MBE = 17.66%, 2nd model: Average difference = 20.87%/MBE = 21.90%. | [56] |

---

[2] Defined by the Institute for Environmental Research at KSU under the ASHRAE contract.

Table 3. (Continued).

| YEAR | STUDY CASE | UNDERLYING AI/ML TOOLS | AI APPLICATION SCENARIO | THERMAL COMFORT METHOD | OPTIMIZATION OBJECTIVE | OUTCOMES & KEY RESULTS | REF. |
|---|---|---|---|---|---|---|---|
| 2016 | A personalized energy management system (PEMS) development for HVAC systems in residential buildings. | Adaptive neuro-fuzzy based model | Predictive control | Personalized comfort bands | Comfort parameters (Temperature), Energy/load, Cost | About 9.7% to 25% reduction in energy consumption and the cost, from 8.2% to 18.2%. | [57] |
| 2017 | Proposing an AI-based heating and cooling energy supply model, responding to abnormal/ abrupt indoor situations, to enhance thermal comfort and energy consumption reduction. | Decision making based ANN model | Optimized setting | PMV-PPD (Fanger's model) | Comfort parameters (PMV/PPD, Temperature, Humidity, Heating/ cooling), Energy/load | Thermal comfort improvement: 2.5% for office building, and ~10.2% for residential building. annual energy consumption reduction: 17.4% for office building and 25.7% for residential buildings. | [58] |
| 2017 | AI-based controller development for improving thermal comfort and reducing peak energy demands in a district heating system. | ANN-based optimized (Opt. ANN) | Optimized setting & predictive control | PMV-PPD (Fanger's model) | Comfort parameters (PMV/PPD, Humidity, Temperature, Cooling/ heating), Energy/load | The proposed model's effectiveness > 27% for thermal comfort, reduction in annual energy loss over 30% for cooling and 40% for heating (compared to a conventional thermostat ON/OFF controller). | [59] |
| 2017 | A low-cost, high-quality decision-making mechanism (DMM) targeting smart thermostats in a smart building environment. | ANN and fuzzy inference system (FIS) | Neural-Fuzzy control | PPD (Fanger's model) | Comfort parameters (PPD, Temperature, Humidity), Energy/load | The proposed framework allowed to reach a higher thermal comfort while reducing energy consumption by an average between 18% and 40%. The use of FL by considering the dynamic behavior of the world allowed to improve the total cost by 7%–19% on average. | [60] |
| 2017 | Designing and implementing a smart controller by integrating the internet of things (IoT) with cloud computing for HVAC within an environment chamber. | Random neural network | Occupancy estimation & optimized setting | PMV (Fanger's model) | Comfort parameters (PMV, Temperature, HVAC, $CO_2$ concentration/Air quality), Energy/load | Results showed that the hybrid RNN-based occupancy estimation algorithm was accurate by 88%. ~27.12% reduction in energy consumption with the smart controller, compared to the simple rule-based controllers. | [61] |
| 2017 | RNN-based smart controller development for HVAC by integrating IoT with cloud computing and web services. | RNN trained with PSO-SQP | Occupancy estimation & optimized setting | PMV (Fanger's model) | Comfort parameters (PMV, Temperature, Humidity, $CO_2$ concentration/Air quality, HVAC), Energy/load | Energy consumption was 4.4% less than Case-1 and 19.23% less than Case-2. The RNN HVAC controller could maintain the user defined set-points and accurate temperature for PMV set-points. | [62] |
| 2017 | Implementing a predictive control strategy in a commercial BEMS for boilers in buildings. | ANN-based model | Predictive control | Predefined temperature (according to daytime) | Comfort parameters (Temperature, Hot/cold water), Energy/load | The predictive strategy allowed to reduce ~20% of energy required to heat the building without compromising the user's comfort, compared to scheduled ON/OFF control. | [63] |
| 2017 | A smart heating set-point scheduler development for an office building control. | ANN coupled with MOGA | Optimized setting & predictive control | PPD (Fanger's model) | Comfort parameters (PPD, Temperature, Humidity), Energy /load | 4.93% energy savings whilst improving thermal comfort by reducing the PPD by an average of 0.76%. | [64] |
| 2017 | A hybrid rule-based energy saving approach development using ANN and GA in buildings. | ANN-based model | Optimized setting | PMV (Fanger's model) | Comfort parameters (PMV, Temperature, Heating/ cooling), Energy/load | Validation results showed an average 25% energy savings while satisfying occupants' (elderly people) comfort conditions (-1≤PMV≤+1). | [65] |
| 2017 | Deploying ML techniques to balance energy consumption and thermal comfort in ACMV systems through computational intelligence techniques in optimizations. | ANN with Extreme learning machine | Optimized setting & predictive control | PMV (Fanger's model) | Comfort parameters (PMV, Temperature, Humidity), Energy/load | Maximum energy saving rate prediction ~31% and maintaining thermal comfort within pre-established comfort zone (PMV≈0) | [66] |
| 2017 | Machine learning-based thermal environment control development | ANN-based model | Predictive control | Individual's thermal preference/feedback | Comfort parameters (Temperature, Humidity), Energy/load | A total of up to 45% more energy savings and 44.3% better thermal comfort performance than the PMV model. | [67] |
| 2018 | A novel real-time automated HVAC control system built on top of an Internet of Things (IoT). | ANN MPL-based predictive model | Optimized setting & predictive control | Personal dissatisfaction level expressed by users[3] | Comfort parameters (Temperature, Humidity, $CO_2$ concentration/Air quality), Energy/load | Between 20% and 40% energy savings were achieved while maintaining temperature within the comfort range (except the pre-peak cooling hour). | [68] |
| 2019 | An indoor-climate framework development for air-conditioning and mechanical ventilation (ACMV) systems control in buildings | Self-layered feedforward-ANN | Predictive control & optimized setting | Thermal sensation index based on ASHRAE 7-point sensation scale | Comfort parameters (ACMV, Temperature, Humidity), Energy/load | An average of 36.5% energy saving was ensured, and 25ºC was found as the ideal comfort temperature with a minimum energy use. | [69] |
| 2019 | A novel optimization framework using a deep learning-based control for building thermal load. | Recurrent neural network (RNN) | Load prediction & optimized setting | Defined temperature setpoints | Comfort parameters (Temperature), Energy/load | Up to 12.8% cost savings compared with a rule-based strategy, while maintaining the users' thermal comfort during the occupied periods. | [70] |
| 2019 | A learning-based optimization framework development for HVAC systems in smart buildings | Deep neural networks | Predictive control & optimized setting | Predicted thermal comfort[4] | Comfort parameters (Temperature, Humidity), Energy/load | DDPG allowed higher degree of thermal comfort with an average value closer to the preset threshold of 0.5. As it could save 6% more energy than the baseline methods. | [71] |

---

[3] Thermal comfort is a function of temperature: $\phi_n = 1 - \left(\frac{|t_n^{room}-t_n^{set}|}{dev_n}\right)$

[4] The predicted value is by time slot t: $M_t = \Phi\left(T_t^{in}, H_t^{in}\right)$

**Table 3.** (Continued).

| YEAR | STUDY CASE | UNDERLYING AI/ML TOOLS | AI APPLICATION SCENARIO | THERMAL COMFORT METHOD | OPTIMIZATION OBJECTIVE | OUTCOMES & KEY RESULTS | REF. |
|---|---|---|---|---|---|---|---|
| 2020 | Hybrid data-driven approaches development for predicting building indoor temperature response in VAV systems. | MLR and ANN trained by Bayesian Regulation | Predictive control | Defined comfort zones | Comfort parameters (Temperature, VAV system, Heating/cooling), Energy/load | The proposed model allowed to improve the control and optimization of buildings space cooling | [72] |
| 2020 | A network-based deterministic model development to respond the ever-changing users' fickle taste that can deteriorate thermal comfort and energy efficiency in building spaces. | Fuzzy inference system (FIS), ANN | Thermostat on/off, ANN, ANN + FDM | PMV (Fanger's model) | Comfort parameters (PMV, Humidity), Energy/load | ANN-FDM showed significant results by improving thermal comfort by up to 4.3% rather than the thermostat model and up to 44.1% of energy efficiency rather than ANN model. | [73] |
| 2020 | ANN-based prognostic models' development for load demand (LD) prediction for a Greek island by capturing three different forecasting horizons: medium, short and very short-terms | Multilayer Perception ANN | Predictive control | Biometeorological human thermal comfort-discomfort index[5] | Comfort parameters (Humidity, Heating/ cooling), Energy/load | Both medium and short-term prognoses showed significant ability to predict LD by errors around 7.9% and 7.2% respectively. | [74] |
| 2020 | An intelligent-based ML model to predict the energy performances in heating loads (HL) and cooling loads (CL) of residential buildings. | ANN, Deep Neural Networks (DNN) | Predictive control | Maintaining comfort conditions[6] | Comfort parameters (Temperature, Humidity, Heating/cooling), Energy/load | Deep NN showed better results compared to ANN (i.e., HL and CL prediction), by applying state-based sensitivity analysis (SBSA) technique allowing to improve the model by selecting the most significant variables. | [75] |
| 2020 | A novel personal thermal comfort prediction method using less physiological parameters. | ANN-based model | Predictive control | Modified thermal sensation vote scale[7] | Comfort parameters (Temperature, Humidity, HVAC), Energy/load | The proposed model showed good prediction accuracy and stability by an average of 89.2% and a standard deviation around 2.0%. | [76] |
| 2020 | Investigating the performances and comparative analyses of combined on-demand and predictive models for thermal conditions control in buildings. | ANN combined with FIS | On-demand & predictive controls | PMV/PPD (Fanger's model) | Comfort parameters (PMV/PPD, Temperature), Energy/load | Combining the predictive and on-demand algorithms improved the energy efficiency from 13.1% to 44.4% and reduced the thermal dissatisfaction by 20% to 33.6%, compared to each independent model. | [77] |
| 2020 | A building intelligent thermal comfort control and energy prediction based on the IoT and artificial intelligence. | Back-propagation (BP) neural networks | Predictive control & optimized setting | PMV (Fanger's model) | Comfort parameters (PMV, Temperature, Humidity), Energy/load | The system performed better than traditional control on comfort and energy savings. Limitations: ~3% error between expected and actual values. | [78] |

**Table 4.** Summary of the works focusing on intelligent management of thermal comfort and energy in buildings using Fuzzy Logic Control (FLC).

| YEAR | STUDY CASE | UNDERLYING AI/ML TOOLS | AI APPLICATION SCENARIO | THERMAL COMFORT METHOD | OPTIMIZATION OBJECTIVE | OUTCOMES & KEY RESULTS | REF. |
|---|---|---|---|---|---|---|---|
| 1998 | Fuzzy controller development for improving thermal comfort and energy saving of HVAC systems. | Fuzzy logic control | Fuzzy control | PMV (Fanger's model) | Comfort parameters (PMV, Temperature, Humidity, HVAC), Energy/load, cost | 20% energy saving, and better comfort level at the lower cost (than provided by thermostatic techniques) | [42] |
| 1999 | Multi-objective supervisory control of building climate and energy. | Fuzzy logic control | Optimized setting | Pre-defined (standardized) temperature | Comfort parameters (Temperature, Humidity, $CO_2$ concentration/Air quality) | The proposed system allows the user to compromise solution (comfort requirements /energy saving) | [41] |
| 2001 | PMV-based fuzzy logic controller for energy conservation and indoor thermal requirements and of a heating system in a building space. | Fuzzy logic control | Fuzzy control | PMV/PPD (Fanger's model) | Comfort parameters (PMV/PPD, Temperature, Humidity, Heating/ cooling), Energy/load | By maintaining PMV index at 0 and PPD with a maximum threshold of 5%, fuzzy controller had better performance with a heating energy of 20% (compared to tuned PID control). | [79] |
| 2001 | Developing fuzzy controller for energy saving and occupants' thermal-visual comfort and IAQ requirements. | Fuzzy logic control | Fuzzy control | PMV (Fanger's model) | Comfort parameters (PMV, $CO_2$ concentration/Air quality, Lighting), Energy/load | Adaptive fuzzy PD showed better performance for energy consumption (up to 25-30%) and the PMV/CO2 responses, for visual comfort, the non-adaptive fuzzy PD was sufficient. | [40] |
| 2003 | Fuzzy controller for indoor environment management. | Fuzzy logic control | Fuzzy control | PMV (Fanger's model) | Comfort parameters (PMV, Temperature, $CO_2$ concentration, Lighting, Heating/cooling), Energy/load | Up to 20.1% heating and cooling energy saving using P-controller by maintaining PMV between 0 and 0.1 and CO2 concentration less than 20 ppm. | [80] |
| 2003 | Fuzzy control for indoor environmental quality, energy and cost efficiencies. | Fuzzy logic control | Fuzzy control | Defined ranges/ Preferred set-points variables | Comfort parameters (Temperature, Humidity, $CO_2$ concentration/Air quality), Energy/load, Cost | Fuzzy's approach showed its ability to deal with multivariate problems by collaborating expert knowledge for decision making at complex level. | [81] |

---

[5] Cooling power (CP) index: $CP = 1.163 \cdot (10.45 + 10 \cdot u^{0.5} - u) \cdot (33 - T)$ with {T: temperature [ºC] and u: wind speed [m/s]}
[6] Comfort conditions considered in the internal design of the buildings, i.e., clothing level of 0.6 Clo with internal temperature of 21ºC, 60% of humidity, 0.3 m/s air speed and 300 Lux lighting level.
[7] Thermal sensation vote classified into 5 categories {cold, cool, neutral, warm, hot}.

Table 4. (Continued).

| YEAR | STUDY CASE | UNDERLYING AI/ML TOOLS | AI APPLICATION SCENARIO | THERMAL COMFORT METHOD | OPTIMIZATION OBJECTIVE | OUTCOMES & KEY RESULTS | REF. |
|---|---|---|---|---|---|---|---|
| 2005 | Integrated indoor environment energy management system (IEEMS) implementation for buildings application | Fuzzy logic control (FLC) | Fuzzy control | PMV (Fanger's model) | Comfort parameters (PMV, Temperature, $CO_2$ concentration, Lighting), Energy/load | Up to 38% energy conservation in both buildings without compromising the indoor comfort requirements. | [82] |
| 2005 | Dynamic illumination and temperature response control in real time conditions. | Fuzzy logic control | Fuzzy control | Temperature preference set-point (by the user) | Comfort parameters (Temperature, Lighting, Heating/cooling), Energy/load, Cost | Adjusting automatically roller blind position and window geometry according to external weather enables to get closer to thermal-visual preferences, contributing to lower energy consumption for lighting, heating/cooling and cost-saving enhancement. | [83] |
| 2005 | Controller development to improve energy conservation with a constraint on the individual dissatisfactions of the indoor environment. | FLC based on kNN approximations | Gradient-based optimization | Degree of individual dissatisfaction $(DID(vote) = (1 + tanh(2|vote| - 3)/2)$ | Comfort parameters (DID, Temperature), Energy/load | Optimized HIYW presented better performance than OFSA (PPD exceeding 20% for ~15% of population and 50% for ~5%) to minimize the energy consumption. | [84] |
| 2006 | Adaptive fuzzy control strategy for comfort air-conditioning (CAC) system performance | Fuzzy logic control | Indirect fuzzy adaptive control | PMV (Fanger's model) | Comfort parameters (PMV, HVAC), Energy/load | The adaptive fuzzy controller could save almost 18.9% of energy, compared to PID controller. | [85] |
| 2007 | Fuzzy controller development for improving indoor environmental conditions while reducing energy requirements for building energy management system | Fuzzy logic control | Fuzzy control | PMV (Fanger's model) | Comfort parameters (PMV, Lighting, $CO_2$ concentration /Air quality), Energy/load | Using a suitable cost function for BEMS allowed to save energy at a level lower than recommended by the literature. Also, users were satisfied by adopting the fuzzy controller | [86] |
| 2011 | Fuzzy adaptive comfort temperature (FACT) model development for intelligent control of smart building. | Fuzzy adaptive control | Fuzzy control and optimized setting | Adaptive comfort model | Comfort parameters (Temperature), Energy/load | Using the FACT model with grey predictor in agent-based control system of a smart building, provided reasonable comfort temperature with less energy consumption to the customers | [43] |
| 2013 | Fuzzy method-based data-driven to model and optimize thermal conditions of smart buildings applications. | Fuzzy logic control (FLC) | Fuzzy control | Comfort temperature ranges (defined by the users). | Comfort parameters (Temperature), Energy/load | The type-2 fuzzy model performs better, with RMSE=12.55 compared to the linear regression model where the RMSE=17.64. | [87] |
| 2013 | Identifying building behaviors related to energy efficiency and comfort for an office building in the Pacific Northwest. | Fuzzy knowledge base | Fuzzy rule base & optimized setting | Comfort levels based on average zone temperature | Comfort parameters (Temperature), Energy/load | The developed framework was able to identify and extract complex building behavior, which improve the building energy management systems (BEMSs) by eliminating the low efficiency and low comfort behavior | [88] |
| 2014 | Deploying and evaluating a user-led thermal comfort driven HVAC control framework in office building on University of Southern California | Fuzzy predictive model | Predictive control | Personalized comfort profiles based on a thermal preference (TPI) scale | Comfort parameters (Temperature, Humidity, Lighting, $CO_2$ concentration /Air quality, HVAC), Energy/load | The developed framework showed promising results for energy saving and comfort improvement. 39% reduction in daily average airflow rates. | [89] |
| 2015 | Fuzzy logic-based advanced on–off control for maintaining thermal comfort in residential buildings | Fuzzy logic control (FLC) | Fuzzy control | Defined (desired) room temperature | Comfort parameters (Temperature), Energy/load | Compared to the conventional on–off controller, the proposed system had better control performance and saved energy. | [90] |
| 2018 | Combining a Comfort Eye sensor with a sub-zonal heating system control for building climate management | Fuzzy logic control (FLC) | Fuzzy PID-PMV control | PMV (Fanger's model) | Comfort parameters (PMV, Temperature, Humidity, Heating/cooling), Energy/load | Up to 17% energy savings with respect to the standard ON/OFF mono-zone control, thermal comfort has been slightly improved with a minimum deviation from the neutral condition. | [91] |
| 2019 | Developing a thermal sensation-based control method to improve thermal comfort and energy saving in indoor environment. | Fuzzy comprehensive evaluation method | Predictive control & optimized setting | Temperature set-points | Comfort parameters (Air temperature, humidity, $CO_2$ concentration, HVAC), Energy/load | Online monitoring of thermal sensation saved 13.8% in daily energy consumption, with an average satisfaction score of 5.56 of all subjects, compared to the set point-based control method. | [92] |
| 2019 | Developing a context-aware model for multi-objective decision making process in Ambient Intelligence (AmI): maintaining thermal comfort while optimizing energy-efficiency in an office building. | Fuzzy logic control (*L*-fuzzy extension) | Multi-objective decision making process & optimized setting | User preferences, defined temperature | Comfort parameters (Temperature, Lighting, ventilation/cooling), Energy/load | By defining the situation of interest, the system allowed to achieve thermal comfort conditions in most of the time, while saving 27% of energy, compared to a scenario without system use. | [93] |
| 2020 | Combining two different types for optimal time response in non-linear systems in smart buildings application. | Fuzzy PI-PD Mamdani type (FPIPDM) and Takagi-Sugeno-Kang (CABTSK) type. | Optimized setting | PMV-PPD (Fanger's model) | Comfort parameters (PMV-PPD, Relative humidity, Temperature, HVAC), Energy/Load, Cost | The integrated framework allowed to reduce $CO_2$ emissions and provide the required thermal comfort, while saving up to 37% of energy. Besides a significant reduction in time response in HVAC systems. | [94] |
| 2021 | Developing and experimental verification of thermal sensation prediction-based fuzzy control method for improving thermal comfort and energy saving of HVAC systems. | Fuzzy logic control (Mamdani fuzzy model and Functioning fuzzy subset inference (FFSI) method) | Predictive control & optimized setting | Temperature set-point, Thermal sensation votes, predicted thermal sensation (as function of physiological variables) | Comfort parameters (Temperature, Humidity, $CO_2$ concentration, HVAC, VAV system), Energy/Load | The proposed method allowed to better adjustment of temperature set-points according to thermal sensations of all occupants, with an average satisfaction score of 5.30, and saving around 20.07% of daily energy consumption (compared to temperature set-points method) and 10.73% (compared to thermal sensation feedback-based control) | [95] |

The third type of techniques is recognized as **Distributed Artificial Intelligence (DAI)** and **Multi-Agent Systems (MAS)** (cf. Table 5). MAS are derived from the Distributed AI, a branch of artificial intelligence. The DAI is structured around three axes:

- **Distributed problem solving**, allows the problem to be divided into a set of subproblems supported by distributed and collaborating entities and studies on how to share problem knowledge in order to find a solution.
- **Parallel AI**, develops parallel languages and algorithms to improve computer system performance.
- **Multi-agent systems**, promote a decentralized modeling approach and emphasize the collective aspects of systems.

The MAS approach, which has evolved considerably over the last twenty years, makes it possible to apprehend, model and simulate complex systems, i.e., systems involving multiple components that interact dynamically with each other and with the outside world. It looks at how to coordinate a set of entities called agents, so that they can collectively solve a global problem. Otherwise, the concept of agent refers to an autonomous entity evolving in interaction with its environment, which is often dynamic and unforecastable. The modeling and interactions of these agents were inspired by the observation of complex biological systems (e.g., organized animal societies such as ant colonies, bird swarms [96] or fish schooling [97]). MAS are therefore a privileged approach to addressing complex systems. Their entirely decentralized nature makes them particularly suitable for this type of system. MAS make it possible to work on the overall functioning of the system by looking at the entities that make it up and their interactions. MAS have been developed in a variety of areas including: image processing, robotics, simulation, among others.

In the building sector, the usage of agent-based and distributed intelligent energy-saving systems while maintaining a satisfactory indoor environment has been adopted in several works. For example, Klein L. et al. [98] proposed a Multi-Agent Comfort and Energy System (MACES) to coordinate equipment and occupants within a building. Also, Davidsson P. and Boman M. [99] presented a MAS for energy control in tertiary sector buildings. The purpose of this system is to provide three services: lighting, heating and ventilation, as well as minimizing office energy consumption. In addition, Mo Z. [100] has built an agent-based platform for individual users and buildings occupants in order to negotiate their control activities. Dounis A. I. and Caraiscos C. [101] suggested the use of an intelligent supervisor to arrange the optimal collaboration of local controller-agents. Consequently, overall control is reached, the occupants' preferences are met, disagreements are avoided and the energy consumption is reduced on a conditional basis. Another agent-based system control developed by Barakat M. and Khoury H. [102], which examines multi-comfort (visual, thermal and acoustic) level control designed to reduce energy consumption. In Ref. [103], an agent-based model was introduced to simulate the effect of occupant's behavior on comfort and energy usage in a residential building. The developed model showed a realistic estimation of the energy consumption levels.

Table 5. Summary of the works focusing on intelligent management of thermal comfort and energy in buildings using distributed AI (DAI) & Multi-Agent Systems (MASs).

| YEAR | STUDY CASE | UNDERLYING AI/ML TOOLS | AI APPLICATION SCENARIO | THERMAL COMFORT METHOD | OPTIMIZATION OBJECTIVE | OUTCOMES & KEY RESULTS | REF. |
|---|---|---|---|---|---|---|---|
| 2005 | Decentralized system development for controlling and monitoring an office building. | Agent-based approach | Distributed AI | Personal comfort based on individual preferences | Comfort parameters (Temperature, Lighting), Energy/load | The MAS approach allowed it to save up to 40% energy, compared to the thermostat approach, and ~12% compared to the timer-based approach. Reactive approach is more energy consuming than proactive, ensuring 100% of thermal satisfaction. | [99] |
| 2006 | Centralized HVAC with multi-agent structure. | Agent-based approach | Distributed AI and optimized setting | PMV (Fanger's model) | Comfort parameters (Temperature, Humidity, HVAC), Energy/load | The control accuracy goes around 89% to 92.5%. which means that the thermal comfort is predicted by 7.5% to 11% of error rate. | [104] |
| 2011 | Multi-agent simulation for building system energy and occupants' comfort optimization | Multi-agent system (MAS) | Distributed AI | PMV (Fanger's model) | Comfort parameters (PMV, Temperature), Energy/load | 17% energy savings while maintaining a high comfort level, approximately 85% occupants' satisfaction. | [105] |
| 2011 | Developing a MAS combined with an intelligent optimizer for intelligent building control. | Multi-agent system (MAS) | Optimized setting | Temperature set-point control | Comfort parameters (Temperature, Lighting, $CO_2$ concentration/Air quality), Energy/load | Implementing PSO optimizer allowed to maintain a high-level of overall comfort, i.e., mainly around 1.0, when the total energy supply was in shortage. | [106] |
| 2012 | Coordinating occupants' behaviors for thermal comfort improvement and energy conservation of an HVAC system. | Agent-based approach | Distributed AI | PMV (Fanger's model) | Comfort parameters (PMV, HVAC), Energy/load | Reducing 12% of energy consumption while maintaining 70%–75% occupant satisfaction for both proactive and proactive-MDP. | [98] |
| 2012 | Distributed AI control with information fusion-based Indoor energy and comfort management for smart building application. | Multi-agent approach | Distributed AI & optimized setting | Defined comfort range | Comfort parameters (Temperature, Lighting, $CO_2$ concentration/Air quality), Energy/load | All case studies showed the effectiveness of the system of the developed system in different operating scenarios. | [107] |
| 2013 | Intelligent management system development for energy efficient and comfort in building environments. | Agent-based approach | Distributed AI | Individual thermal comfort based on the indoor temperature | Comfort parameters (PMV, Temperature, Lighting, $CO_2$ concentration), Energy/load | Case studies simulation results showed that the developed MAS could manage comfort needs and reduce energy consumption simultaneously (PMV was kept around 0.61). | [108] |
| 2014 | A human and building interaction toolkit (HABIT) development for building performance simulation. | Agent-based model (ABM) | Distributed AI | Individual comfort distribution based on PMV (Fanger's model) | Comfort parameters (PMV, Temperature, Heating/cooling), Energy/load | Up to 32% reduction of total energy use in all zones in summer without significant increase in winter are expected, and a promising decrease in thermal discomfort in all zones in both seasons. | [109] |
| 2014 | NN-based approach with a MAS infrastructure to improve energy efficiency, while maintaining acceptable thermal comfort level for occupants of an academic building[8]. | MAS combined with gARTMAP | Distributed AI | Learning the user's thermal preferences | Comfort parameters (Temperature, Hot/cold water), Energy/Load | The proposed gARTMAP-MAS IHMS might use less heat to achieve the desired indoor temperature, compared to the existing rule-base BMS and fuzzy ARTMAP IHMS | [110] |
| 2015 | Multi-agent control architecture for cooling and heating processes in smart residential buildings. | Multi-agent system + ML algorithms | ML & distributed AI | Desired temperature based on occupant's behavior | Comfort parameters (Temperature, Heating/cooling), Energy/load | The proposed system allowed to significantly improve the occupants comfort with a slight increase in energy consumption, with respect to 'sense behavior' (compared to simple strategies) | [111] |

---

[8] UCLan's West Lakes Samuel Lindow Building Maizura

Another technique known as **Reinforcement Learning** (**RL**) and **Deep Reinforcement Learning** (**DRL**) has been used in building control (cf. Table 6). Reinforcement learning is a type of machine learning through which an intelligent agent learns on the basis of rewards (or reinforcements) that may be positive or negative depending on how the action is taken by the agent brings it closer to its goal [112]. In RL, the agent interacts with the environment and receives information from this interaction that helps to manage the environment better over time. In each interaction, the agent is in a state s, from a set of all possible $S$ ($s \in S$) states, and performs an action a, from a set of all possible $A$ ($a \in A$) actions. After performing the action, the agent goes to a new state $s'$ and receives a reward $r$ from the environment. This process can be shown in Figure 6.

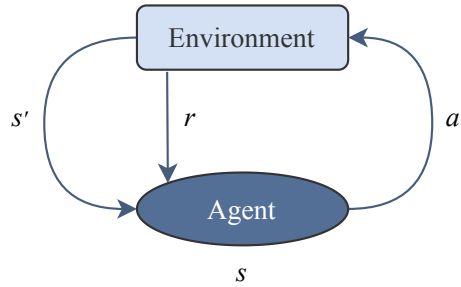

**Figure 6.** Illustration of the general framework of reinforcement learning.

The agent must carry out those actions that increase the total amount of received rewards, i.e., it is necessary to locate a movement policy that optimizes the accumulated reinforcement over the long-term. A policy $\pi$ is a mapping of states to actions that determines the probability $\pi(a|s)$ of an action being performed in a state $s$. This map is updated on the basis of the experience acquired by the agent during training.

**Advanced predictive control**, including predictive functions of ANN and Model-based Predictive Control (MBPC) (cf. Table 7), is a widely recognized comfort control technique using a model (system, noise, and disturbance) to predict future output. These predictions are integrated into the cost function of closed-loop action and control activity, which is reduced with regard to the sequence of anticipated signals, taking into account the problem constraints. Finally, a rolling-horizon strategy is implemented, applying at time k the control signal calculated for that time and repeating the calculations for the next sampling period. Many variants of these techniques have emerged and, within the context of this paper, we consider the most relevant in the field of comfort and energy management in built-environments, such as Linear MPC, Non-Linear MPC, Distributed MPC.

The application of model-based predictive control in building control systems can provide energy savings and more cost efficiency, than any other non-predictive control. Other benefits can be reached such as robustness to disturbances and changes, multi-variables control, improved steady-state response, prediction of future disturbances and control actions and many others [113,114].

Subsequently, we may cite another approach, described as **Hybrid** methods (cf. Table 8), resulting from a combination of intelligent techniques and classical or advanced techniques, such as: **FLC** and **Genetic Algorithm** (**GA**) [115–120], **MAS** and **FLC** [121,122], **ANN** and **GA** [123–125], among others [126–131]. Hybrid controllers are useful, since this incorporation can solve problems that cannot be solved by the individual controller. Nevertheless, the design of the "intelligent" component involves the expertise of the user and a large amount of training data, while the "classic" or "advanced" part is difficult to adjust (or tuning), particularly for HVAC systems, which is a constraint on the control system.

Table 6. Summary of the works focusing on intelligent management of thermal comfort and energy in buildings using Reinforcement Learning (RL) & Deep Reinforcement Learning (DRL).

| YEAR | STUDY CASE | UNDERLYING AI/ML TOOLS | AI APPLICATION SCENARIO | THERMAL COMFORT METHOD | OPTIMIZATION OBJECTIVE | OUTCOMES & KEY RESULTS | REF. |
|---|---|---|---|---|---|---|---|
| 2007 | Linear reinforcement learning controller (LRLC) for energy saving while sustaining comfort requirements. | Linear reinforcement learning | Machine Learning (ML) | PMV-PPD (Fanger's model) | Comfort parameters (PMV/PPD, Temperature, Humidity, $CO_2$ concentration/Air quality), Energy/load | Over a period of 4 years, training the LRLC, the energy consumption has been increased from 4.77Mwh to 4.85Mwh, however the PPD index has been decreased from 13.4% to 12.1%. | [132] |
| 2014 | Reinforcement learning for tenant comfort and energy use optimization in HVAC systems. | Q-learning based supervisory approach | Optimized setting | Occupant's comfort is learnt from the tenant preferences and occupancy patterns | Comfort parameters (Temperature, HVAC), Energy/load | Learning to adjust/schedule, appropriately, thermostat temperature setpoints for energy efficiency while keeping the tenant comfortable. | [133] |
| 2015 | Implementing and evaluating a multi-grid reinforcement learning method for energy conservation and comfort control of HVAC systems in buildings. | Multi-grid methods for Q-learning | Optimized setting | PPD-PMV (Fanger's model) | Comfort parameters (PMV/PPD, Temperature, Humidity, HVAC), Energy/load | The proposed multi-grid approach helped to accelerate the convergence of Q-learning, and performed better on energy saving and comfort than the constant grid versions. | [134] |
| 2017 | A deep reinforcement learning based data–driven approach development for building HVAC control. | Deep reinforcement learning (DRL) | Optimized setting | Desired temperature range based on ASHRAE standard | Comfort parameters (Temperature), Energy/load, Cost | Up to 20%-70% energy cost reduction while meeting the room temperature requirements, compared to a conventional rule-based approach. | [34] |
| 2017 | A reinforcement learning-based thermostat schedule controller development using long–short–term memory recurrent neural network for an office HVAC system. | Actor-critic RL + LSTM-RNN | Optimized setting | PMV (Fanger's model) | Comfort parameters (PMV, Temperature), Energy/load | An average 2.5% energy savings was achieved while improving thermal comfort by an average of 15%, compared to other control baselines (Ideal PMV & Control Variable). | [135] |
| 2018 | A novel type of decentralized and cooperative method development for decision-making strategies in the buildings' context, based on reinforcement learning. | Extended joint action learning (eJAL) | Distributed AI | Thermal comfort index as a function of indoor temperature | Comfort parameters (Temperature, Humidity, Lighting), Energy/load | The long-term learning analysis showed that Q-learning and eJAL gave acceptable comfort losses ($\Delta C \leq 0.4$), for demand/response balance, eJAL (Median=1.67) was slightly better than Q-learning (Median=2.21) | [136] |
| 2018 | Plug & play solution of an HVAC thermostat's set-point scheduling inspired by reinforcement learning technique | Neural Fitted Q-iteration (NFQ)-RL | RL-based control | PMV (Fanger's model) | Comfort parameters (PMV, Temperature, Humidity, HVAC), Energy/load | With energy/comfort trade-off balance, an average up to 32.4% energy savings and up to 27.4% comfort improvements in average. | [137] |
| 2018 | A whole BEM-DRL framework development for HVAC optimal control in a real office building (Intelligent Workplace) | Deep reinforcement learning (DRL) | Optimized setting | PPD (Fanger's model) | Comfort parameters (PPD, Hot/cold water, Temperature, Humidity), Energy/load | About 15% heating energy savings with similar comfort conditions as the base-case | [138] |
| 2019 | AI-based agent development for indoor environment control while optimizing energy use of air-conditioning and ventilation fans in a classroom and a laboratory | Deep RL (double Q-learning) | Optimized setting | PMV (Fanger's model) | Comfort parameters (PMV, Temperature, HVAC, $CO_2$ concentration/Air quality), Energy/load | AI-agent has successfully managed the indoor environment within an acceptable PMV values between -0.1 and +0.07, and 10% lower $CO_2$ levels, while reducing energy consumption by about 4% to 5%. | [139] |
| 2020 | An event-triggered paradigm based on RL approach for smart learning and autonomous micro-climate control in buildings. | Stochastic/deterministic policy gradient RL | Optimized setting & event-triggered control | Discomfort rate derived from desired temperature | Comfort parameters (Temperature, Heating/cooling), Energy/load | the proposed algorithms learn the optimal policy in an appropriate time, i.e., optimal thresholds were found $T_{ON}^{th} = 12.5°C$ and $T_{OFF}^{th} = 17.5°C$ resulting in an optimal rewards rate. | [140] |
| 2020 | A framework development for optimal control over AHUs by combining DRL methods and long-short-term-memory networks (LSTM). | Deep reinforcement learning (DRL) | Predictive control & optimized setting | PMV (Fanger's model) | Comfort parameters (PPD, Temperature, $CO_2$ concentration/Air quality), Energy/load | 27% to 30% lower energy consumption compared to rule-based control, while maintaining the average PPD at 10%. | [141] |

**Table 7.** Summary of the works focusing on intelligent management of thermal comfort and energy in buildings using Advanced predictive control methods.

| YEAR | STUDY CASE | UNDERLYING AI/ML TOOLS | AI APPLICATION SCENARIO | THERMAL COMFORT METHOD | OPTIMIZATION OBJECTIVE | OUTCOMES & KEY RESULTS | REF. |
|---|---|---|---|---|---|---|---|
| 2012 | Improving the energy efficiency in an AC by reducing transient and steady-state electricity consumption on BRITE[9] platform. | Learning-based model predictive control | Predictive control | Comfort specifications based on OSHA guidelines | Comfort parameters (Temperature, HVAC), Energy/load | 30%–70% reduction in energy consumption while maintaining comfortable room temperature by keeping temperature close to the specified comfort middle (22ºC) | [142] |
| 2012 | Model-based predictive control development for thermal comfort improvement with auction of available energy of a limited shared energy resource in three houses. | Distributed model predictive control | Predictive control | Defined comfort temperature bounds | Comfort parameters (Temperature), Cost, Energy/load, | The developed system is flexible, allowing the customer to shift between comfort and lower cost. | [143] |
| 2012 | A discrete model-based predictive control for thermal comfort and energy conservation in an academic building. | MBPC based (RBF) ANN | Discrete models-based predictive control | PMV (Fanger's model) | Comfort parameters (PMV, Temperature, Humidity), Energy/load | Up to 50% energy savings are achieved by using the MBPC, which provides good coverage of the thermal sensation scale, when used with radial basis function-NN models. | [144] |
| 2013 | Model-based predictive control development for optimal personalized comfort and energy consumption management in an office workplace | Learning-based model predictive control | Predictive control & optimized setting | PPV function defined as an affine transform of PMV index | Comfort parameters (PPV-PMV, Temperature), Energy/load | About 60% energy savings when compared with fixed temperature set-point, and discomfort reduction from 0.36 to 0.02 compared to baseline methods. | [145] |
| 2014 | Predicting an integrated building heating and cooling control based on weather forecasting and occupancy behavior detection in the Solar House test-bed in real-time located in Pittsburgh. | Nonlinear model predictive control | Predictive control & optimized setting | Personalized thermal comfort (based on occupancy and weather) | Comfort parameters (Temperature, Humidity, HVAC, Lighting, $CO_2$ concentration), Energy/load | 30.1% of energy reduction in the heating season, besides 17.8% in the cooling season. NMPC allowed reducing time not met comfort (from 4.8% to 1.2% in the heating season, and from 2.5% to 1.2% for the cooling season). | [146] |
| 2015 | Hybrid predictive control model development for energy and cost savings in a commercial building (Adelaide airport). | Linear MPC combined with ANN | Hybrid predictive control | Defined comfort range based on ASHRAE | Comfort parameters (Temperature, Hot/cold water), cost, Energy/load | About 13% of energy cost saving was achieved and up to 41% of energy saving, compared to the baseline control. | [147] |
| 2016 | Simulation-based MPC procedure for multi-objective optimization of HVAC system performance and thermal comfort, applied to a multi-zone residential building in Naples. | Model-based predictive control | Predictive control & optimized setting | $PPD^{MAX}$: the maximum hourly value of PPD (Fanger's model) | Comfort parameters (PPD, HVAC), cost, Energy/load | Up to 56% operating cost reduction and improvement in thermal comfort, compared to the standard control strategy. | [148] |
| 2020 | A novel MPC relied on artificial intelligence-based approach development for institutional and commercial buildings control. | MPC relied on AI-based approach | Predictive control & optimized setting | Predefined set-point ramps (temperature) profiles | Comfort parameters (Temperature, Heating), Time efficiency, Energy/load | Reduction of the natural gas consumption and the building heating demand by 22.2% and 4.3% resp. Improving thermal comfort, while minimizing the required amount of time and information, compared with *business-as-usual* control strategies. | [149] |
| 2020 | A neural network-based approach for energy management and climate control optimization of buildings (applied to two-story building in Italy). | MPC with ANN-based models | Predictive control & optimized setting | Constant set-point temperature (defined as $T_{ref} = 25°C$) for each zone. | Comfort parameters (Temperature, Humidity), Energy/load | The proposed model showed significant results in energy savings (5.7% energy reduction of one zone) and better comfort compared to the baseline controller. | [150] |

**Table 8.** Summary of the works focusing on intelligent management of thermal comfort and energy in buildings using the Hybrid methods.

| YEAR | STUDY CASE | UNDERLYING AI/ML TOOLS | AI APPLICATION SCENARIO | THERMAL COMFORT METHOD | OPTIMIZATION OBJECTIVE | OUTCOMES & KEY RESULTS | REF. |
|---|---|---|---|---|---|---|---|
| 2002 | Controller development for indoor environmental conditions management for users' satisfaction while minimizing energy consumption inside a building. | GA-based fuzzy control | Optimized setting | PMV (Fanger's model) | Comfort parameters (PMV, Temperature, lighting, $CO_2$ concentration/Air quality, Humidity)/ Energy/load | Overall energy saving up to 35%, with a steady-state error of 0.5 for PMV, ~ 80ppm for $CO_2$, and ~80 lx for illuminance (after applying GA). | [120] |
| 2003 | Developing controller for HVAC system to improve indoor comfort requirements and energy performance in two real sites. | GA-based fuzzy control | Optimized setting | PMV (Fanger's model) | Comfort parameters (PMV, Temperature, $CO_2$ concentration/Air quality), Energy/load | While maintaining steady-state indoor conditions, the developed controller showed best experimentation results in the real test cells, with up to 30% energy saving for CNRS–ENTPE case and 12.5% for ATC (anonymous enterprise). | [115] |
| 2005 | Development of fuzzy rule-based controller using GA for HVAC system | GA-based fuzzy control | Optimized setting | PMV (Fanger's model) | Comfort parameters (PMV, Temperature, HVAC, $CO_2$ concentration/Air quality), Energy/load | By considering the rule weights and rule selection, results showed that the FLC controller presented improvement by 14% in energy saving and about 16.5% in system stability. | [116] |
| 2007 | Development of an intelligent coordinator of fuzzy controller-agents (FCA) for indoor environmental control conditions using 3-D fuzzy comfort model | Agent-based FLC | Intelligent system-based fuzzy control | PMV (Fanger's model) | Comfort parameters (PMV, Lighting, $CO_2$ concentration /Air quality), Energy/load | The combined controller[10] showed significant results by maintaining the controlled variables in acceptable ranges (PMV between -0.5 and +0.6) besides up to 30% of energy savings | [121] |

---

[9] Berkeley Retrofitted and Inexpensive HVAC Testbed for Energy Efficiency
[10] Combining the fuzzy controller-agent (FCA) with the intelligent coordinator (IC)

Table 8. (Continued).

| YEAR | STUDY CASE | UNDERLYING AI/ML TOOLS | AI APPLICATION SCENARIO | THERMAL COMFORT METHOD | OPTIMIZATION OBJECTIVE | OUTCOMES & KEY RESULTS | REF. |
|---|---|---|---|---|---|---|---|
| 2011 | Intelligent control system development to optimize comfort and energy savings using soft computing techniques for building application | GA-based fuzzy control | Optimized setting | PMV (Fanger's model) | Comfort parameters (PMV, Lighting, $CO_2$ concentration /Air quality), Energy/load | The proposed system has successfully managed the user's preferences for comfort requirements and energy consumption (while maintaining PPD < 10%). | [117] |
| 2011 | Controller development for a heating and cooling energy system | GA-based fuzzy control | Predictive control | Fixed set-point temperature for the thermal zone (24ºC) | Comfort parameters (Temperature, Heating/cooling, cost, Energy/load | The proposed methodologies allowed to achieve higher energy efficiency and comfort requirements by lowering equipment initial and operating costs up to 35%, and comfort costs up to 45%. | [119] |
| 2013 | Intelligent control system deployment for energy and comfort management in commercial buildings | MAS + FLC | Distributed AI & Fuzzy control | User preferences (temperature setpoint) | Comfort parameters (Temperature, Lighting), Energy/load | Up to 0.9 is achieved by comfort factors, i.e., the customer's satisfaction is ensured. The GA-based optimization allowed to minimize the energy consumption | [122] |
| 2014 | Improving the fuzzy controller's performance for comfort energy saving in HVAC system | GA-based fuzzy control | Fuzzy control & optimized setting | Individual comfort classes: ISO 7730 based on PMV/PPD (Fanger's model) | Comfort parameters (PMV/PPD, HVAC), Energy/load | The overall energy consumption is decreased by 16.1% in case of cooling and 18.1% in case of heating. Also, the PMV is reduced from -0.3735 to -0.3075 compared to EnergyPlus. | [118] |
| 2014 | Stochastic optimized controller development to improve the energy consumption and indoor environmental comfort in smart buildings | MAS + GA | Distributed AI and optimized setting | User preferences (temperature setpoint) | Comfort parameters (Temperature, Lighting, $CO_2$ concentration), Energy/load | Overall occupant comfort with GA was kept between 0.97 and 0.99, and the error between setpoints and the sensor data became smaller with GA. A significant reduction in the overall energy consumption (~20% compared to system without GA) | [126] |
| 2015 | Agent-based particle swarm optimization development for inter-operation of Smart Grid-BEMS framework | Agent-based approach | Distributed AI & optimized setting | Comfort was modeled as a temperature Gaussian function | Comfort parameters (Temperature, Humidity, $CO_2$ concentration), Energy/load | The proposed system could effectively improve the voltage profile of the feeder, while ensuring acceptable comfort levels. | [127] |
| 2016 | Deploying an intelligent MBPC solution for HVAC systems in a University building | MOGA framework + RBF-NN | Predictive control | PMV (Fanger's model) | Comfort parameters (PMV, Temperature, Humidity, HVAC), Cost, Energy/load | The IBMPC HVAC showed significant results in reducing energy cost and maintaining thermal comfort level during the whole occupation period. | [124] |
| 2018 | Optimizing the passive design of newly-built residential buildings in hot summer and cold winter region of China | NSGA-II combined with ANN | Optimized setting | Annual indoor thermal indices: CTR and DTR[11] | Comfort parameters (Comfort indices: CTR/DTR), Energy/load | The annual thermal comfort hours were extended by 516.8–560.6 hours, and the annual building energy demand was reduced by 27.86–33.29% compared to base-case design | [123] |
| 2018 | A demand-driven cooling control (DCC) based on machine learning techniques for HVAC systems in office buildings. | k-means clustering & kNN | ML & predictive control | Predefined comfort conditions (temperature setpoints) | Comfort parameters (HVAC, Temperature, Humidity, $CO_2$ concentration/Air quality), Energy/load | Between 7% and 52% energy savings were ensured compared to the conventionally-scheduled cooling systems (by maintaining temperature deviations means less than 0.1ºC). | [128] |
| 2020 | Comfort and energy management of daily and seasonally used appliances for smart buildings application in hottest areas. | Binary-PSO + FLC (BPSOFMAM[12], BPSOFSUG[13]) | Fuzzy logic and optimization setting | Fanger's PMV method | Comfort parameters (PMV, Temperature), Energy/load | Simulation results showed that the BPSOFSUG controller outperformed the BPSOFMAM in terms of energy efficiency by 16%, while comfort computation, via PMV, was kept in satisfactory range. | [151] |
| 2020 | A multi-objective optimization method for a passive house (PH) design by considering energy demand, thermal comfort and cost. | Combining: RDA, GBDT and NSGA-II | Optimized setting | The annual cumulative comfort ratio (CTR)-based adaptive model[14] | Comfort parameters (CTR index), Cost, Energy/load | The optimization results showed around 88.2% energy savings rate and improvement in thermal comfort by 37.7% compared to base-case building. | [131] |
| 2020 | A predictive model for thermal energy by integrating IoT architecture based on *Edge* Computing and classifier ensemble techniques for smart buildings application. | Combining: SVM, LR and RF | Predictive control | Indoor temperature set by the user or by the learning algorithm | Comfort parameters (Temperature, Humidity, $CO_2$ concentration/Air quality, lighting), Energy/load | Simulation results showed that the proposed approach presented the highest accuracy, by 91.526% compared to neural networks, ensemble RF and SVM. | [129] |
| 2020 | A novel optimization method for building environment design by integrating a GA, an ANN, MRA and an FLC based on the results of computational fluid dynamics (CFD) analysis. | Combining GA + ANN + multivariate regression analysis (MRA) + FLC | Optimized setting | PMV (Fanger's model) | Comfort parameters (PMV, Temperature, Cost, Time efficiency, Energy/load | Integrating GA, ANN, MRA and FLC in the design process allowed to reduce the variable space and computational cost by 50% and 35.7% respectively. | [125] |
| 2020 | An energy flexibility quantification methodology based on supervised machine learning techniques for hybrid demand-side control for high-rise office building. | MLR + SVR + backpropagation NN | Predictive control | Indoor setpoint temperature | Comfort parameters (Temperature, Hot/cold water), Time efficiency, Energy/load | The hybrid controller allowed to reduce the time duration of the peak power, which was reduced by 61% of the grid importation | [130] |

---

[11] The annual indoor thermal Comfort Time Ratio: $CTR\,[\%] = \frac{1}{n} \times \sum_{i=1}^{n} \frac{N_i}{8760} \times 100$, and Discomfort Time Ratio $DTR = 100 - CTR$.

[12] BPSOFMAM: Binary Particle Swarm Optimization Fuzzy Mamdani

[13] BPSOFSUG: Binary Particle Swarm Optimization Fuzzy Sugeno

[14] The annual cumulative comfort ratio (CTR)-based adaptive model: $CTR = \frac{1}{m}\sum_{k=1}^{k=m}\left(\sum_{j=1}^{N_p} wf_j \cdot \frac{1}{N_p}\right)^m \in [0,1]$.

Table 9. Summary of the works focusing on intelligent management of thermal comfort and energy in buildings using other AI-assisted tools.

| YEAR | STUDY CASE | UNDERLYING AI/ML TOOLS | AI APPLICATION SCENARIO | THERMAL COMFORT METHOD | OPTIMIZATION OBJECTIVE | OUTCOMES & KEY RESULTS | REF. |
|---|---|---|---|---|---|---|---|
| 1993 | An intelligent operation support system (IOSS) to improve HVAC operations for IAQ control and energy saving for industrial application. | Knowledge-based system (KBS) | Optimized setting | PMV (Fanger's method) | Comfort parameters (PMV, HVAC), Time Efficiency, Energy/load | The developed system can provide real-time planning, and assisting the interaction between the operator and the HVAC process | [152] |
| 2004 | Two-objective optimization of HVAC system control with two variable air volume (VAV) systems. | Genetic Algorithm (GA) | Optimized setting | PMV-PPD (Fanger's model) | Comfort parameters (PMV/PPD, Temperature, HVAC), Energy/load | The on-line implementation of GA optimization allowed to save up to 19.5% of energy consumption while minimizing the zone airflow rates and satisfying thermal comfort | [153] |
| 2007 | Modelling indoor temperature using autoregressive models for intelligent building application. | Autoregressive exogenous (ARX) | Predictive control | Black-box model to predict indoor temperature based on defined variables[15] | Comfort parameters (Temperature, Humidity), Energy/load | Results showed that ARX model gave better temperature prediction than ARMAX model by the structure $ARX(2,3,0)$ with a coefficient of determination of 0.9457 and the $ARX(3,2,1)$ with a coefficient of determination of 0.9096. | [154] |
| 2009 | Exploring the impact of optimal control strategies of a multi-zone HVAC system on the energy consumption while maintaining thermal comfort and IAQ of a built environment. | Genetic Algorithm (GA) | Optimized setting & predictive control | PMV (Fanger's model) | Comfort parameters (PMV, Temperature, HVAC, Air quality), Cost, Energy/load | Up to 30.4% savings in energy costs when compared to conventional base strategy whilst sustaining comfort and indoor air quality | [155] |
| 2009 | Estimating occupant mental performance and energy consumption of determining acceptable thermal conditions under different scenarios. | Bayesian Networks (BN) | Predictive control | PMV (Fanger's model) and the adaptive comfort model | Comfort parameters (PMV, Temperature), Energy/load | Results concluded that determining acceptable thermal conditions with the adaptive model of comfort can result in significant energy saving with no large consequences for the mental performance of occupants. | [156] |
| 2010 | Energy consumption optimization and thermal comfort management using data mining approach in built environment | Decision tree classifier (C4.5 Algorithm) | Optimized setting & predictive control | Comfort levels based on CIBSE[16] standard | Comfort parameters (Temperature, $CO_2$ concentration/Air quality, Humidity), Energy/load | Based on decision tree analysis and results relying ambient environmental conditions with user comfort, designers and facility managers can determine the optimal energy use | [157] |
| 2014 | Improving HVAC systems operations by coupling personalized thermal comfort and zone level energy consumption for selecting energy-aware and comfort-driven set-points. | Knowledge-based approach | Optimized setting | Personalized comfort profiles | Comfort parameters (Temperature), Energy/load | About 12.08% (57.6m3/h) average daily air-flow rates were reduced in three target zones, compared to operational strategies that focus on comfort only. | [158] |
| 2016 | Simulation-based multi-objective optimization for building energy efficiency and indoor thermal comfort | MOABC optimizer | Optimized setting | PPD (Fanger's model) | Comfort parameters (PPD, Temperature, Heating/cooling), Energy/load | The multi-objective optimization + TOPSIS showed that in different climates, even the energy consumption increased a bit by 2.9-11.3%, the PPD significantly reduced by 49.1-56.8%, compared to the baseline model. | [159] |
| 2016 | An operation collaborative optimization framework development for a building cluster with multiple buildings and distributed energy systems while maintaining indoor thermal comfort | Multi-objective optimization (PSO) | Optimized setting | PMV (Fanger's model) | Comfort parameters (PMV, Temperature), Cost, Energy/load | Around 12.1–58.3% of energy cost saving under different electricity pricing plans and thermal comfort requirements. | [160] |
| 2016 | Multi-objective control and management for smart energy buildings | Hybrid multi-objective GA | Optimized setting | Discomfort parameter based on the user preferences | Comfort parameters (Temperature, Lighting, $CO_2$ concentration/Air quality), Energy/load | 31.6% energy saving could be achieved for smart control building, and the comfort index was improved by 71.8%, compared to the conventional optimization methods. | [161] |
| 2016 | Real-time information-based energy management controller development for smart homes applications | Genetic Algorithm | Optimized setting | User preferences | Comfort parameters (Temperature), Cost, Energy/load | The proposed algorithms are flexible enough to maintain the user's comfort while reducing the peak to average ratio (PAR) and electricity cost up to 22.77% and 22.63% resp. | [162] |
| 2017 | A personalized thermal comfort model (BCM) development for smart HVAC systems control | Bayesian Network-based model | Optimized setting & predictive control | Personalized comfort model (combining the static and the adaptive models) | Comfort parameters (HVAC, Temperature), Energy/load | By using alternative comfort scale, the proposed model outperformed the existing approaches by 13.2%–25.8%. The heating algorithm allowed to reduce energy consumption by 6.4% to 10.5% for heating, and by 15.1% to 39.4% for air-conditioning, while reducing discomfort by 24.8%. | [35] |
| 2017 | A newly developed Epistemic-Deontic-Axiologic (EDA) agent-based solution supporting the energy management system (EMS) in office buildings | Support vector machine (SVM & C-SVC) | Distributed AI & ML | Personal thermal sensation model and Group-of-people-based thermal sensation model | Comfort parameters (Temperature, Humidity), Energy/load | Case studies simulations showed the abilities of the developed model in energy saving by 3.5–10%, compared to the preset control systems, while fulfilling the individual thermal comfort requirements (mean value of TSV in [-0.5, +0.5]). | [163] |

---

[15] Outside air temperature ($T_o$); Global solar radiation flux ($R_a$); Wind speed ($V_w$); Outside air relative humidity ($R_{HO}$).
[16] CIBSE – Chartered Institution of Building Services Engineers (https://www.cibse.org/)

**Table 9.** (Continued)

| YEAR | STUDY CASE | UNDERLYING AI/ML TOOLS | AI APPLICATION SCENARIO | THERMAL COMFORT METHOD | OPTIMIZATION OBJECTIVE | OUTCOMES & KEY RESULTS | REF. |
|---|---|---|---|---|---|---|---|
| 2017 | Deploying a software application based mobile sensing technology (Occupant Mobile Gateway (OMG)) for occupant-aware energy management of mix of buildings in California | Logistic regression (LR) | ML & predictive control | Occupants' subjective feedbacks | Comfort parameters (Temperature, Humidity), Energy/load | Implementing occupant-driven models showed that thermal management learned by subjective feedback had the potential energy savings while maintaining acceptable levels of thermal comfort | [164] |
| 2017 | An HVAC optimization framework deployment for energy-efficient predictive control for HVAC systems in office buildings | Random Forest (RF) regression | Predictive control and optimized setting | Comfort ranges defined by Royal Decree 1826/2009. | Comfort parameters (Temperature, Humidity, HVAC), Energy/load | Next 24h-Energy framework allowed reduce energy consumption for heating (48%) and cooling (39%), without affecting the user's comfort. | [165] |
| 2018 | The benefits of including ambient intelligent systems for building's EMS control to optimize the energy/comfort trade-off | k-means algorithm | Optimized setting | Occupants' preferences | Comfort parameters (HVAC), Energy/load | The energy consumption was reduced by an average of 5KWh while maintaining the majority of the occupants within acceptable comfort levels (the comfort rate was 5% lower than the baseline). | [166] |
| 2018 | Agent-based control system for and optimized and intelligent control of the built environment | Evolutionary MOGA | Distributed AI & optimized setting | User preferences | Comfort parameters (Temperature, lighting, Humidity, $CO_2$ concentration/Air quality), Energy/load | By applying MOGA optimizer allowed to save up to 67% energy consumption and about 99.73% overall comfort improvement. | [167] |
| 2020 | Thermal comfort control relying on a smart WiFi-based thermostat deployment for residential applications | Nonlinear Autoregressive exogenous (NARX) | Linear-based predictive control | Fanger's PMV method | Comfort parameters (PMV, Heating/cooling, Temperature, Humidity), Energy/load | In both High- and low-efficiency residences, cooling energy savings were around 85% and 95% respectively, while the PMV index was maintained within the desired range [0 – 0.5]. | [168] |
| 2020 | Defining new occupant comfort ranges using Bayesian-based data-driven approach for U.S. office buildings using the ASHRAE global thermal comfort database | Bayesian Inference (BI) (Bayes Theorem) | Active learning and data-driven control | Setpoint temperatures/ Occupants' feedback | Comfort parameters (HVAC, Temperature) | Data-driven and Bayesian approach allowed to reach realistic setpoint temperature values which facilitate more building performance, load prediction, and better informing better HVAC design as well as technology selection. | [169] |

In addition, there are other AI-based methods including: **Genetic Algorithm** method [153,155,161,162,167], **Knowledge-Based System (KBS)** [152,158] for reasoning and resolving complex problems, **Autoregressive Exogenous (ARX)** technique [154,168], **Bayesian Network (BN)** [35,156], **Decision Tree (DT)** [157], **Multi-Objective Artificial Bee Colony (MOABC)** and **Multi-Objective Particle Swarm Optimization (MOPSO)** [159,160] for multi-objective optimization control strategies, **Radial Basis Function (RBF)** [116,144], **Support Vector Machine (SVM/C-SVC)** [129,163], **Logistic regression (LR)** [164], **Random Forest (RF)** [129,165], and **k-Nearest Neighbor (kNN)** [145] for classification purpose, **K-means** algorithm for clustering [166], and the **Hidden Markov Model (HMM)** for modeling [64], while the **Bayesian Inference (BI)** which is useful to quantify the uncertainty in the estimated parameters of a given model [169].

Table 10 gives a summary of the main AI tools from the reviewed papers with their applicability in energy and comfort management.

**Table 10.** Comparison of the main AI/ML techniques.

| AI/ML TECHNIQUE | ADVANTAGES | LIMITATIONS | APPLICABILITY IN ENERGY & COMFORT SERVICES |
|---|---|---|---|
| NEURAL NETWORKS | • Ability to handle a large number of input variables [170]. <br> • Ability to handle a large amount of input data [170]. <br> • Ability to represent any function, linear or not, simple or complex. <br> • Faculty of learning from representative examples, by "retro propagation of Errors" [37]. <br> • Resistance to noise or unreliable data. <br> • Less bad behavior in case of small amount of data. | • Lack of a systematic method to define the best network topology and the number of neurons to be placed in the hidden layer(s) [171]. <br> • Difficulty in the choice of the initial values of the network weights and the setting of the learning step, which play an important role in the speed of convergence [171]. <br> • Problem of overlearning (learning at the cost of generalization) [171]. | • ANN methods are assumed to be more reliable for the prediction of energy consumption in HVAC systems [170]. <br> • ANNs can be introduced to define the notion of thermal comfort, in cases where the calculation of PMV index in not feasible. <br> • ANN can be used to calculate the optimal time to start heating after a period of unoccupancy [52]. <br> • ANN are usually combined with GA for optimization purpose [46,64]. |
| FUZZY LOGIC | • No need for modeling (however it may be useful to have a suitable model) [172]. <br> • Ability to implement (linguistic) knowledge of the process operator [172]. <br> • Expertise in systems to be tuned with complex behavior (highly non-linear and difficult to model) [39]. <br> • Frequently obtaining better dynastic performance (non-linear regulator) [172]. <br> • Can also be used for fast processes (via dedicated processors) [39]. <br> • Ability to manage uncertainty and imprecision, and the ability to model reasoning mechanisms and human decision-making. | • They can only use a limited number of inputs since the increased number of membership functions and fuzzy inference rules makes the system more complex to solve [173]. <br> • Lack of precise guidelines for the design of a setting (choice of quantities to be measured, determination of fuzzification, inferences and defuzzification) [173]. <br> • The implementation of the operator's knowledge often difficult [173]. <br> • The possibility of limit cycles due to non-linear operation [173]. <br> • The precision of the adjustment is often low. <br> • The consistency of inferences is not guaranteed a priori (appearance of contradictory inference rules possible) [173] | • Fuzzy logic can express thermal comfort in a linguistic way and, hence, can describe thermal comfort levels rather than temperature or humidity levels, resulting in greater efficiency. <br> • FLC methods are well suited for non-linear systems, e.g., HVAC systems. They are able to uniformly approximate a nonlinear function to any degree of accuracy and also provide fast operation [39]. <br> • Used to overcome the non-linearity of PMV index which can cause difficulties when monitoring HVAC systems (by calculating PMV values from measured data). |
| DAI/MAS | • Ability to scale an architecture, as multiple agents can dynamically add or remove themselves from a system [174]. <br> • Ability to be automatically configured [174]. <br> • System flexibility: Systems are composed of several agents that can therefore solve different problems [174]. <br> • MAS reflect the reality that the majority of problems are distributed, which easily fit into the MAS. <br> • MAS can have a great diversity of constituent agents, which gives designers the possibility to integrate different agents (reactive, cognitive, etc.) [174]. <br> • MASs can cooperate with each other to solve more complex problems [174]. <br> • Each agent has his own way of solving problems (the same problem may have different solutions for different agents) [174,175]. | • Lack of security: agents can communicate without any control. <br> • As the agents are executed in parallel, it is even more difficult to understand how they work from their code, due to the inherent nondeterminism of parallelism [175]. <br> • MASs are complex software, difficult to understand and design [175]. | • Used to negotiate the control activities in smart buildings control subsystems [98]. <br> • Used to coordinate equipment and occupants' services in buildings [99]. <br> • Used to simulate social behaviors, in building energy optimization, without affecting occupants' comfort [99]. <br> • Agent-based control can be used to schedule heating/cooling operation by considering the knowledge of human behavior in smart building applications [176]. <br> • Agent-oriented approach based on thermal sensation can be used to define occupant behavior (the occupants are modeled as autonomous agents) [111] |

| AI/ML TECHNIQUE | ADVANTAGES | LIMITATIONS | APPLICABILITY IN ENERGY & COMFORT SERVICES |
|---|---|---|---|
| RL/DRL | <ul><li>It is useful to solve very complex problems [177].</li><li>It allows to achieve long-term results [177].</li><li>It allows the correction of errors that occur during the training process [177].</li><li>It allows machines to automatically determine the ideal behavior in a specific context in order to maximize their performance [177].</li></ul> | <ul><li>Requires a lot of data/computation.</li><li>It is often too expensive in memory, because it has to store values for each state. As the problems can be quite complex, it can thus become very expensive in memory [177].</li><li>Developing this technology consists of implementing value approximation techniques, e.g., decision trees or NN.</li><li>The problems are also generally very modular; similar behaviors often reappear. It is therefore very often impossible to determine the current state entirely. This affects the performance of the algorithm [177].</li></ul> | <ul><li>RL (or DRL) methods are applied for continuous sensors inputs/actions enabling an approximate real-world HVAC operation [138].</li><li>Used to learn the state-value for optimal control of heat-pump, HVAC systems, or water heaters while using sensory data for energy and comfort optimization [132].</li><li>Used to intelligently learn the efficient strategy for building HVAC systems operation.</li></ul> |
| ADVANCED MPC | <ul><li>Robust to disturbances and changes [178,179].</li><li>Ability to simplify the analysis of stability and robustness of the corrector [180].</li><li>Multi-variables control [181,182].</li><li>Improved steady-state response [180].</li><li>Prediction of future disturbances [180].</li><li>Prediction of future control actions [180].</li><li>Ability to consider the constraints during command synthesis [180].</li><li>The formulation of MPC is preferable for large processes composed of several sub-systems.</li></ul> | <ul><li>The difficulty related to online computing time is a barrier to the use of this technique in certain areas [180].</li><li>For large buildings, the centralized predictive approach creates computational complexity issues.</li><li>Its main drawback in building control, is the need for a model of the process to be controlled and the relatively high costs of implementation [180].</li><li>Difficulty to obtain a mathematical model have long penalized the use of 'predictive' controls in buildings</li></ul> | <ul><li>They can bring energy savings in buildings control systems [142,144].</li><li>Applicable to reduce peak electricity demand in buildings.</li><li>It makes it possible to take into account future changes in climate in order to optimize control over time.</li><li>MPC controllers usually perform better in heating/cooling systems than PID or FLC alone [145,150].</li><li>In energy performance prediction, MPC usually use simplified resistance-capacity (RC) models.</li></ul> |
| HYBRID METHODS | <ul><li>Hybrid regulators are beneficial as the combination can solve problems that cannot be solved by the individual regulator [114].</li></ul> | <ul><li>The design of the "intelligent" part requires user experience and a huge amount of training data. While the "conventional" or "advanced" part is difficult to set-up [114].</li></ul> | <ul><li>Combined models always perform better than a single model, as they showed significant results in reducing energy consumption, maintaining comfort conditions, user preferences, etc. [119,122,151].</li></ul> |

## 3.2. *Optimized Functions in Support of AI-based control*

Building designers increasingly need to use simulation tools to analyze performance of scenarios for the purpose of understanding how strategies reduce environmental impact, ameliorate energy usage, and enhance comfort in buildings. These techniques can also be used to infer the adequate parameters to the AI-based building control. Another application of optimization function in AI-based control is the use of co-simulators that provide the possible parameters in real-time to the AI-based controller for optimum operation. The co-simulator has a global view of the system, while the controller has a local view of the sub-system and they complement each other in the entire process of optimizing AI-based energy and thermal comfort of sustainable buildings.

The introduction of these optimization techniques to the design simulation field started in the 1980's, and gained renewed interest from the 2000's [14,157]. In [157], the authors reported an increase in the number of scientific works on building model optimization since 2005. This reflects the interest and importance given to the development and application of numerical optimization methods by the building community around the world. At this point, the focus of this work is not to make a literature review of all these methods, but rather to present the most advanced and adopted optimization techniques in AI-assisted building control, in particular the **Genetic Algorithm** (**GA**) [46,49,52,64,65,115–118,120,123,125,126,131,148,153,155,161,162,167] and **Particle Swarm Optimization** (**PSO**) [43,51,61,61,107,127,151,160].

Genetic algorithms are an optimization technique that imitates the evolution of species through natural selection in a very simplified way. In genetic algorithms, a population is generated and submitted to the selection and recombination genetic operators (i.e., Crossover). These operators evaluate each individual, i.e., they use a quality feature of each individual as a solution to the problem. As a result, a process of natural evolution of the individuals in the created population is generated, which will consequently generate an individual with features of a good solution to the addressed problem. A flowchart describing the classical GA is shown in Figure 7 [183].

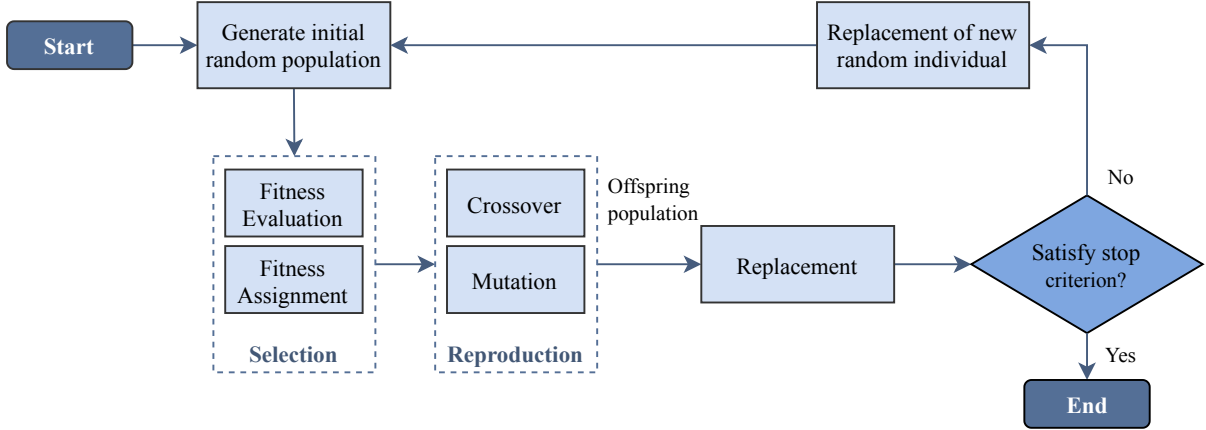

**Figure 7.** Flowchart of a classical Genetic Algorithm (GA).

Particle Swarm Optimization is a population-based stochastic optimization technique pioneered by Eberhart R. and Kennedy J. [184]. PSO is an AI technique that seeks to imitate the social behavior of animals such as fish and birds that live in colonies. The algorithm is initialized with an initial population *candidate* to solve the problem, called *particles* [185,186]. Similar to the GA technique, the PSO initializes a swarm with a quantity of particles $i$, and each particle has a dimension $d$ representing a possible solution to the problem. This occurs in such a way that all the elements of the swarm are within the pre-established range $[x_{min}, x_{max}]$, in the same way as the best evaluation solution (global evaluation) that should guide the hyperspace search for the sub-optimal solution, i.e., solutions that have approximate values to the optimum of the function. The best individual values for each particle are stored and, therefore, the best one estimated will represent a new optimal assessment if it overlaps with that established in the previous iteration. In this way, each particle has its own velocity, which will be updated along the iterations according to the best individual values and the global value of the swarm to then update the value of each particle, as depicted in Equations (6) and (7) [185].

$$v_i^{t+1} = w * v_i^t + c_1 * r_1(p_i - x_i^t) + c_2 * r_2(g - x_i^t) \qquad (6)$$

$$x_i^{t+1} = x_i^t + v_i^{t+1} \qquad (7)$$

Whereas, $v_i^t$ and $v_i^{t+1}$ represent the velocity vector of the particles of position $v_i^{t+1}$ respectively at iteration $t$ and $t+1$, $w$ defines the coefficient of inertia, $c_1$ and $c_2$ are the positive constants, $r_1$ and $r_2$ are the arbitrary values defined in the interval [0,1], while $p_i$ and $g$ represent, respectively, the vectors of the best solution for position $i$ and the best global solution, and finally, $x_i^t$ and $x_i^{t+1}$ represent the particle vector in position $i$ of the swarm, respectively, at iterations $t$ and $t+1$.

A flowchart describing a typical PSO algorithm is illustrated in Figure 8.

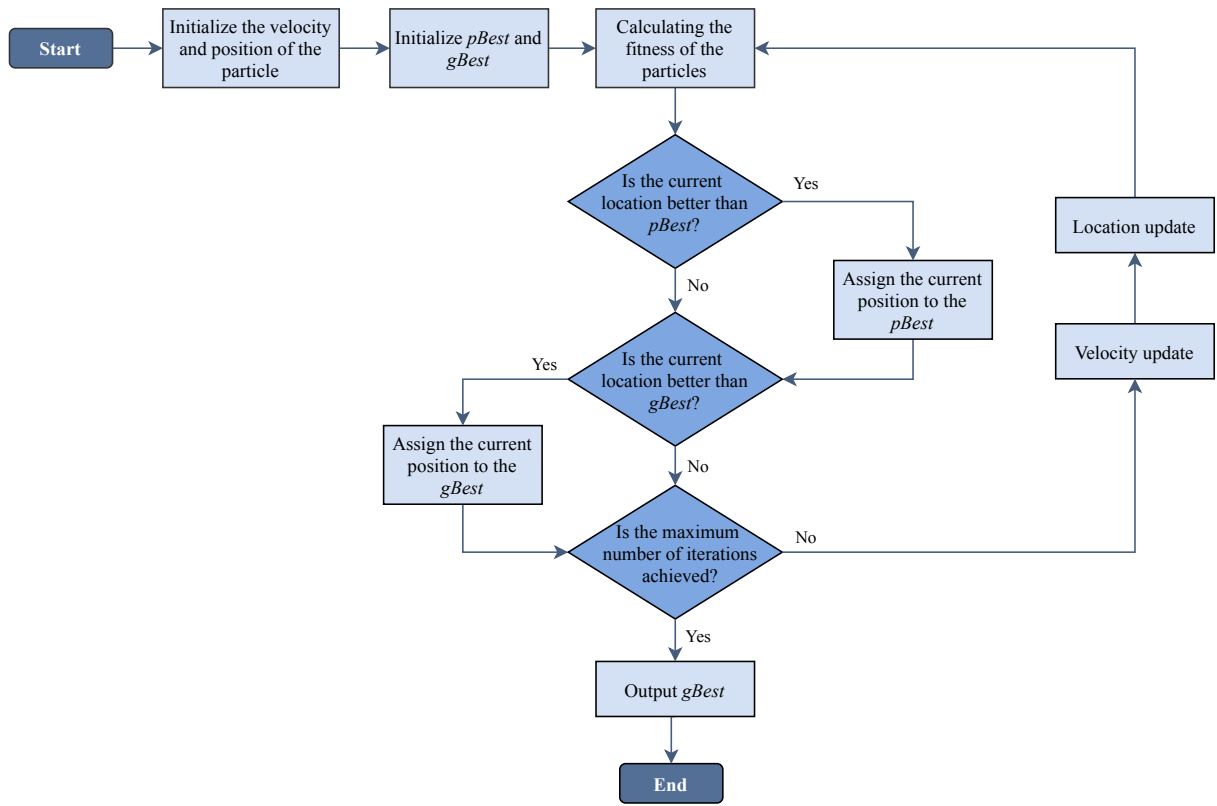

**Figure 8.** Flowchart of a classical particle swarm optimization (PSO) algorithm.

At this level, we give a summary table (cf. Table 11) of the optimization methods and their characteristics.

**Table 11.** Table of the considered optimization methods and their characteristics.

| METHOD | ADVANTAGES | LIMITATIONS |
|---|---|---|
| **GENETIC ALGORITHM [183]** | ● Robust, able to manage several parameters.<br>● Adaptability responsiveness and consideration of the environment.<br>● Allows to process large search spaces (many solutions, no exhaustive browsing envisaged).<br>● Relativity of the quality of the solution according to the degree of precision required | ● Requires more computations than other meta-heuristic algorithms (especially the evaluation function, which makes it very consuming of time).<br>● Parameters difficult to set (population size, % mutation).<br>● Choice of the evaluation function is tricky.<br>● The solution found is not guaranteed to be the best, but just an approximation of the optimal solution.<br>● Problems with local optimums |
| **PARTICLE SWARM OPTIMIZATION [184]** | ● Robust in solving optimization problems.<br>● Simplicity of implementation.<br>● Very short calculation time.<br>● Inexpensive neighborhood management.<br>● Simple parameterization.<br>● Efficient to solve problems which require accurate mathematical models. | ● The wrong choice of parameters can have an effect on both the algorithm's operation and the resulting solution.<br>● Cannot work out with the problems of non-coordinate and scattering systems.<br>● Possible difficulties in defining initial parameters. |

## 4. Theoretical Analysis of the AI Applied for Building Control

Improving energy efficiency and maintaining indoor comfort conditions, while taking into account user preferences, have led researchers to develop intelligent Building Energy Management Systems (iBEMS), primarily for large-scale buildings such as hotels, office and commercial buildings, among others. The iBEMS are developed to be used in a wide array of applications. Such solutions are designed to track and manage the building's microclimate and to reduce energy use and operating costs. The literature includes a significant number of works on the application of AI techniques to iBEMS. The results are more persuasive than those of conventional control systems.

General advances in the development of automated control systems are the need for a mathematical model for building operation, which is a drawback of applying traditional control systems in buildings. Through incorporating high-level variables that describe comfort into smart controllers, comfort could be managed without having to control lower-level variables such as temperature, humidity and air-speed. The consumer starts to get involved in specifying the ideal comfort, in these systems. Hence, through this section, the reviewed 125 publications in which AI-assisted tools were deployed and summarized in Tables 3-9 are therefore extensively examined. In the first place, the case studies are discussed on the basis of the most selected inputs and their associated outputs used by the implemented models, in the second place, the control performance of the AI techniques used for energy saving and thermal comfort optimization are quantified and, finally, the thermal comfort measurement methods are characterized and classified according to the AI tools used. In this regard, Figure 9 presents the block diagram of a typical structure for AI-assisted building control resulting from the reviewed articles.

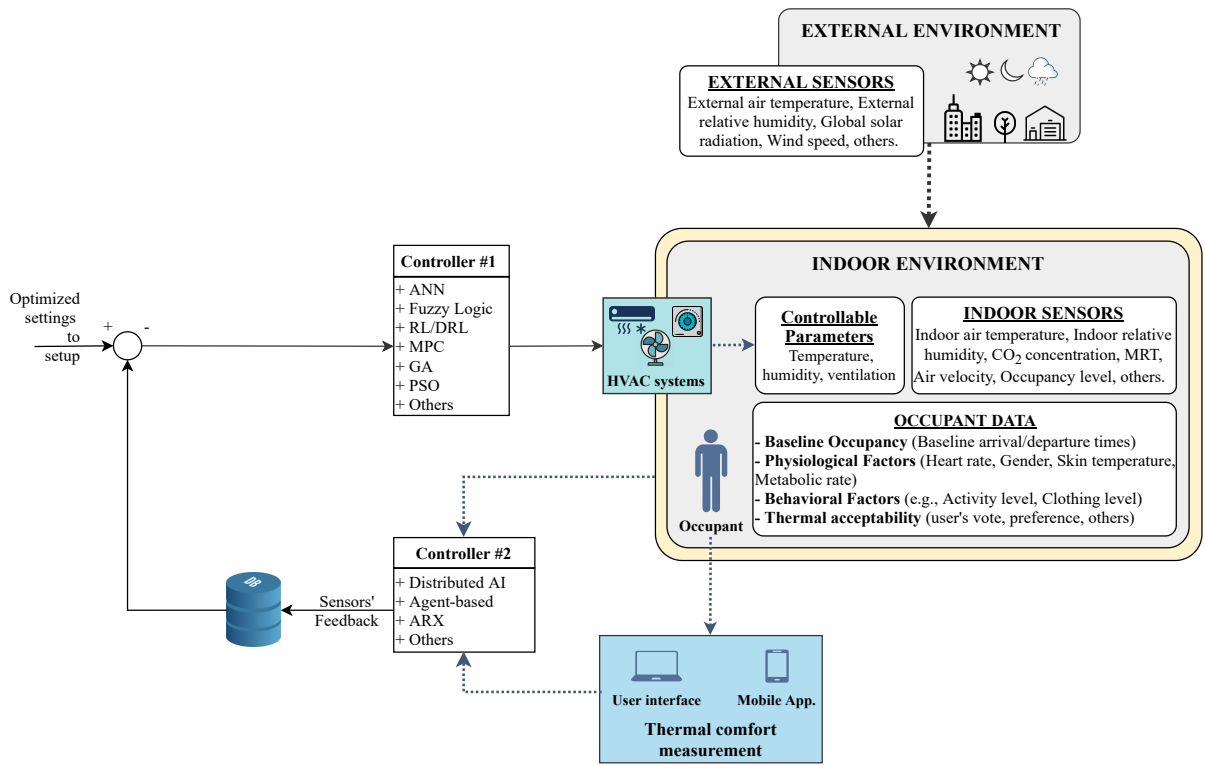

**Figure 9.** Block-diagram of the AI-assisted for HVAC and thermal comfort controls in buildings.

As a result, from the previous section, the core of AI-assisted building control is the neural networks and fuzzy logic, which are illustrated is Controller #1 in Figure 9. Moreover, the key feature of this system control is sensor feedback. The multiple sensors collect various types of variables (e.g., environmental, personal, among others), saved in a database and used by the system to make intelligent decisions. It is worth noting that the AI-assisted tools are also applied on sensors part for more intelligent control, which is illustrated as Controller #2 in Figure 9. In addition, the most adopted functions are the optimized setting and predictive control functions, as shown in the Figure 9. The optimized setting is mainly performed by GA or PSO, as shown by the previous section.

## 4.1. Study Cases: Inputs and Outputs

In the context of thermal comfort and energy-saving control systems, the inputs and outputs considered to generate AI-based models are closely linked. Concerning the inputs, they are mainly associated with comfort conditions and design variables as well as other indicators that could be useful for such control systems. The inputs are therefore divided into seven groups: heating, ventilation and air conditioning (HVAC) systems, indoor and outdoor climatic conditions, occupant-related variables, building geometry and components, among others.

Regarding the outputs, they can be categorized into four major groups: (i) **Comfort indices** (PMV, PPD, and others), (ii) **Microclimate indicators** (temperature, $CO_2$ concentration), (iii) **Energy/Load** (HVAC, cooling/heating, cool/hot water), and (iv) **Others** (including cost and time efficiency). In this regard, the relationship between the selected inputs to describe specific outputs is illustrated in Figure 10.

| Inputs | | | Comfort Index | | | Microclimate Indicators | | Energy/Load | | | | Others | |
|---|---|---|---|---|---|---|---|---|---|---|---|---|---|
| | | | PMV | PPD | Others | Temperature | $CO_2$ concentration | HVAC | Cooling/Heating | Cool/Hot Water | Others | Cost | Time Efficiency |
| Space Control Devices | | Component | 4 | 1 | 3 | 8 | | 6 | 4 | 1 | 7 | 1 | 2 |
| | | Component Efficiency | 9 | 1 | 1 | 5 | | 9 | 4 | 1 | 5 | 2 | 2 |
| | | Set-point temperature | 8 | 2 | 5 | 6 | | 20 | 4 | 3 | 8 | 3 | 3 |
| | | Others | 4 | | 3 | 6 | 1 | 7 | 2 | 2 | 3 | 2 | 1 |
| Climatic Conditions | Indoor | Air temperature | 29 | 4 | 16 | 29 | 2 | 45 | 12 | 3 | 15 | 7 | 3 |
| | | Relative humidity | 21 | 3 | 10 | 12 | 1 | 26 | 6 | | 9 | 3 | 1 |
| | | Air Velocity | 13 | 2 | 4 | 2 | 1 | 11 | 4 | | 4 | 1 | |
| | | $CO_2$ concentration | 10 | 1 | 8 | 12 | 3 | 21 | 1 | | 7 | 1 | 1 |
| | | MRT | 9 | 1 | 2 | 1 | | 5 | 3 | | 3 | 1 | |
| | | Others | 13 | 2 | 9 | 15 | 1 | 23 | 8 | | 8 | 4 | 2 |
| | Outdoor | Air temperature | 16 | 4 | 8 | 17 | 2 | 24 | 9 | 4 | 9 | 1 | 2 |
| | | Relative humidity | 4 | 3 | 3 | 5 | | 7 | 2 | 1 | 5 | | |
| | | Air Velocity | 2 | 1 | | 4 | | 1 | 3 | 1 | 2 | | |
| | | $CO_2$ concentration | 1 | | 1 | 1 | | 2 | | | 1 | | |
| | | Solar radiation | 6 | 3 | 1 | 6 | | 6 | 4 | 2 | 3 | 2 | |
| | | Others | 2 | 1 | | 6 | | 6 | 3 | 1 | | | |
| Occupant-related variables | | Clothing level | 10 | 1 | 3 | 1 | | 4 | 4 | | 5 | | 1 |
| | | Activity level | 9 | 1 | 2 | | | 4 | 3 | | 3 | | |
| | | Comfort Information | 15 | 3 | 8 | 1 | 4 | 14 | 5 | | 6 | 1 | |
| | | Preference | 6 | 2 | 10 | 16 | 2 | 21 | 5 | 1 | 7 | 3 | |
| | | Others | 2 | | 4 | 1 | | 3 | 1 | | 2 | | |
| Building Component | | Window | 5 | | 3 | 2 | 1 | 4 | 3 | | 1 | 1 | |
| | | Wall | 3 | 1 | 2 | 2 | | 1 | 7 | | | | |
| | | Roof | 1 | | 2 | | | 1 | 2 | | | | |
| | | Floor | 2 | | | 2 | | 1 | 3 | | | | |
| | | Door | 2 | | | 1 | | | 2 | | 1 | | |
| | | Envelope | 1 | | 1 | 1 | | | 2 | | 1 | | |
| | | Others | 1 | | 1 | 3 | | | 5 | | | | |
| Building Property | | Location | 1 | | 2 | | | 1 | 1 | | 1 | | |
| | | Geometry | 3 | 2 | 2 | 1 | | 3 | 4 | | | | |
| | | Configuration | 1 | | 2 | | | 1 | 1 | | 1 | | 1 |
| | | Others | 3 | 1 | 1 | | | 3 | 1 | | | | |
| Others | | Energy Information | 1 | | 1 | 1 | | 3 | | | 1 | 1 | |
| | | Power consumption | 1 | | | 3 | | 3 | | | 1 | | 2 |
| | | Time/Date/Hours | 3 | 2 | | 7 | | 4 | 5 | 1 | 4 | | 1 |
| | | Occupancy | 7 | 3 | 7 | 16 | | 19 | 5 | 2 | 9 | 4 | 2 |
| | | Disturbances | | | | 2 | | 2 | | | | | |

**Figure 10.** Heat-map of the number of times of using a given input and the corresponding output of the AI-based models.

It is worth noting that the numbers in the heat-map represent the times when a specific input is used to approximate certain outputs. For example, the air temperature was used 45 times to estimate the HVAC load.

Such information is useful by highlighting the most influential and selected variables used by AI-based models as inputs to building installations/system targets (as shown in Section 4.2).

## *4.2. Study Cases: Energy Control*

We are now turning our attention to how AI techniques have been applied to improve energy efficiency and thermal comfort. Figure 11 shows the diversity between the building control system component/installations adopted in the reviewed works, responsible for ensuring comfort conditions and suitable indoor air quality in indoor settings. It is apparent that AI techniques are relevant for implementation in different parts of the building control systems.

It is worth noting that among the solutions proposed for improving energy efficiency and indoor comfort conditions, we can distinguish the "passive" solutions, which consist of reducing the energy consumption of equipment and materials through better intrinsic performance, such as the building's architecture, thermal insulation, airtightness, hot or cold water, household appliances such as lighting. Passive solutions are a key sustainable way for energy efficiency and comfort control. However, the drawback of most of these solutions are that they depend on the local weather and the outdoor air quality. Besides, of course, "active" solutions, which aim at using "just the right amount of energy" through active management of equipment, such as intelligent systems. These systems make it possible to measure, control and regulate the energy consumption of buildings (sensors for temperature, presence for lighting, $CO_2$ emissions for ventilation, among others), and thus avoid unnecessary consumption.

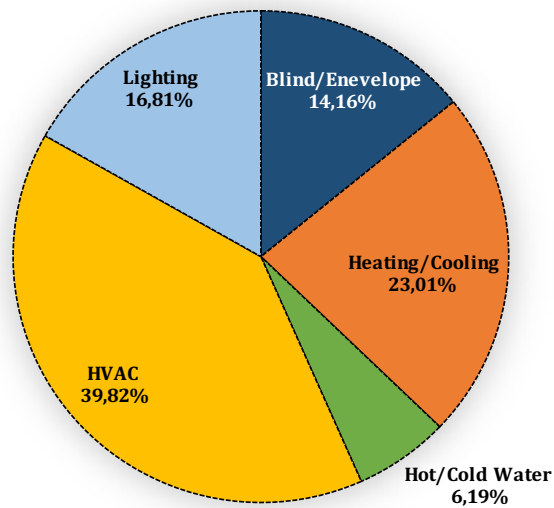

**Figure 11.** The relative distribution of AI tools used in building control systems among the reviewed works.

In addition, the proposed AI-assisted methods applied to the building control systems, in most of the revised papers, are qualified as predictive, since the strategies for optimal loads operation (heating and cooling) are implemented while respecting the maximum authorized power rate and thermal comfort. The implementation of the optimization functions in typical loads such as ACMV is made through modeling the energy-consuming units, which consist of the Liquid Dehumidification Unit (LDU), the Air-Handling Unit (AHU), and the Water Chiller Unit (WCU). In this context, such an example can be found in [69], in which the authors have used an optimization algorithm based on a Single Layered Feedforward Neural Network (SLFF-ANN) for modeling energy consumption in an ACMV system. In their case study, they have considered the main energy-consuming component in the ACMV system as model inputs, i.e., the energy consumption of ACMV was modeled as a function of air temperature as an indicator for thermal environment, supply-air fan (as a primary part of AHU), and water-pump and compressor in cooling system. As a result, the model afforded a saving of 36.5% of on average.

Therefore, we can define three kinds of methods of optimal load management: (1) Heating and cooling management methods based on minimizing consumption peaks; (2) Methods for heating management that are based on maintaining a comfortable indoor temperature; and (3) Methods for heating management based on minimizing consumption costs.

### *4.3. Study Cases: Thermal Comfort Measurement*

Thermal comfort assessment approaches can be categorized into two groups, according to the reviewed works: **General Comfort Model (GCM)** and **Individual Comfort Model (ICM)**.

#### *4.3.1. General Comfort Model*

The conventional approaches focused on the thermal equilibrium between man and his surroundings [187] (cf. Equation (1)) allow the development of internationally recognized environmental indices, such as the Fanger PMV-PPD model, considered to be a GCM. In addition, this model was statistically based on experimental studies involving 1,300 subjects in climatic chambers. Its main limitation lies in the fact that the PMV index estimates the average comfort level of the subjects, which was also determined under homogenous and stationary conditions, representing theoretical conditions rarely encountered in actual buildings.

Personal models based on the PMV model, such as the Predicted Personal Vote (PPV) model, may be considered GCM, defined as the PMV transform affine: $ppv = f_{ppv}(pmv)$ [145]. The idea behind PPV is to assess the level of comfort within a single worker within a workplace. The inverse-PMV model may also be considered as GCM, used to calculate thermal comfort temperatures based on the desired target PMV and measured assessed air-speed and humidity [54]. Apart from the Comfort Time Ratio (CTR) index, also considered as GCM, which is based on Szokolay's theory and assesses the annual indoor thermal comfort for residential buildings [123]. These comfort indices are used as inputs to the temperature control system to adjust the comfort level of the building.

Furthermore, in commercial applications, models such as conventional methods (i.e., fixed temperature settings that can be adjusted for complaints and predefined indoor conditions in accordance with standards and legislation that can be considered GCM) are adopted in order to identify comfort ranges. Among these standards: CIBSE which defines the comfort levels in office buildings between 21ºC and 23ºC [157], OSHA guidelines specifying the comfort zone between 20ºC and 24.2ºC [142], ASHRAE 55 which limits indoor temperatures between 21.5ºC and 24ºC during occupancy hours [147], comfort margins based on Royal Decree 1826/2009 by setting indoor temperatures between 21ºC and 26ºC.

#### *4.3.2. Individual Comfort Model*

Although the ASHRAE 55 is considered as a global standard for assessing thermal comfort in buildings, there are a variety of drawbacks and concerns: the main issue is that the comfort models existing in this standard are considered valid for anyone (i.e., even though the models indicate thermal zones for 80% or 90% of thermal comfort/acceptability; they do not discriminate which users' group would not be in comfort or would not be accepting the thermal conditions). However, different groups of people can have different thermal perceptions. Individual Comfort Models (ICMs) can therefore provide individual treatment that can give better satisfaction for occupants within a given environment. ICM is a recent paradigm predicting individual-level thermal comfort and is typically based on data-driven learning algorithms. A Bayesian comfort model (BCM) was developed by combining a human-body-centered approach of static models with an external environment-based technique of adaptive models [35]. A data-driven thermal comfort model was also created by learning subjective feedback from the occupants in real-time through the application of the smartphone/server (OMG) and objective thermal information [164]. In addition to the personal thermal sensation model (for MET, the personal activity of the occupant) and the group-of-people-based thermal sensation model (for MET, the average activity group of people) generated by the SVM algorithm for assessing the occupants' thermal sensation [163].

In addition, other works suggested personalized models by investigating the "human-in-the-loop" approach that allows HVAC to be adapted to user preferences. Personalized comfort profiles established on a participatory sensing approach by embracing a thermal perception index (TPI) scale (slider values) that shows thermal preferences of votes ranging from -5 to +5 [158]. The Degree of Individual Dissatisfaction (DID) index was defined as the function of the user's vote and, depending on the ambient temperature, the desired individual temperature ($T_0$) and the individual temperature tolerance ($\Delta T$) [147]. A comfort-driven framework based on the scale of user preferences using the Thermal State Index (TSI) (*Cool-Discomfort/Comfort/Warm-Discomfort*) is provided in [69].

It is worth noting that the discussed indices consider short-term and long-term evaluation of thermal comfort conditions in buildings. In short-term indices, different classifications have been proposed to categorize indices in identical families. The most adopted/useful classification is the one proposed by MacPherson [188], in which these indices have been classified into: (1) rational indices; (2) empirical indices; and (3) direct indices. However, the drawback of these indices is they are limited in a certain time or position in a given space. Whereas, the long-term indices for a global evaluation of thermal comfort in a building over certain periods of time and considering the different building zones. In their complete study, Carlucci S. and Pagliano L. [189], in which they have divided these indices into four groups: (1) percentage indices, (2) cumulative indices; (3) risk indices; and (4) average indices. It has been noted that the most common approach for thermal comfort evaluation is showing discomfort rather than comfort level in long-term.

## 5. Trend Analysis and Discussions

The graphic detail of the studies considered in this review is displayed in Figures 12 to 20. Among the various AI techniques, neural networks are the most popular approach adopted by researchers among the research papers in our study (cf. Figure 12). Fuzzy logic is also widely used for energy saving and thermal comfort improvement due to its suitability to imitate human behavior and enable linguistic descriptions of thermal comfort sensation. Hybrid methods were also preferred by combining two different techniques (e.g., FL and ANN; ANN and GA/or PSO; FL and GA/or PSO). In most cases, GA and PSO have been introduced to provide optimal solutions to building optimization problems. Although there are fewer works using DAI and MAS, they have been used in complex control systems by incorporating a set of controllers instead of a single controller system.

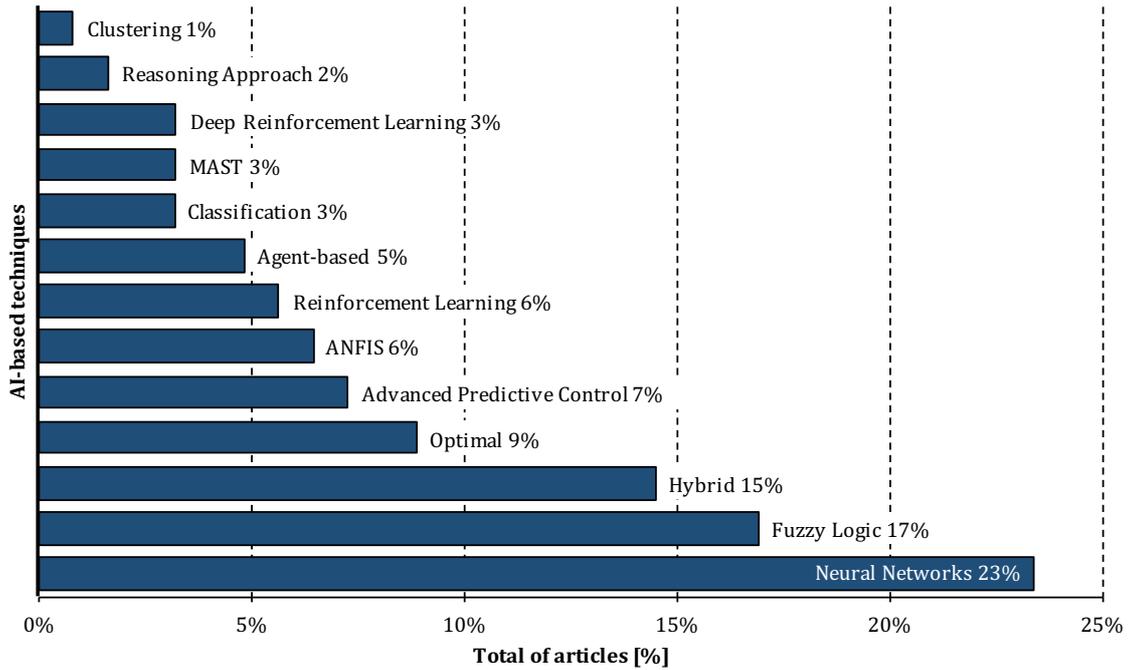

**Figure 12.** The frequency of use of the AI-based tools extracted from the reviewed publications related to the building control.

Statistical results have shown that, from 1993 to 2020, the average energy savings in buildings by applying AI/ML techniques reached up to 31% (cf. Figure 13). Maximum energy savings (~90%) were achieved by applying a Bayesian network-based model to determine acceptable thermal conditions, with the aim of estimating employee mental performance under different scenarios [156]. Moreover, advanced predictive models have shown promising results in the reduction of energy consumption, for example in [142], in which a learning-based model predictive control (LBMPC) was applied to improve energy efficiency (~50% reduction in energy consumption) in an HVAC-Testbed platform located in a room laboratory. Along the same line, a model-based predictive control of neural rule base function (RBF) networks was implemented and identified through the MOGA technique for HVAC control in large public buildings [144]. The model has shown significant results in terms of energy savings, by allowing to save more than 50% of energy while providing a good coverage of the thermal sensation scale. In [145], the authors proposed a smart personalized office thermal control (SPOT+) system using the LBMPC and kNN algorithm used to estimate room occupancy and optimum room temperature within the office building. Based on the predictive model, SPOT+ identified a control schedule that allowed to save about 60% of energy use and optimized thermal comfort. Shaikh et al. [167] recently proposed an agent-based control system using an evolutionary multi-objective genetic algorithm (MOGA) for energy and comfort optimization. The developed optimizer has saved up to 67% of energy consumption in addition to about 99.73% of comfort improvement.

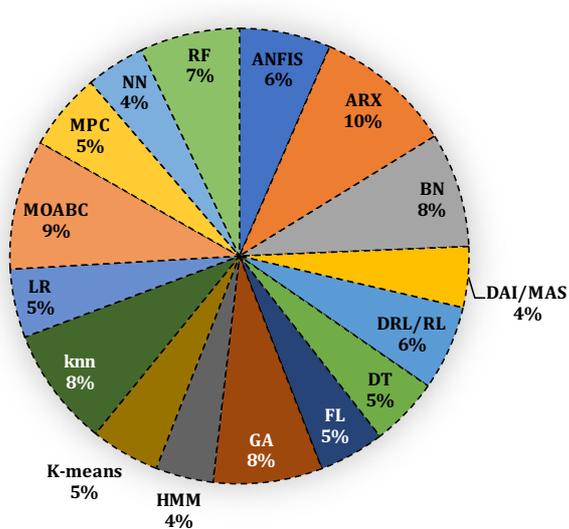
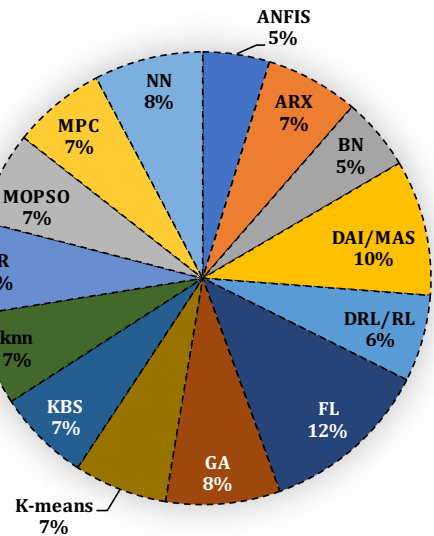

**Figure 13.** Average of key results: Implications of AI/ML techniques on Energy Savings (Results are given in average).

**Figure 14.** Average of key results: Implications of AI/ML techniques on Comfort Level (Results are given in average).

Furthermore, the average comfort level improvement using AI/ML-based techniques was around 50%, while the maximum comfort level reached 100% through the use of neural networks [36,46,48,50,51,71], DAI and MAS [99,106,111], as well as GA [119] (cf. Figure 14). Such comfort improvement was demonstrated in [32] by the development of an Intelligent Comfort Control System (ICCS) incorporating human learning with techniques for reduced energy usage in HVAC systems. The system enabled a higher level of comfort (100%) by keeping the PMV within the comfort zone while saving energy. The GA method has also shown its potential by achieving better energy efficiency and comfort criteria (100%) for a heating and cooling system, by lowering preliminary and operating costs by up to 35% and decreasing the comfort cost by 45% [119]. Another objective was targeted by only 6.7% of the reviewed works (including thermal comfort and energy savings improvement), which are cost-effective, including operating costs, energy savings and electricity costs, as well as comfort costs. The average cost reduction using AI/ML methods was up to 34%, with a maximum of 58% of energy saving costs [160] (cf. Figure 15).

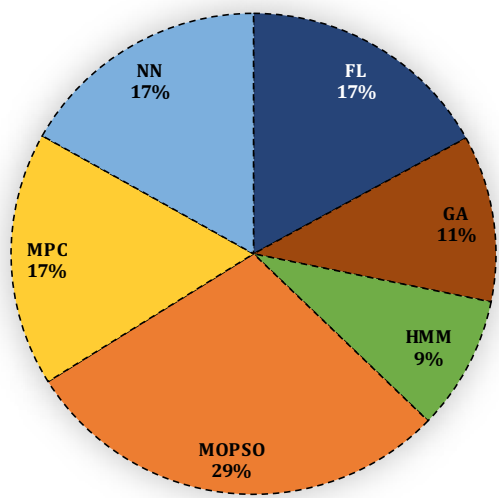

**Figure 15.** Average of key results: Implications of AI/ML techniques on Cost (Results are given in average).

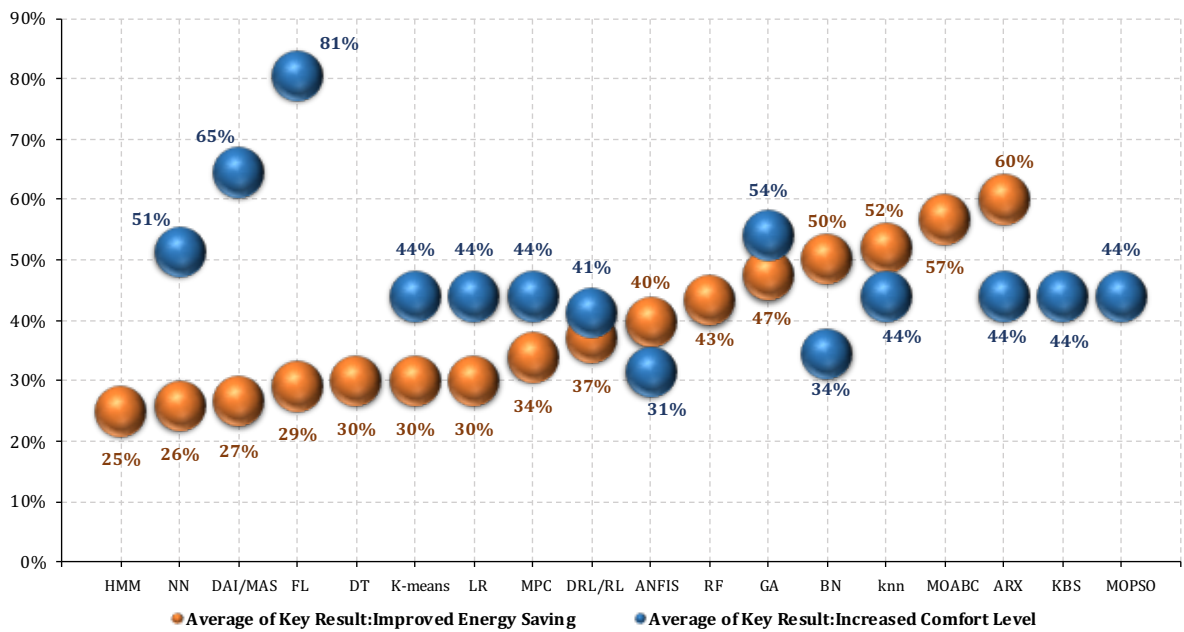

**Figure 16.** Implications of AI/ML techniques on both key results: energy savings and thermal comfort level.

Intelligent control requires no mathematical model for the configuration of the controller and is based solely on the human perception of thermal comfort. Furthermore, in thermal comfort control, which is based on the set temperature values, there is no need to keep the indoor temperature at a fixed value, although a range of these quantities is sufficient to create a situation of comfort (cf. Figure 17). Reducing energy demand, and therefore energy costs, while maintaining thermal comfort indices within the permissible range, is a goal to be achieved in selecting the appropriate control technique. For example, fuzzy controllers have shown significant results in thermal building control, as they can properly imitate the behavior of building users and create linguistic descriptions of thermal comfort sensation which estimate PMV model calculations to facilitate system control (cf. Figures 18 & 19).

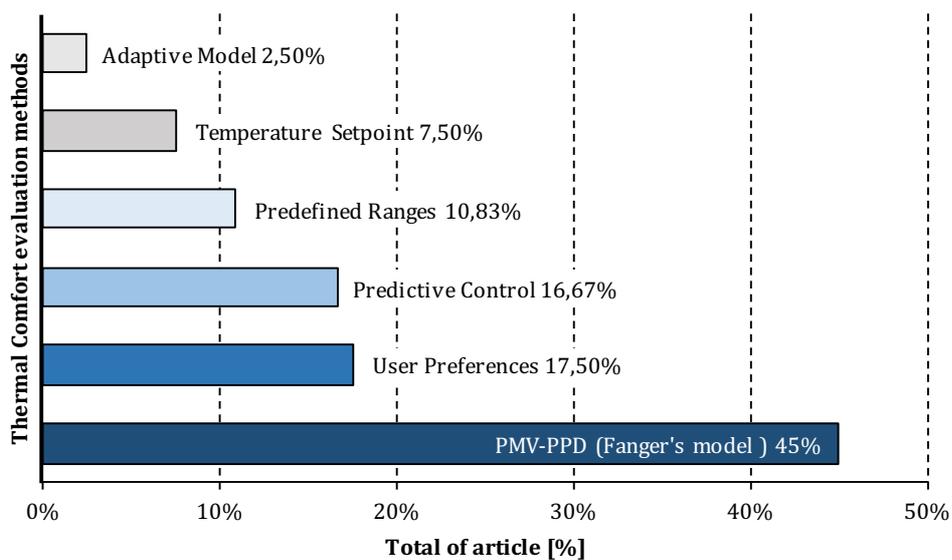

**Figure 17.** The percentage of methods used for assessing thermal comfort in the reviewed works.

In this way, the fuzzy control scheme proposed in [190] is characterized by the explicit consideration in the control law of a range of permissible values for indoor ambient temperature rather than a fixed value. Recently, several studies have been directed towards suggesting personalized models dealing with both thermal comfort and energy savings, by investigating a "human-in-the-loop" approach that allows HVAC to be adapted to the individual preferences of each user. In [35], the authors proposed a Bayesian Comfort Model (BCM) that showed significant results by giving 13.2% to 25.8% accuracy of the user's preference estimate compared to the existing method, and can save up to 13.5% of energy consumption by minimizing 24.8% of discomfort. While researchers have also shown the potential of using the smartphone/server application to generate a data-driven thermal comfort model through training, in real time, the subjective feedback from occupants in real-time [164]. The results showed that the learned settings had potential energy efficiency, while meeting standard expectations of thermal comfort, i.e., ≥ 80% of thermal satisfaction.

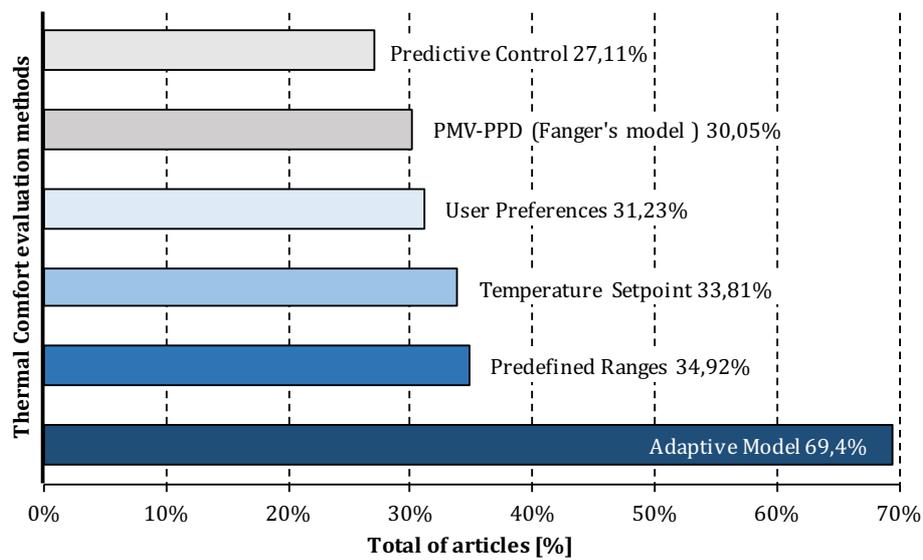

**Figure 18.** Average key result: Implications of Thermal Comfort-based model/method on energy saving.

These intelligent tools can also be used to improve existing conventional controllers, as can be found in [47], where three AI-based thermal control logics have been used to improve existing conventional controllers: (i) Fuzzy-based control; (ii) ANFIS-based control; and (iii) ANN-based control. The efficiency of each approach is examined in a two-story residential building. It is concluded that ANFIS- and ANN-based control methods are potentially better than conventional methods for maintaining indoor thermal comfort conditions (~98% in winter and 100% in summer) by setting up comfort bands for each season (20–23ºC in winter/23–26ºC in summer). However, none of the three techniques showed significantly more energy savings than the others.

Moreover, a hierarchical multi-agent system for multi-objective monitoring and maintenance of intelligent building applications was handled in [161]. The developed control system used stochastic optimization using a hybrid MOGA and saved 31.6% of energy, while the comfort index (based on user preferences) was improved by about 71.8%, compared to traditional optimization techniques. The work of Davidsson P. and Boman M. [99] is another contribution based on the MAS approach in which a decentralized framework for managing and monitoring an office building has been established. In this work, the proposed system facilitated the optimization of energy use (up to 40% of average energy savings compared to baseline) in three services: lighting, heating and ventilation, while ensuring 100% thermal satisfaction of users.

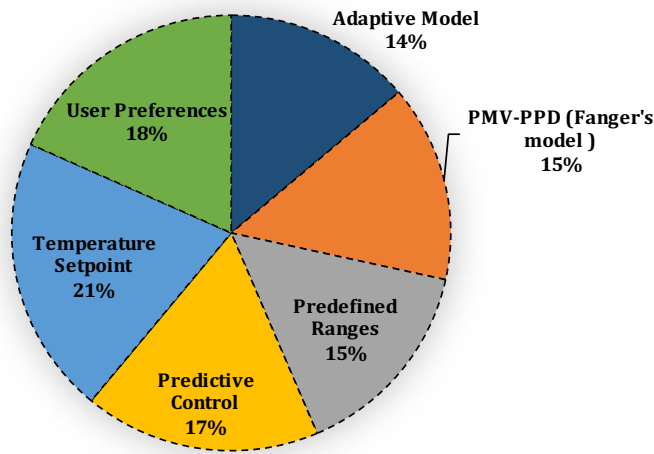

**Figure 19.** Average key result: Implications of Thermal comfort-based Model/Method on improved comfort level.

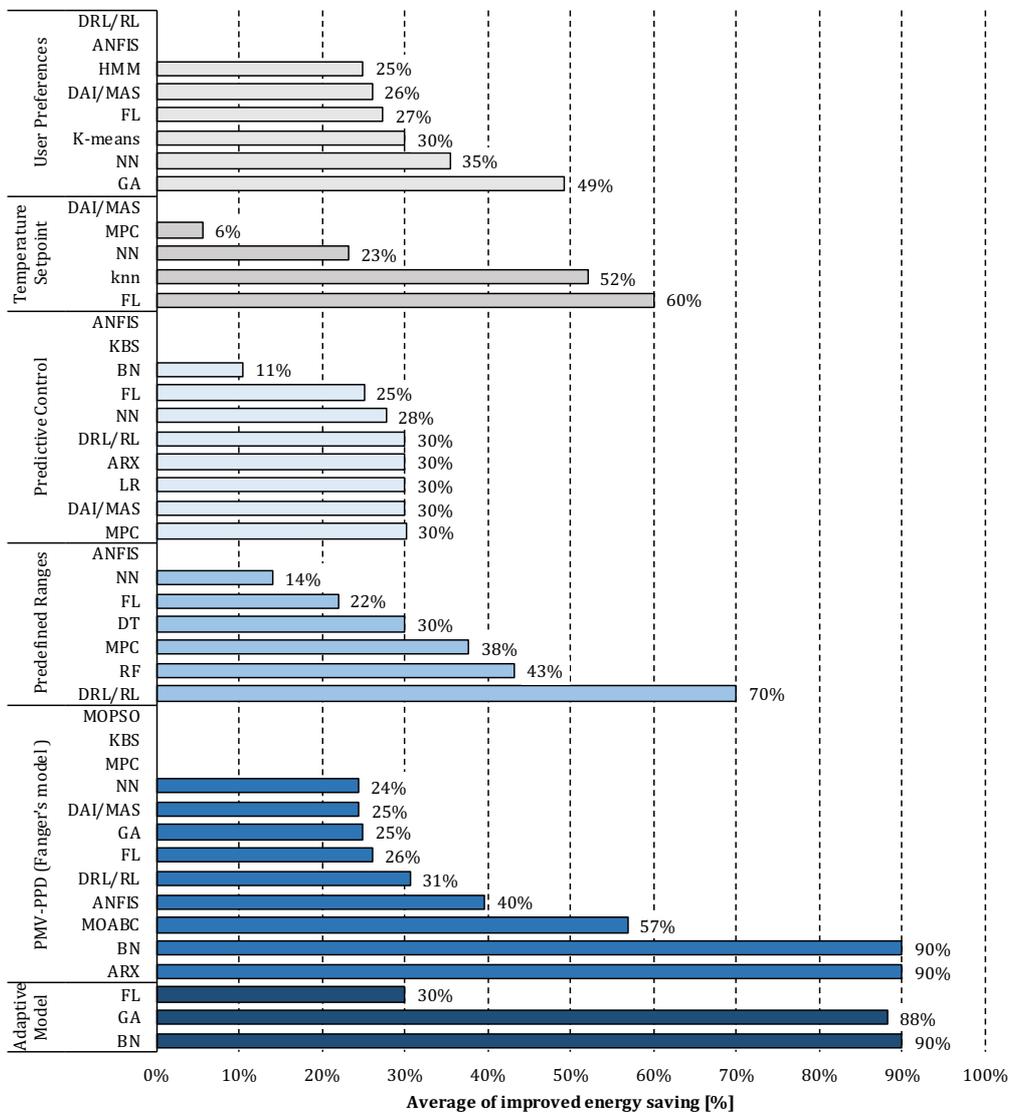

**Figure 20.** Implications of AI/ML techniques and Thermal Comfort-based model/method on energy saving.

# 6. Open Challenges and Future Research Directions

The ability of sustainable buildings to provide thermal comfort to their occupants while improving energy productivity is still an open research problem that presents numerous research challenges. Using AI to improve energy productivity while meeting occupant's comfort presents various research challenges and opens future research directions. These challenges and future research directions are presented in an application-oriented organization in this section.

## *6.1. Smart and connected buildings*

As buildings are getting smarter through AI, networking these buildings using the Internet of Things (IoT) and will open up a window of research opportunities to address greater challenges in operation, design, and user experience in building. Connected building will also facilitate developing smart and sustainable solutions to challenges in smart cities and smart grid. In smart cities, buildings constitute a basic block of its structural formation that is used to deliver key smart city functionalities [191]. Connecting these buildings would further enhance life quality and sustainability of citizens. From an energy perspective, connected buildings will play a key role in improving energy efficiency and resilience through community Microgrids and district energy concepts [192]. Community Microgrids and district energy systems are being developed for campus environments to improve energy resilience through Combined Heat and Power (CHP), renewable energy and storage. They will have a more significant impact on smart grid peak power management through demand response, compared to single building management. This is because connected buildings will have the ability to interact with neighboring buildings to create new opportunities for energy management in the smart grid application [193]. Connected buildings will promote sustainability by enabling the implementation of a net-zero community in terms of energy and carbon footprint. Sustainability opens up challenges in both long-term planning and building day-to-day operations, while taking into account the community as a whole. Buildings will connect with their occupants and sensors as well as their surroundings. Connected buildings will allow us to develop solutions to research challenges from different perspectives using network-centric technologies.

## *6.2. Comfort modeling and dynamic temperature set-point adjustment*

The assessment of thermal comfort in indoor environments is essential for the development of environmental and energy efficiency regulations in buildings, which are primarily accomplished in compliance with the ASHRAE 55 standard [194]. However, the analytical models suggested by this norm involve the knowledge of several parameters, the calculation of which is complex and hence can be skewed (because they are based on a set of assumptions leading to the concept of approximate comfort). Research has demonstrated that, in the real situations of the in-situ studies, the comfort conditions experienced differ according to the studies context, in particular the climate and the way in which the thermal environments within the buildings are managed. In non-air-conditioned (i.e., naturally ventilated) buildings where thermal conditions differ with seasonal variations, the variety of comfort is broader than that offered by the PMV, which appears to overestimate the warmth sensation in this kind of building during the summer period, particularly in hot climates. In a study of various types of buildings situated in three different climatic cities in Australia (air-conditioned and non-air-conditioned), de Dear R. [195] found that PMV was unable to estimate conditions of thermal neutrality in both types of buildings, while linear regression equations better predict thermal neutrality as determined by previous surveys. The same observation shall apply to Busch J. [196], who conducted a study in office buildings in Thailand, found thermal acceptability to be between 22ºC and 30.5ºC during the summer period (4ºC above the norm).

The discrepancies observed between the comfort conditions encountered in situ and those predicted (e.g., PMV index) are not only due to measurement errors and uncertainties in metabolism and clothing insulation estimates, but also reflect the dynamic interaction of subjects and their environment. Nicol J. and Humphreys M. [197,198] postulated the influence of certain psychological and sociological factors. The diversity and variability of acceptable thermal conditions in the in-situ studies suggest that subjects are adapted to the thermal environments in their usual living areas. Such gaps have led to a different approach to thermal comfort, while considering a man as an active element that reacts to variations in his environment in order to guarantee his comfort. This is achieved by highlighting the inability of rational models, developed in climate chambers under stationary and homogeneous conditions, to predict thermal comfort in real-life situations, influenced by the dynamics of multidisciplinary interactions between the subject and his environment.

In addition, a number of technological solutions, such as Wireless Sensor Networks (WSN) and IoT-based solutions, have been proposed in order to adjust indoor climate conditions. However, in order to reach a convenient and energy-efficient environment, the occupant is supposed to become an expert in these technologies that can challenge his daily habits. In view of the complexity of these technologies, the user could choose the solution of smart buildings equipped with sensors to adjust everything (temperature, ventilation, window opening/closing) to promote energy savings and comfort. Moreover, we might think that building automation makes it possible to achieve comfort conditions and reduce energy use, but studies have shown that when the user is able to act on his environment, he sets the conditions that allow him to achieve optimum comfort. It is therefore necessary for the occupant to interconnect with the surrounding environment to achieve the expected savings, since the human being is always the ultimate sensitive sensor (i.e., the user is an actor of comfort and can act to meet the conditions that are favorable to the risk of opposing social or technical practices designed to lower energy consumption or enhance comfort).

In their study, Nicol J. and Humphreys M. [198] showed that an individual is more tolerant towards comfort situations if he has the capacity to act (by himself) on the regulation of systems. For example, in buildings where control is centralized, occupants must adapt to a certain temperature that may make them feel uncomfortable. Thus, according to them, when the occupants have access to temperature change control, they find the atmosphere more comfortable.

### *6.3. Human-in-the loop research challenges and directions*

Since people spend a substantial part of their time inside buildings, this has a significant effect on their well-being, comfort, and productivity. Various studies have also shown that at least 5% of people in each group report that they are uncomfortable with the pre-defined conditions of comfort and that, this percentage may increase until exceeding the whole population if the environmental conditions become more adverse [199]. Furthermore, existing thermal comfort standards are valid for anyone (cf. Section 4.3.2), i.e., they do not discriminate thermal variability between occupants. Also, the parameters used by these models cannot be dynamically tracked by the building control systems to change their settings. Dynamic models have been proposed to address this gap in order to provide a satisfactory level of comfort and to ensure the satisfactory quality of indoor environments.

Several studies have recently been directed towards developing individual comfort models based on AI/ML techniques and users' personal characteristics within a given environment [35,137,200–204]. For example, in [200] a decision support framework is proposed for personal thermal comfort prediction – in real time, especially for senior citizens, using environmental, psychological, and physiological features. The obtained results showed a significant improvement in predictive accuracy (76.7%) compared to the conventional Fanger model (35.4%) by including two new factors: age and outdoor temperature that are not considered in the Fanger model. In other personalized models, both thermal comfort and energy savings are addressed through a "human-in-the-loop" approach that allows HVAC to be adapted to user preferences [35]. While Li D. et al. [201] proposed the use of a smartphone application to dynamically track the optimal conditioning mode in a single or multi-occupied area, the reduction in uncomfortable reporting is almost 53.7%.

In addition, creating thermally comfortable and energy-efficient environments is an important yet challenging issue. Particularly with the increasing cost of energy to be taken into account when determining optimum temperature settings in different zones for different groups of building occupants. Also, personalized comfort standards pose contradictory situations in multi-occupancy areas, in which each occupant has his/her own comfort range.

### 6.4. *Context-awareness computing in energy and thermal comfort control challenges and research directions*

Context-Awareness (CA) corresponds to the capability of the system to recognize changes that will have an impact on overall system operation and functionality. These changes include direct system inputs as well as other parameters such as occupant location, device presence, temperature, humidity, sound and lighting. CA is considered as the basis for building intelligence development and is essential for optimizing energy consumption and thus minimizing production. CA promotes the use of electricity only when it is needed and with appropriate levels. For example, heating up a classroom with 10 "present" students should not be the same as 40 students. Besides the presence, the heating level should account for other variables, such as outside ambient temperature, open/closed windows, user preferences, among others. These variables are the *Context* and every Building Energy Management System (BEMS) should be *Aware* of them.

CA relies heavily on sensing the environment, acquiring information about users, and actuating equipment. Sensing and actuation have been facilitated by the advent of Wireless Sensor Networks (WSN) and the Internet of Things (IoT). For example, it is possible to locate and count people using Bluetooth technology on their mobile phones. It is also possible to deploy sensors through IoT sensors (e.g., temperature, $CO_2$, window lock, etc.). In order to deploy CA in buildings, we need to interconnect these IoT devices with the BEMS control plane [205] and have the ability to detect occupant activity and behavior. There are plenty of IoT solutions in the market. However, the decision as to which solution to deploy depends primarily on the existing BEMS and whether it is an open or closed source solution.

To promote *easy-to-deploy* solutions, which are essential from the customer and building operators perspectives, IoT devices should be battery-powered. In addition, these devices should allow mesh connectivity to avoid the need to search for the best location in the room/office next to an electrical socket and with direct connection to BEMS. Besides, by opting for a wireless mesh networking topology, IoT devices will dynamically and seamlessly adapt their routing paths and choose the best route to the control plane [206]. This said, it still comes at a significant price, as this poses another problem to solve: optimizing the *limited battery-life* of the sensors while using low energy harvesting techniques to keep charging the battery.

In this context, a novel routing algorithm Energy Aware Context Recognition Algorithm (EACRA) that dynamically customizes sensors to transmit specific data under specific conditions and at a specific time is still an open research endeavor [207]. Such algorithms should leverage the fact that most sensed data are not that frequently changing (e.g., the temperature in a room would be sensed every 5 minutes) and thus avoiding sending redundant data. By reducing the rate at which the data is sent by the sensor, the battery-life of the sensor is optimized. In addition, control and routing protocols can take advantage of machine learning algorithms to optimize routing decisions. Another research opportunity is how to use energy harvesting techniques inside the building to charge the batteries of these IoT devices and increase the battery life. The modeling of context awareness-based techniques to optimize the way the system interacts intelligently is still an open research question that needs to be addressed. Context awareness is a key component of Information and Communication Technologies (ICT) and promotes smart energy-efficient buildings [205] and enables smart optimization of energy consumption that adapts to user activity and behavior.

### 6.5. *Personal big-data streaming challenges in smart buildings*

Sensing technologies in smart buildings generate massive amounts of heterogeneous personal data. Typically, terabytes of personal streaming data can be generated daily from a variety of sources in smart buildings. The mechanism of long-term storage and pre-processing this data is crucial. A typical batch processing approach does not work with personal streaming data collected from a variety of data sources. Many applications for thermal comfort control need fast response and processing time. Advanced thermal comfort control and monitoring systems are relying on more and more complex models using both real-time and historical data, as well as aggregations and correlations in the range of several years. For these reasons, current research efforts are mainly focusing on the problems and opportunities resulting from personal big data streaming in smart buildings. Researchers are starting to focus on different techniques to store, organize and process these massive datasets using high-performance platforms that provide resources to explore the insights into big data. These datasets also require customized ML techniques, flexible cloud computing and virtualization solutions, distributed computing techniques, and stream processing solutions such as Apache Hadoop and Spark.

### 6.6. *Security, privacy and data sensitivity issues in smart building*

Although smart buildings can buy users and operators of smart buildings significant facility, convenience, and building management efficiency, there are significant security and privacy risks involved. News of embedded sensing, actuating, communication devices having some vulnerability due to some software bug or exploit comes up almost every day. This is in part due to the fact that creating secure software is both hard and time consuming – this ultimately often results in smart equipment manufacturers giving a short drift to security in their bid to expedite development and reduce costs [208]. Since the security of a system is as strong as its weakest point, the resulting risk rises exponentially when we consider the networking of multiple IoT devices within a smart building and across buildings for smart factories or smart cities and the dangers of a cybersecurity attack on such a networked system.

Producing trustworthy smart buildings requires that we give due attention to the five design considerations (safety, resilience, reliability, cybersecurity, and privacy) (cf. Figure 19) [209]. These five design considerations have traditionally been considered in isolation, but such an approach is not sufficient to ensure the overall security of the system. NIST Framework for Cyber-Physical Systems (CPS) [209] recommends that these considerations should be studied in an interdisciplinary holistic manner at various levels of components (be it physical, analog, or at the more abstract cyber level) as illustrated in Figure 19. A more detailed description of the NIST Framework for CPS and related recommendations can be seen in [209]. Future work should focus on how to create efficient smart buildings that are secure at different levels (hardware, network, middleware, as well as at users and application level) [208,210,211].

Apart from the considerable security risks, the use of sensors of different types also raises privacy concerns. The availability of private data generated by smart building sensors opens up individuals to potential harm and exploitation through the use of algorithms that can be used to dig out intimate information that the user may not want to divulge [212]. Strong ethical oversight and regulations, building upon efforts such as EU General Data Protection Regulation (GDPR), are needed to ensure that unscrupulous elements do not use private data without the informed consent of individuals [213].

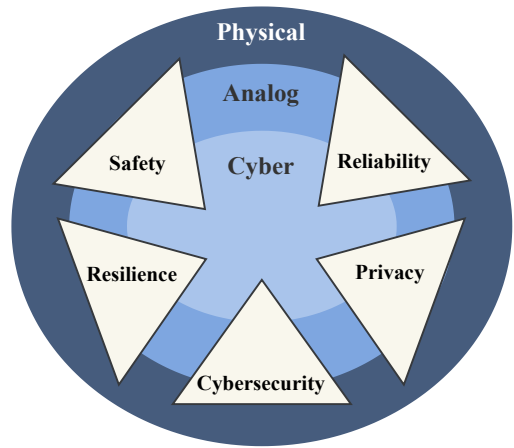

**Figure 21**. The three components of smart building cyber physical systems (physical, analog, and digital) and the five interdisciplinary design considerations (safety, resilience, reliability, cybersecurity, and privacy) that must be addressed for secure smart buildings. (Figure Credit: NIST Framework for CPS [209])

A fundamental open question in this area is how to perform effective privacy-preserving machine learning in smart buildings without sacrificing accuracy. In this context, differential privacy-based techniques appear technically promising but more research is needed to evaluate its suitability in smart building environments [214].

## *6.7. Lack of sufficient amounts of data issues and lack of AI-based modeling in buildings*

Data is an essential part and the fuel needed for any successful AI predictive modeling exercise in thermal comfort prediction and control systems. It's a proven reality that the framework you create is just as good as the data that it's provided. In some thermal comfort services, it's possible that sufficient data is not available due to copyrighted datasets or privacy considerations which are more common in domains with human daily activities in smart building environments. With small amounts of data, and the resulting lack of generalization, AI predictive methods tend to overfit. In some applications, you might have data for training but you have to make sure it is high-quality. The datasets need to be representative and balanced; otherwise, the model will make biased decisions based on the available data. In addition, there are challenges of labeling the massive data collected from smart buildings, many of the energy and thermal comfort control systems utilizing AI techniques are trained using supervised learning techniques and require the data to be labeled. Therefore, it becomes essential to endeavor future research in this direction to collect sustainable building data that is genuine and heterogeneous to lessen this issue.

## 7. Conclusions

This paper presented a comprehensive review discussing Artificial Intelligence (AI) techniques for Building Energy Management Systems (BEMS) that enable energy efficiency while taking thermal comfort into consideration. In addition, in order to evaluate the outputs of AI-based methods in energy savings and thermal comfort enhancement, assessments of the implementations of these techniques conducted in the published works have been reviewed and compiled according to the eligibility criteria. The research method used in the peer-reviewed publications was primarily empirical case-study, with data sources and data on thermal comfort and energy usage were collected predominantly through the execution of real-world studies (questionnaires or interviews with the occupants and data measurements) or by using current and publicly accessible datasets.

The findings of the study showed that multiple types of AI-based techniques were used in different parts of building control systems. In particular, the artificial neural networks (ANNs) have been used to overcome problems related to recognition and identification, and their function focused on learning algorithms which allow them to retain and classify data. In building management, ANNs were implemented to describe thermal comfort and estimate the Predicted Mean Vote (PMV) index. Fuzzy Logic (FL) is one of the recent tendencies which was developed to model human decision-making. Research works using FL have been documented since the late 1990's to treat thermal comfort as a subjective or fuzzy parameter. They were built to monitor conditions where the highest level of satisfaction and optimum energy-efficiency were achieved, most of the FL based studies used the PMV comfort index. This line of management approaches, focused on experience and judgement, aim at achieving simple, scalable and effective regulation, without resorting to a system model. Their efficiency is generally compared to traditional controls, and their advantage resides primarily attributed to the fact that additional awareness of system behavior (expressed in natural language – fuzzy or incorporated learning techniques – ANNs) or a degree of optimality (i.e., Genetic Algorithms) is assumed.

The review shows that the implementation of AI and ML technology in the building industry is still an ongoing research endeavor. This is partly attributed to the fact that this type of algorithms typically needs a massive quantity of high-quality real-world data, yet buildings or, more specifically, the energy sector has so far had little data. Adjustments and technical advancements are contributing to a rise in the quantity and sophistication of data (Smart Meter Installation, Internet of Things (IoT), Cloud Storage, and so forth) allowing to build much more effective data-driven research. The paper concludes by describing the research challenges facing the research community namely the need for more data for AI-based modeling in buildings, IoT based smart and connected buildings to facilitate efficient management and data collection for further studies. Smart buildings will also present security, privacy and data sensitivity issues as well as big-data streaming. Context awareness mechanisms that improve the intelligence of buildings in adapting to human behavior to adjust dynamically comfort and improve energy in a more fine-grain manner is also of high importance to the community. Another line of research includes also humans in the loop, comfort modeling for dynamic temperature set-point adjustments. This type of research will need mixed methods types of research where AI and ML techniques will open up opportunities for more energy savings while keeping the building comfortable. In particular adjusting dynamic set-points in commercial buildings will depend on how comfort modeling is connected to human activity in the building. Tracking human activity brings the notion of context-awareness as another line of research that will provide value to efficient building management with satisfactory comfort levels to its occupants. As these models are exchanging data about the building and its occupants, security and privacy become important issues to investigate in smart buildings. Indeed, smart buildings bring about many interesting research challenges that are still active line of research with many opportunities with the application of AI and ML.

# References

[1] Roaf S, Nicol F, Humphreys M, Tuohy P, Boerstra A. Twentieth century standards for thermal comfort: promoting high energy buildings. Architectural Science Review 2010;53:65–77. https://doi.org/10.3763/asre.2009.0111.

[2] Cândido C, de Dear R, Lamberts R, Bittencourt L. Cooling exposure in hot humid climates: are occupants 'addicted'? Transforming Markets in the Built Environment, Routledge; 2012, p. 59–64.

[3] Saman W, Boland J, Pullen S, Dear R de, Soebarto V, Miller W, et al. A framework for adaptation Australian households to heat waves: Final Report. Australia: National Climate Change Adaptation Research Facility [NCCARF Publication 87/13]; 2013.

[4] Air conditioning use emerges as one of the key drivers of global electricity-demand growth. IEA 2018. https://www.iea.org/news/air-conditioning-use-emerges-as-one-of-the-key-drivers-of-global-electricity-demand-growth (accessed February 7, 2020).

[5] Soares N, Bastos J, Pereira LD, Soares A, Amaral A, Asadi E, et al. A review on current advances in the energy and environmental performance of buildings towards a more sustainable built environment. Renewable and Sustainable Energy Reviews 2017;77:845–60.

[6] Humphreys MA, Fergus Nicol J. The validity of ISO-PMV for predicting comfort votes in every-day thermal environments. Energy and Buildings 2002;34:667–84. https://doi.org/10.1016/S0378-7788(02)00018-X.

[7] Yang R, Newman MW. Learning from a learning thermostat: lessons for intelligent systems for the home. Proceedings of the 2013 ACM international joint conference on Pervasive and ubiquitous computing - UbiComp '13, Zurich, Switzerland: ACM Press; 2013, p. 93. https://doi.org/10.1145/2493432.2493489.

[8] Asakawa K, Takagi H. Neural networks in Japan. Commun ACM 1994;37:106–13. https://doi.org/10.1145/175247.175258.

[9] Dounis AI, Caraiscos C. Advanced control systems engineering for energy and comfort management in a building environment—A review. Renewable and Sustainable Energy Reviews 2009;13:1246–61. https://doi.org/10.1016/j.rser.2008.09.015.

[10] Shaikh PH, Nor NBM, Nallagownden P, Elamvazuthi I, Ibrahim T. A review on optimized control systems for building energy and comfort management of smart sustainable buildings. Renewable and Sustainable Energy Reviews 2014;34:409–29. https://doi.org/10.1016/j.rser.2014.03.027.

[11] Evins R. A review of computational optimisation methods applied to sustainable building design. Renewable and Sustainable Energy Reviews 2013;22:230–45. https://doi.org/10.1016/j.rser.2013.02.004.

[12] Behrooz F, Mariun N, Marhaban M, Mohd Radzi M, Ramli A. Review of Control Techniques for HVAC Systems—Nonlinearity Approaches Based on Fuzzy Cognitive Maps. Energies 2018;11:495. https://doi.org/10.3390/en11030495.

[13] Cheng C-C, Lee D. Artificial Intelligence-Assisted Heating Ventilation and Air Conditioning Control and the Unmet Demand for Sensors: Part 1. Problem Formulation and the Hypothesis. Sensors 2019;19:1131. https://doi.org/10.3390/s19051131.

[14] Royapoor M, Antony A, Roskilly T. A review of building climate and plant controls, and a survey of industry perspectives. Energy and Buildings 2018;158:453–65. https://doi.org/10.1016/j.enbuild.2017.10.022.

[15] Qolomany B, Al-Fuqaha A, Gupta A, Benhaddou D, al-wajidi S, Qadir J, et al. Leveraging Machine Learning and Big Data for Smart Buildings: A Comprehensive Survey. IEEE Access 2019;7:90316–56. https://doi.org/10.1109/ACCESS.2019.2926642.

[16] Ngarambe J, Yun GY, Santamouris M. The use of artificial intelligence (AI) methods in the prediction of thermal comfort in buildings: energy implications of AI-based thermal comfort controls. Energy and Buildings 2020;211:109807. https://doi.org/10.1016/j.enbuild.2020.109807.

[17] ACM Digital Library n.d. https://dl.acm.org/ (accessed December 26, 2020).

[18] Scopus n.d. https://www.scopus.com/home.uri (accessed December 26, 2020).

[19] Google Scholar n.d. https://scholar.google.gr/ (accessed December 26, 2020).

[20] IEEE Xplore n.d. https://ieeexplore.ieee.org/Xplore/home.jsp (accessed December 26, 2020).

[21] Web of Science. Web of Science Group, a Clarivate Company n.d. https://mjl.clarivate.com/ (accessed December 26, 2020).

[22] Science Direct n.d. https://www.sciencedirect.com/ (accessed December 26, 2020).

[23] Moher D, Liberati A, Tetzlaff J, Altman DG, PRISMA Group. Preferred reporting items for systematic reviews and meta-analyses: the PRISMA statement. PLoS Med 2009;6:336–41. https://doi.org/10.1371/journal.pmed.1000097.

[24] Liberati A, Altman DG, Tetzlaff J, Mulrow C, Gøtzsche PC, Ioannidis JP, et al. The PRISMA statement for reporting systematic reviews and meta-analyses of studies that evaluate health care interventions: explanation and elaboration. Annals of Internal Medicine 2009;151:W-65. https://doi.org/10.1371/journal.pmed.1000100.

[25] Rosenblatt F. The perceptron: A Perceiving and Recognizing Automaton. Buffalo, New York: Cornell Aeronautical Laboratory; 1957.
[26] Rosenblatt F. The perceptron: A probabilistic model for information storage and organization in the brain. Psychological Review 1958;65:386–408. https://doi.org/10.1037/h0042519.
[27] Hopfield JJ. Neural networks and physical systems with emergent collective computational abilities. Proceedings of the National Academy of Sciences 1982;79:2554–8.
[28] Rumelhart DE, Hinton GE, Williams RJ. Learning representations by back-propagating errors. Nature 1986;323:533–6. https://doi.org/10.1038/323533a0.
[29] Rumelhart DE, Hinton GE, Williams RJ. Learning Internal Representations by Error Propagation. Parallel Distributed Processing: Explorations in the Microstructure of Cognition, Vol. 1: Foundations, Cambridge, MA, USA: MIT Press; 1986, p. 318–62.
[30] Hinton GE, Osindero S, Teh Y-W. A fast-learning algorithm for deep belief nets. Neural Computation 2006;18:1527–54. https://doi.org/10.1162/neco.2006.18.7.1527.
[31] Ruslan Salakhutdinov, Geoffrey Hinton. Deep Boltzmann Machines. In: David van Dyk, Max Welling, editors. Proceedings of the Twelth International Conference on Artificial Intelligence and Statistics, PMLR; 2009, p. 448–55.
[32] Jian Liang, Ruxu Du. Thermal comfort control based on neural network for HVAC application. Proceedings of 2005 IEEE Conference on Control Applications, 2005. CCA 2005., Toronto, Canada: IEEE; 2005, p. 819–24. https://doi.org/10.1109/CCA.2005.1507230.
[33] Pombeiro H, Machado MJ, Silva C. Dynamic programming and genetic algorithms to control an HVAC system: Maximizing thermal comfort and minimizing cost with PV production and storage. Sustainable Cities and Society 2017;34:228–38. https://doi.org/10.1016/j.scs.2017.05.021.
[34] Wei T, Wang Y, Zhu Q. Deep Reinforcement Learning for Building HVAC Control. Proceedings of the 54th Annual Design Automation Conference 2017 on - DAC '17, Austin, TX, USA: ACM Press; 2017, p. 1–6. https://doi.org/10.1145/3061639.3062224.
[35] Auffenberg F, Snow S, Stein S, Rogers A. A Comfort-Based Approach to Smart Heating and Air Conditioning. ACM Trans Intell Syst Technol 2017;9:1–20. https://doi.org/10.1145/3057730.
[36] Liang J, Du R. Design of intelligent comfort control system with human learning and minimum power control strategies. Energy Conversion and Management 2008;49:517–28. https://doi.org/10.1016/j.enconman.2007.08.006.
[37] Torres J, Martin M, Juan S, San J. Adaptive Control of Thermal Comfort Using Neural Networks. Argentine Symposium on Computing Technology 2008:12.
[38] Kusiak A, Xu G. Modeling and optimization of HVAC systems using a dynamic neural network. Energy 2012;42:241–50. https://doi.org/10.1016/j.energy.2012.03.063.
[39] Fraisse G, Virgone J, Roux JJ. Thermal control of a discontinuously occupied building using a classical and a fuzzy logic approach. Energy and Buildings 1997;26:303–16. https://doi.org/10.1016/S0378-7788(97)00011-X.
[40] Kolokotsa D, Tsiavos D, Stavrakakis GS, Kalaitzakis K, Antonidakis E. Advanced fuzzy logic controllers design and evaluation for buildings' occupants thermal–visual comfort and indoor air quality satisfaction. Energy and Buildings 2001;33:531–43. https://doi.org/10.1016/S0378-7788(00)00098-0.
[41] Bernard Th, Kuntze H-B. Multi-objective optimization of building climate control systems using fuzzy-logic. 1999 European Control Conference (ECC), Karlsruhe: IEEE; 1999, p. 2572–7. https://doi.org/10.23919/ECC.1999.7099712.
[42] Hamdi M, Lachiver G. A fuzzy control system based on the human sensation of thermal comfort. 1998 IEEE International Conference on Fuzzy Systems Proceedings. IEEE World Congress on Computational Intelligence (Cat. No.98CH36228), vol. 1, Anchorage, AK, USA: IEEE; 1998, p. 487–92. https://doi.org/10.1109/FUZZY.1998.687534.
[43] Wang Z, Yang R, Wang L, Green RC, Dounis AI. A fuzzy adaptive comfort temperature model with grey predictor for multi-agent control system of smart building. 2011 IEEE Congress of Evolutionary Computation (CEC), New Orleans, LA, USA: IEEE; 2011, p. 728–35. https://doi.org/10.1109/CEC.2011.5949691.
[44] Jassar S, Liao Z, Zhao L. Adaptive neuro-fuzzy based inferential sensor model for estimating the average air temperature in space heating systems. Building and Environment 2009;44:1609–16. https://doi.org/10.1016/j.buildenv.2008.10.002.
[45] Soyguder S, Alli H. Predicting of fan speed for energy saving in HVAC system based on adaptive network based fuzzy inference system. Expert Systems with Applications 2009;36:8631–8. https://doi.org/10.1016/j.eswa.2008.10.033.
[46] Magnier L, Haghighat F. Multiobjective optimization of building design using TRNSYS simulations, genetic algorithm, and Artificial Neural Network. Building and Environment 2010;45:739–46. https://doi.org/10.1016/j.buildenv.2009.08.016.

[47] Moon JW, Jung SK, Kim Y, Han S-H. Comparative study of artificial intelligence-based building thermal control methods – Application of fuzzy, adaptive neuro-fuzzy inference system, and artificial neural network. Applied Thermal Engineering 2011;31:2422–9. https://doi.org/10.1016/j.applthermaleng.2011.04.006.

[48] Collotta M, Messineo A, Nicolosi G, Pau G. A Dynamic Fuzzy Controller to Meet Thermal Comfort by Using Neural Network Forecasted Parameters as the Input. Energies 2014;7:4727–56. https://doi.org/10.3390/en7084727.

[49] Javed A, Larijani H, Ahmadinia A, Emmanuel R. Modelling and optimization of residential heating system using random neural networks. 2014 IEEE International Conference on Control Science and Systems Engineering, Yantai, China: IEEE; 2014, p. 90–5. https://doi.org/10.1109/CCSSE.2014.7224515.

[50] Moon JW, Lee J-H, Kim S. Application of control logic for optimum indoor thermal environment in buildings with double skin envelope systems. Energy and Buildings 2014;85:59–71. https://doi.org/10.1016/j.enbuild.2014.09.018.

[51] Emmanuel R, Clark C, Ahmadinia A, Javed A, Gibson D, Larijani H. Experimental testing of a random neural network smart controller using a single zone test chamber. IET Networks 2015;4:350–8. https://doi.org/10.1049/iet-net.2015.0020.

[52] Garnier A, Eynard J, Caussanel M, Grieu S. Predictive control of multizone heating, ventilation and air-conditioning systems in non-residential buildings. Applied Soft Computing 2015;37:847–62. https://doi.org/10.1016/j.asoc.2015.09.022.

[53] Moon JW. Comparative performance analysis of the artificial-intelligence-based thermal control algorithms for the double-skin building. Applied Thermal Engineering 2015;91:334–44. https://doi.org/10.1016/j.applthermaleng.2015.08.038.

[54] Ku KL, Liaw JS, Tsai MY, Liu TS. Automatic Control System for Thermal Comfort Based on Predicted Mean Vote and Energy Saving. IEEE Transactions on Automation Science and Engineering 2015;12:378–83. https://doi.org/10.1109/TASE.2014.2366206.

[55] Moon JW, Jung SK. Algorithm for optimal application of the setback moment in the heating season using an artificial neural network model. Energy and Buildings 2016;127:859–69. https://doi.org/10.1016/j.enbuild.2016.06.046.

[56] Moon JW, Jung SK. Development of a thermal control algorithm using artificial neural network models for improved thermal comfort and energy efficiency in accommodation buildings. Applied Thermal Engineering 2016;103:1135–44. https://doi.org/10.1016/j.applthermaleng.2016.05.002.

[57] Soudari M, Srinivasan S, Balasubramanian S, Vain J, Kotta U. Learning based personalized energy management systems for residential buildings. Energy and Buildings 2016;127:953–68. https://doi.org/10.1016/j.enbuild.2016.05.059.

[58] Ahn J, Cho S. Anti-logic or common sense that can hinder machine's energy performance: Energy and comfort control models based on artificial intelligence responding to abnormal indoor environments. Applied Energy 2017;204:117–30. https://doi.org/10.1016/j.apenergy.2017.06.079.

[59] Ahn J, Cho S. Development of an intelligent building controller to mitigate indoor thermal dissatisfaction and peak energy demands in a district heating system. Building and Environment 2017;124:57–68. https://doi.org/10.1016/j.buildenv.2017.07.040.

[60] Danassis P, Siozios K, Korkas C, Soudris D, Kosmatopoulos E. A low-complexity control mechanism targeting smart thermostats. Energy and Buildings 2017;139:340–50. https://doi.org/10.1016/j.enbuild.2017.01.013.

[61] Javed A, Larijani H, Ahmadinia A, Emmanuel R, Mannion M, Gibson D. Design and Implementation of a Cloud Enabled Random Neural Network-Based Decentralized Smart Controller With Intelligent Sensor Nodes for HVAC. IEEE Internet of Things Journal 2017;4:393–403. https://doi.org/10.1109/JIOT.2016.2627403.

[62] Javed A, Larijani H, Ahmadinia A, Gibson D. Smart Random Neural Network Controller for HVAC Using Cloud Computing Technology. IEEE Transactions on Industrial Informatics 2017;13:351–60. https://doi.org/10.1109/TII.2016.2597746.

[63] Macarulla M, Casals M, Forcada N, Gangolells M. Implementation of predictive control in a commercial building energy management system using neural networks. Energy and Buildings 2017;151:511–9. https://doi.org/10.1016/j.enbuild.2017.06.027.

[64] Reynolds J, Hippolyte J-L, Rezgui Y. A smart heating set point scheduler using an artificial neural network and genetic algorithm. 2017 International Conference on Engineering, Technology and Innovation (ICE/ITMC), Funchal: IEEE; 2017, p. 704–10. https://doi.org/10.1109/ICE.2017.8279954.

[65] Yuce B, Rezgui Y. An ANN-GA Semantic Rule-Based System to Reduce the Gap Between Predicted and Actual Energy Consumption in Buildings. IEEE Transactions on Automation Science and Engineering 2017;14:1351–63. https://doi.org/10.1109/TASE.2015.2490141.

[66] Zhai D, Soh YC. Balancing indoor thermal comfort and energy consumption of ACMV systems via sparse swarm algorithms in optimizations. Energy and Buildings 2017;149:1–15. https://doi.org/10.1016/j.enbuild.2017.05.019.

[67] Zhong C, Choi J-H. Development of a Data-Driven Approach for Human-Based Environmental Control. Procedia Engineering 2017;205:1665–71. https://doi.org/10.1016/j.proeng.2017.10.341.

[68] Rajith A, Soki S, Hiroshi M. Real-time optimized HVAC control system on top of an IoT framework. 2018 Third International Conference on Fog and Mobile Edge Computing (FMEC), Barcelona: IEEE; 2018, p. 181–6. https://doi.org/10.1109/FMEC.2018.8364062.

[69] Chaudhuri T, Soh YC, Li H, Xie L. A feedforward neural network based indoor-climate control framework for thermal comfort and energy saving in buildings. Applied Energy 2019;248:44–53. https://doi.org/10.1016/j.apenergy.2019.04.065.

[70] Chen Y, Chandna V, Huang Y, Alam MJE, Ahmed O, Smith L. Coordination of Behind-the-Meter Energy Storage and Building Loads: Optimization with Deep Learning Model. Proceedings of the Tenth ACM International Conference on Future Energy Systems - e-Energy '19, Phoenix, AZ, USA: ACM Press; 2019, p. 492–9. https://doi.org/10.1145/3307772.3331025.

[71] Gao G, Li J, Wen Y. Energy-Efficient Thermal Comfort Control in Smart Buildings via Deep Reinforcement Learning. ArXiv:190104693 [Cs] 2019.

[72] Ghofrani A, Nazemi SD, Jafari MA. Prediction of building indoor temperature response in variable air volume systems. Journal of Building Performance Simulation 2020;13:34–47. https://doi.org/10.1080/19401493.2019.1688393.

[73] Moon JW, Ahn J. Improving sustainability of ever-changing building spaces affected by users' fickle taste: A focus on human comfort and energy use. Energy and Buildings 2020;208:109662. https://doi.org/10.1016/j.enbuild.2019.109662.

[74] Moustris K, Kavadias KA, Zafirakis D, Kaldellis JK. Medium, short and very short-term prognosis of load demand for the Greek Island of Tilos using artificial neural networks and human thermal comfort-discomfort biometeorological data. Renewable Energy 2020;147:100–9. https://doi.org/10.1016/j.renene.2019.08.126.

[75] Sadeghi A, Younes Sinaki R, Young WA, Weckman GR. An Intelligent Model to Predict Energy Performances of Residential Buildings Based on Deep Neural Networks. Energies 2020;13:571. https://doi.org/10.3390/en13030571.

[76] Shan C, Hu J, Wu J, Zhang A, Ding G, Xu LX. Towards non-intrusive and high accuracy prediction of personal thermal comfort using a few sensitive physiological parameters. Energy and Buildings 2020;207:109594. https://doi.org/10.1016/j.enbuild.2019.109594.

[77] Sung L-Y, Ahn J. Comparative Analyses of Energy Efficiency between on-Demand and Predictive Controls for Buildings' Indoor Thermal Environment. Energies 2020;13:1089. https://doi.org/10.3390/en13051089.

[78] Zhao Y, Genovese PV, Li Z. Intelligent Thermal Comfort Controlling System for Buildings Based on IoT and AI. Future Internet 2020;12:30. https://doi.org/10.3390/fi12020030.

[79] Gouda MM, Danaher S, Underwood CP. Thermal comfort based fuzzy logic controller. Building Services Engineering Research and Technology 2001;22:237–53. https://doi.org/10.1177/014362440102200403.

[80] Kolokotsa D. Comparison of the performance of fuzzy controllers for the management of the indoor environment. Building and Environment 2003;38:1439–50. https://doi.org/10.1016/S0360-1323(03)00130-6.

[81] Shepherd AB, Batty WJ. Fuzzy control strategies to provide cost and energy efficient high quality indoor environments in buildings with high occupant densities. Building Services Engineering Research and Technology 2003;24:35–45. https://doi.org/10.1191/0143624403bt059oa.

[82] Kolokotsa D, Niachou K, Geros V, Kalaitzakis K, Stavrakakis GS, Santamouris M. Implementation of an integrated indoor environment and energy management system. Energy and Buildings 2005;37:93–9. https://doi.org/10.1016/j.enbuild.2004.05.008.

[83] Lah MT, Zupančič B, Krainer A. Fuzzy control for the illumination and temperature comfort in a test chamber. Building and Environment 2005;40:1626–37. https://doi.org/10.1016/j.buildenv.2004.11.008.

[84] Ari S, Cosden IA, Khalifa HE, Dannenhoffer JF, Wilcoxen P, Isik C. Constrained Fuzzy Logic Approximation for Indoor Comfort and Energy Optimization. NAFIPS 2005 - 2005 Annual Meeting of the North American Fuzzy Information Processing Society, Detroit, MI, USA: IEEE; 2005, p. 500–4. https://doi.org/10.1109/NAFIPS.2005.1548586.

[85] Y. Huang, N. Li, Y. Huang. Indoor Thermal Comfort Control Research Based on Adaptive Fuzzy Strategy. The Proceedings of the Multiconference on "Computational Engineering in Systems Applications," vol. 2, 2006, p. 1969–72. https://doi.org/10.1109/CESA.2006.4281961.

[86] Sierra E, Hossian A, Britos P, Rodriguez D, Garcia-Martinez R. Fuzzy Control for Improving Energy Management within Indoor Building Environments. Electronics, Robotics and Automotive Mechanics

[87] Li C, Zhang G, Wang M, Yi J. Data-driven modeling and optimization of thermal comfort and energy consumption using type-2 fuzzy method. Soft Computing 2013;17:2075–88. https://doi.org/10.1007/s00500-013-1117-4.

[88] Wijayasekara D, Manic M, Rieger C. Fuzzy linguistic knowledge-based behavior extraction for building energy management systems. 2013 6th International Symposium on Resilient Control Systems (ISRCS), San Francisco, CA, USA: IEEE; 2013, p. 80–5. https://doi.org/10.1109/ISRCS.2013.6623755.

[89] Jazizadeh F, Ghahramani A, Becerik-Gerber B, Kichkaylo T, Orosz M. User-led decentralized thermal comfort driven HVAC operations for improved efficiency in office buildings. Energy and Buildings 2014;70:398–410. https://doi.org/10.1016/j.enbuild.2013.11.066.

[90] Kang C-S, Hyun C-H, Park M. Fuzzy logic-based advanced on–off control for thermal comfort in residential buildings. Applied Energy 2015;155:270–83. https://doi.org/10.1016/j.apenergy.2015.05.119.

[91] Zampetti L, Arnesano M, Revel GM. Experimental testing of a system for the energy-efficient sub-zonal heating management in indoor environments based on PMV. Energy and Buildings 2018;166:229–38. https://doi.org/10.1016/j.enbuild.2018.02.019.

[92] Li W, Zhang J, Zhao T. Indoor thermal environment optimal control for thermal comfort and energy saving based on online monitoring of thermal sensation. Energy and Buildings 2019;197:57–67. https://doi.org/10.1016/j.enbuild.2019.05.050.

[93] Teixeira MS, Maran V, de Oliveira JPM, Winter M, Machado A. Situation-aware model for multi-objective decision making in ambient intelligence. Applied Soft Computing 2019;81:105532. https://doi.org/10.1016/j.asoc.2019.105532.

[94] Homod RZ, Gaeid KS, Dawood SM, Hatami A, Sahari KS. Evaluation of energy-saving potential for optimal time response of HVAC control system in smart buildings. Applied Energy 2020;271:115255. https://doi.org/10.1016/j.apenergy.2020.115255.

[95] Li W, Zhang J, Zhao T, Ren J. Experimental study of an indoor temperature fuzzy control method for thermal comfort and energy saving using wristband device. Building and Environment 2021;187:107432. https://doi.org/10.1016/j.buildenv.2020.107432.

[96] Ferber J, Weiss G. Multi-agent systems: an introduction to distributed artificial intelligence. vol. 1. Addison-Wesley Reading; 1999.

[97] Reynolds CW. Flocks, herds and schools: A distributed behavioral model. Proceedings of the 14th annual conference on Computer graphics and interactive techniques, 1987, p. 25–34.

[98] Klein L, Kwak J, Kavulya G, Jazizadeh F, Becerik-Gerber B, Varakantham P, et al. Coordinating occupant behavior for building energy and comfort management using multi-agent systems. Automation in Construction 2012;22:525–36. https://doi.org/10.1016/j.autcon.2011.11.012.

[99] Davidsson P, Boman M. Distributed monitoring and control of office buildings by embedded agents. Information Sciences 2005;171:293–307. https://doi.org/10.1016/j.ins.2004.09.007.

[100] Mo Z. An Agent-Based Simulation-Assisted Approach to Bi-Lateral Building Systems Control. PhD Thesis. Carnegie Mellon University, 2003.

[101] Dounis A, Caraiscos C. Intelligent technologies for energy efficiency and comfort in a building environment. International conference of technology and automation, 2005, p. 91–5.

[102] Barakat M, Khoury H. An agent-based framework to study occupant multi-comfort level in office buildings. 2016 Winter Simulation Conference (WSC), Washington, DC, USA: IEEE; 2016, p. 1328–39. https://doi.org/10.1109/WSC.2016.7822187.

[103] Alfakara A, Croxford B. Using Agent-Based Modelling to Simulate Occupants' Behaviours in Response to Summer Overheating. Proceedings of the Symposium on Simulation for Architecture & Urban Design, San Diego, CA, USA: Society for Computer Simulation International; 2014.

[104] Hadjiski M, Sgurev V, Boishina V. Multi-Agent Intelligent Control of Centralized HVAC Systems. IFAC Proceedings Volumes 2006;39:195–200. https://doi.org/10.3182/20061002-4-BG-4905.00033.

[105] Klein L, Kavulya G, FarrokhJazizadeh, Kwak J, Becerik-Gerber B, Varakantham P, et al. Towards Optimization of Building Energy and Occupant Comfort Using Multi-Agent Simulation. International Symposium on Automation and Robotics in Construction, Seoul, Korea: 2011. https://doi.org/10.22260/ISARC2011/0044.

[106] Yang R, Wang L. Energy management of multi-zone buildings based on multi-agent control and particle swarm optimization. 2011 IEEE International Conference on Systems, Man, and Cybernetics, Anchorage, AK, USA: IEEE; 2011, p. 159–64. https://doi.org/10.1109/ICSMC.2011.6083659.

[107] Wang Z, Wang L, Dounis AI, Yang R. Multi-agent control system with information fusion-based comfort model for smart buildings. Applied Energy 2012;99:247–54. https://doi.org/10.1016/j.apenergy.2012.05.020.

[108] Yang R, Wang L. Development of multi-agent system for building energy and comfort management based on occupant behaviors. Energy and Buildings 2013;56:1–7. https://doi.org/10.1016/j.enbuild.2012.10.025.
[109] Langevin J, Wen J, Gurian PL. Including Occupants in Building Performance Simulation: Integration of an Agent-Based Occupant Behavior Algorithm with EnergyPlus. 2014 ASHRAE/IBPSA-USA Buidling Simulation Conference. Atlanta, GA, Atlanta, GA, USA: ASHRAE; 2014, p. 8.
[110] Mokhtar M, Liu X, Howe J. Multi-agent Gaussian Adaptive Resonance Theory Map for building energy control and thermal comfort management of UCLan's WestLakes Samuel Lindow Building. Energy and Buildings 2014;80:504–16. https://doi.org/10.1016/j.enbuild.2014.05.045.
[111] Zupančič D, Luštrek M, Gams M. Multi-Agent Architecture for Control of Heating and Cooling in a Residential Space. The Computer Journal 2015;58:1314–29. https://doi.org/10.1093/comjnl/bxu058.
[112] Sutton RS, Barto AG. Reinforcement Learning: An Introduction. vol. 135. Second Edition. MIT press Cambridge; 1998.
[113] Gwerder M, Tödtli Siemens J. Predictive control for integrated room automation. 2005.
[114] Afram A, Janabi-Sharifi F. Theory and applications of HVAC control systems – A review of model predictive control (MPC). Building and Environment 2014;72:343–55. https://doi.org/10.1016/j.buildenv.2013.11.016.
[115] Alcalá R, Benítez JM, Casillas J, Casillas J, Cordón O, Pérez R. Fuzzy Control of HVAC Systems Optimized by Genetic Algorithms. Applied Intelligence 2003;18:155–77. https://doi.org/10.1023/A:1021986309149.
[116] Alcalá R, Casillas J, Cordón O, González A, Herrera F. A genetic rule weighting and selection process for fuzzy control of heating, ventilating and air conditioning systems. Engineering Applications of Artificial Intelligence 2005;18:279–96. https://doi.org/10.1016/j.engappai.2004.09.007.
[117] Dounis AI, Tiropanis P, Argiriou A, Diamantis A. Intelligent control system for reconciliation of the energy savings with comfort in buildings using soft computing techniques. Energy and Buildings 2011;43:66–74. https://doi.org/10.1016/j.enbuild.2010.08.014.
[118] Hussain S, Gabbar HA, Bondarenko D, Musharavati F, Pokharel S. Comfort-based fuzzy control optimization for energy conservation in HVAC systems. Control Engineering Practice 2014;32:172–82. https://doi.org/10.1016/j.conengprac.2014.08.007.
[119] Jahedi G, Ardehali MM. Genetic algorithm-based fuzzy-PID control methodologies for enhancement of energy efficiency of a dynamic energy system. Energy Conversion and Management 2011;52:725–32. https://doi.org/10.1016/j.enconman.2010.07.051.
[120] Kolokotsa D, Stavrakakis GS, Kalaitzakis K, Agoris D. Genetic algorithms optimized fuzzy controller for the indoor environmental management in buildings implemented using PLC and local operating networks. Engineering Applications of Artificial Intelligence 2002;15:417–28. https://doi.org/10.1016/S0952-1976(02)00090-8.
[121] Dounis AI, Caraiscos C. Intelligent Coordinator of Fuzzy Controller-Agents for Indoor Environment Control in Buildings Using 3-D Fuzzy Comfort Set. 2007 IEEE International Fuzzy Systems Conference, London, UK: IEEE; 2007, p. 1–6. https://doi.org/10.1109/FUZZY.2007.4295573.
[122] Smitha SD, Savier JS, Mary Chacko F. Intelligent control system for efficient energy management in commercial buildings. 2013 Annual International Conference on Emerging Research Areas and 2013 International Conference on Microelectronics, Communications and Renewable Energy, Kanjirapally, India: IEEE; 2013, p. 1–6. https://doi.org/10.1109/AICERA-ICMiCR.2013.6575942.
[123] Gou S, Nik VM, Scartezzini J-L, Zhao Q, Li Z. Passive design optimization of newly-built residential buildings in Shanghai for improving indoor thermal comfort while reducing building energy demand. Energy and Buildings 2018;169:484–506. https://doi.org/10.1016/j.enbuild.2017.09.095.
[124] Ruano AE, Pesteh S, Silva S, Duarte H, Mestre G, Ferreira PM, et al. The IMBPC HVAC system: A complete MBPC solution for existing HVAC systems. Energy and Buildings 2016;120:145–58. https://doi.org/10.1016/j.enbuild.2016.03.043.
[125] Zhang T, Liu Y, Rao Y, Li X, Zhao Q. Optimal design of building environment with hybrid genetic algorithm, artificial neural network, multivariate regression analysis and fuzzy logic controller. Building and Environment 2020;175:106810. https://doi.org/10.1016/j.buildenv.2020.106810.
[126] Shaikh PH, Nor NBM, Nallagownden P, Elamvazuthi I. Stochastic optimized intelligent controller for smart energy efficient buildings. Sustainable Cities and Society 2014;13:41–5. https://doi.org/10.1016/j.scs.2014.04.005.
[127] Hurtado LA, Nguyen PH, Kling WL. Smart grid and smart building inter-operation using agent-based particle swarm optimization. Sustainable Energy, Grids and Networks 2015;2:32–40. https://doi.org/10.1016/j.segan.2015.03.003.

[128] Peng Y, Rysanek A, Nagy Z, Schlüter A. Using machine learning techniques for occupancy-prediction-based cooling control in office buildings. Applied Energy 2018;211:1343–58. https://doi.org/10.1016/j.apenergy.2017.12.002.
[129] Xu H, He Y, Sun X, He J, Xu Q. Prediction of thermal energy inside smart homes using IoT and classifier ensemble techniques. Computer Communications 2020;151:581–9. https://doi.org/10.1016/j.comcom.2019.12.020.
[130] Zhou Y, Zheng S. Machine-learning based hybrid demand-side controller for high-rise office buildings with high energy flexibilities. Applied Energy 2020;262:114416. https://doi.org/10.1016/j.apenergy.2019.114416.
[131] Wang R, Lu S, Feng W. A three-stage optimization methodology for envelope design of passive house considering energy demand, thermal comfort and cost. Energy 2020;192:116723. https://doi.org/10.1016/j.energy.2019.116723.
[132] Dalamagkidis K, Kolokotsa D, Kalaitzakis K, Stavrakakis GS. Reinforcement learning for energy conservation and comfort in buildings. Building and Environment 2007;42:2686–98. https://doi.org/10.1016/j.buildenv.2006.07.010.
[133] Fazenda P, Veeramachaneni K, Lima P, O'Reilly U-M. Using reinforcement learning to optimize occupant comfort and energy usage in HVAC systems. JAISE 2014;6:675–90. https://doi.org/10.3233/AIS-140288.
[134] Li B, Xia L. A multi-grid reinforcement learning method for energy conservation and comfort of HVAC in buildings. 2015 IEEE International Conference on Automation Science and Engineering (CASE), Gothenburg, Sweden: IEEE; 2015, p. 444–9. https://doi.org/10.1109/CoASE.2015.7294119.
[135] Yuan Wang, Kirubakaran Velswamy, Biao Huang. A Long-Short Term Memory Recurrent Neural Network Based Reinforcement Learning Controller for Office Heating Ventilation and Air Conditioning Systems. Processes 2017;5:46. https://doi.org/10.3390/pr5030046.
[136] Hurtado LA, Mocanu E, Nguyen PH, Gibescu M, Kamphuis RIG. Enabling Cooperative Behavior for Building Demand Response Based on Extended Joint Action Learning. IEEE Transactions on Industrial Informatics 2018;14:127–36. https://doi.org/10.1109/TII.2017.2753408.
[137] Marantos C, Lamprakos CP, Tsoutsouras V, Siozios K, Soudris D. Towards plug&play smart thermostats inspired by reinforcement learning. Proceedings of the Workshop on INTelligent Embedded Systems Architectures and Applications - INTESA '18, Turin, Italy: ACM Press; 2018, p. 39–44. https://doi.org/10.1145/3285017.3285024.
[138] Zhang Z, Chong A, Pan Y, Zhang C, Lu S, Lam KP. A deep reinforcement learning approach to using whole building energy model for hvac optimal control. 2018 Building Performance Analysis Conference and SimBuild, 2018, p. 9.
[139] Valladares W, Galindo M, Gutiérrez J, Wu W-C, Liao K-K, Liao J-C, et al. Energy optimization associated with thermal comfort and indoor air control via a deep reinforcement learning algorithm. Building and Environment 2019;155:105–17. https://doi.org/10.1016/j.buildenv.2019.03.038.
[140] Hosseinloo AH, Ryzhov A, Bischi A, Ouerdane H, Turitsyn K, Dahleh MA. Data-driven control of micro-climate in buildings; an event-triggered reinforcement learning approach. ArXiv:200110505 [Cs, Eess] 2020.
[141] Zou Z, Yu X, Ergan S. Towards optimal control of air handling units using deep reinforcement learning and recurrent neural network. Building and Environment 2020;168:106535. https://doi.org/10.1016/j.buildenv.2019.106535.
[142] Aswani A, Master N, Taneja J, Culler D, Tomlin C. Reducing Transient and Steady State Electricity Consumption in HVAC Using Learning-Based Model-Predictive Control. Proceedings of the IEEE 2012;100:240–53. https://doi.org/10.1109/JPROC.2011.2161242.
[143] Barata FA, Neves-Silva R. Distributed model predictive control for thermal house comfort with auction of available energy. 2012 International Conference on Smart Grid Technology, Economics and Policies (SG-TEP), Nuremberg, Germany: IEEE; 2012, p. 1–4. https://doi.org/10.1109/SG-TEP.2012.6642375.
[144] Ferreira PM, Ruano AE, Silva S, Conceição EZE. Neural networks based predictive control for thermal comfort and energy savings in public buildings. Energy and Buildings 2012;55:238–51. https://doi.org/10.1016/j.enbuild.2012.08.002.
[145] Gao PX, Keshav S. Optimal Personal Comfort Management Using SPOT+. Proceedings of the 5th ACM Workshop on Embedded Systems for Energy-Efficient Buildings - BuildSys'13, Roma, Italy: ACM Press; 2013, p. 1–8. https://doi.org/10.1145/2528282.2528297.
[146] Dong B, Lam KP. A real-time model predictive control for building heating and cooling systems based on the occupancy behavior pattern detection and local weather forecasting. Building Simulation 2014;7:89–106. https://doi.org/10.1007/s12273-013-0142-7.

[147] Huang H, Chen L, Hu E. A new model predictive control scheme for energy and cost savings in commercial buildings: An airport terminal building case study. Building and Environment 2015;89:203–16. https://doi.org/10.1016/j.buildenv.2015.01.037.

[148] Ascione F, Bianco N, De Stasio C, Mauro GM, Vanoli GP. Simulation-based model predictive control by the multi-objective optimization of building energy performance and thermal comfort. Energy and Buildings 2016;111:131–44. https://doi.org/10.1016/j.enbuild.2015.11.033.

[149] Cotrufo N, Saloux E, Hardy JM, Candanedo JA, Platon R. A practical artificial intelligence-based approach for predictive control in commercial and institutional buildings. Energy and Buildings 2020;206:109563. https://doi.org/10.1016/j.enbuild.2019.109563.

[150] Jain A, Smarra F, Reticcioli E, D'Innocenzo A, Morari M. NeurOpt: Neural network based optimization for building energy management and climate control. ArXiv:200107831 [Cs, Eess] 2020.

[151] Javaid S, Javaid N. Comfort evaluation of seasonally and daily used residential load in smart buildings for hottest areas via predictive mean vote method. Sustainable Computing: Informatics and Systems 2020;25:100369. https://doi.org/10.1016/j.suscom.2019.100369.

[152] Zhou H, Rao M, Chuang KT. Knowledge-based automation for energy conservation and indoor air quality control in HVAC processes. Engineering Applications of Artificial Intelligence 1993;6:131–44. https://doi.org/10.1016/0952-1976(93)90029-W.

[153] Nassif N, Kajl S, Sabourin R. Two-objective on-line optimization of supervisory control strategy. Building Services Engineering Research and Technology 2004;25:241–51. https://doi.org/10.1191/0143624404bt105oa.

[154] Ríos-Moreno GJ, Trejo-Perea M, Castañeda-Miranda R, Hernández-Guzmán VM, Herrera-Ruiz G. Modelling temperature in intelligent buildings by means of autoregressive models. Automation in Construction 2007;16:713–22. https://doi.org/10.1016/j.autcon.2006.11.003.

[155] Mossolly M, Ghali K, Ghaddar N. Optimal control strategy for a multi-zone air conditioning system using a genetic algorithm. Energy 2009;34:58–66. https://doi.org/10.1016/j.energy.2008.10.001.

[156] Toftum J, Andersen RV, Jensen KL. Occupant performance and building energy consumption with different philosophies of determining acceptable thermal conditions. Building and Environment 2009;44:2009–16. https://doi.org/10.1016/j.buildenv.2009.02.007.

[157] Gao Y, Tumwesigye E, Cahill B, Menzel K. Using Data Mining in Optimisation of Building Energy Consumption and Thermal Comfort Management. The 2nd International Conference on Software Engineering and Data Mining, Chengdu, China: IEEE; 2010, p. 434–9.

[158] Ghahramani A, Jazizadeh F, Becerik-Gerber B. A knowledge-based approach for selecting energy-aware and comfort-driven HVAC temperature set points. Energy and Buildings 2014;85:536–48. https://doi.org/10.1016/j.enbuild.2014.09.055.

[159] Delgarm N, Sajadi B, Delgarm S. Multi-objective optimization of building energy performance and indoor thermal comfort: A new method using artificial bee colony (ABC). Energy and Buildings 2016;131:42–53. https://doi.org/10.1016/j.enbuild.2016.09.003.

[160] Li X, Wen J, Malkawi A. An operation optimization and decision framework for a building cluster with distributed energy systems. Applied Energy 2016;178:98–109. https://doi.org/10.1016/j.apenergy.2016.06.030.

[161] Shaikh PH, Nor NBM, Nallagownden P, Elamvazuthi I, Ibrahim T. Intelligent multi-objective control and management for smart energy efficient buildings. International Journal of Electrical Power & Energy Systems 2016;74:403–9. https://doi.org/10.1016/j.ijepes.2015.08.006.

[162] Rasheed M, Javaid N, Awais M, Khan Z, Qasim U, Alrajeh N, et al. Real Time Information Based Energy Management Using Customer Preferences and Dynamic Pricing in Smart Homes. Energies 2016;9:542. https://doi.org/10.3390/en9070542.

[163] Jiang L, Yao R, Liu K, McCrindle R. An Epistemic-Deontic-Axiologic (EDA) agent-based energy management system in office buildings. Applied Energy 2017;205:440–52. https://doi.org/10.1016/j.apenergy.2017.07.081.

[164] Konis K, Annavaram M. The Occupant Mobile Gateway: A participatory sensing and machine-learning approach for occupant-aware energy management. Building and Environment 2017;118:1–13. https://doi.org/10.1016/j.buildenv.2017.03.025.

[165] Manjarres D, Mera A, Perea E, Lejarazu A, Gil-Lopez S. An energy-efficient predictive control for HVAC systems applied to tertiary buildings based on regression techniques. Energy and Buildings 2017;152:409–17. https://doi.org/10.1016/j.enbuild.2017.07.056.

[166] Carreira P, Costa AA, Mansur V, Arsénio A. Can HVAC really learn from users? A simulation-based study on the effectiveness of voting for comfort and energy use optimization. Sustainable Cities and Society 2018;41:275–85. https://doi.org/10.1016/j.scs.2018.05.043.

[167] Shaikh PH, Nor NBM, Nallagownden P, Elamvazuthi I. Intelligent multi-objective optimization for building energy and comfort management. Journal of King Saud University - Engineering Sciences 2018;30:195–204. https://doi.org/10.1016/j.jksues.2016.03.001.

[168] Lou R, Hallinan KP, Huang K, Reissman T. Smart Wifi Thermostat-Enabled Thermal Comfort Control in Residences. Sustainability 2020;12:1919. https://doi.org/10.3390/su12051919.

[169] Wang Z, Hong T. Learning occupants' indoor comfort temperature through a Bayesian inference approach for office buildings in United States. Renewable and Sustainable Energy Reviews 2020;119:109593. https://doi.org/10.1016/j.rser.2019.109593.

[170] Wenqi Guo, Mengchu Zhou. Technologies toward thermal comfort-based and energy-efficient HVAC systems: A review. 2009 IEEE International Conference on Systems, Man and Cybernetics, San Antonio, TX: IEEE; 2009, p. 3883–8. https://doi.org/10.1109/ICSMC.2009.5346631.

[171] Tu JV. Advantages and disadvantages of using artificial neural networks versus logistic regression for predicting medical outcomes. Journal of Clinical Epidemiology 1996;49:1225–31. https://doi.org/10.1016/S0895-4356(96)00002-9.

[172] W. Pedrycz, M. Reformat. Rule-based modeling of nonlinear relationships. IEEE Transactions on Fuzzy Systems 1997;5:256–69. https://doi.org/10.1109/91.580800.

[173] de Reus N. Assessment of Benefits and Drawbacks of Using Fuzzy Logic, Especially in Fire Control Systems 1994:39.

[174] Wooldridge M. An introduction to multiagent systems, 2nd Edition. England: John Wiley & Sons; 2009.

[175] Zambonelli F, Omicini A. Challenges and Research Directions in Agent-Oriented Software Engineering. Autonomous Agents and Multi-Agent Systems 2004;9:253–83. https://doi.org/10.1023/B:AGNT.0000038028.66672.1e.

[176] Alfakara A, Croxford B. Using agent-based modelling to simulate occupants' behaviours in response to summer overheating. Proceedings of the Symposium on Simulation for Architecture & Urban Design, Society for Computer Simulation International; 2014, p. 13.

[177] Joy A. Pros and Cons Of Reinforcement Learning. Pythonista Planet n.d. https://www.pythonistaplanet.com/pros-and-cons-of-reinforcement-learning/ (accessed January 2, 2021).

[178] Xu M, Li S. Practical generalized predictive control with decentralized identification approach to HVAC systems. Energy Conversion and Management 2007;48:292–9. https://doi.org/10.1016/j.enconman.2006.04.012.

[179] Huang G. Model predictive control of VAV zone thermal systems concerning bi-linearity and gain nonlinearity. Control Engineering Practice 2011;19:700–10. https://doi.org/10.1016/j.conengprac.2011.03.005.

[180] Perera DWU, Pfeiffer CF. Control of temperature and energy consumption in buildings - A review. International Journal of Eenergy and Environment 2014;5:471–84.

[181] Karlsson H, Hagentoft C-E. Application of model based predictive control for water-based floor heating in low energy residential buildings. Building and Environment 2011;46:556–69. https://doi.org/10.1016/j.buildenv.2010.08.014.

[182] Yuan S, Perez R. Multiple-zone ventilation and temperature control of a single-duct VAV system using model predictive strategy. Energy and Buildings 2006;38:1248–61. https://doi.org/10.1016/j.enbuild.2006.03.007.

[183] Goldberg DE. Genetic Algorithms in Search, Optimization and Machine Learning. 1st ed. USA: Addison-Wesley Longman Publishing Co., Inc.; 1989.

[184] Eberhart R, Kennedy J. Particle swarm optimization. Proceedings of the IEEE international conference on neural networks, vol. 4, Citeseer; 1995, p. 1942–8.

[185] Shi Y, Eberhart R. A modified particle swarm optimizer. 1998 IEEE international conference on evolutionary computation proceedings. IEEE world congress on computational intelligence (Cat. No. 98TH8360), IEEE; 1998, p. 69–73.

[186] Soudan B, Saad M. An evolutionary dynamic population size PSO implementation. 2008 3rd International Conference on Information and Communication Technologies: From Theory to Applications, IEEE; 2008, p. 1–5.

[187] Parsons K. Human thermal environments: the effects of hot, moderate, and cold environments on human health, comfort, and performance. Third Edition. CRC press; 2014.

[188] Macpherson R. The assessment of the thermal environment. A review. Occupational and Environmental Medicine 1962;19:151–64.

[189] Carlucci S, Pagliano L. A review of indices for the long-term evaluation of the general thermal comfort conditions in buildings. Energy and Buildings 2012;53:194–205. https://doi.org/10.1016/j.enbuild.2012.06.015.

[190] Oliveira G, Coelho LS, Mendes N, Araujo HX. Using Fuzzy Logic in Heating Control Systems. Proceedings of the 6th ASME/JSME Thermal Engineering Joint Conference, vol. 2003, The Japan Society of Mechanical Engineers; 2003, p. 74.
[191] Benhaddou D. Living building: a building block of smart cities. Proceedings of the 2017 International Conference on Smart Digital Environment, Rabat, Morocco: 2017, p. 182–8.
[192] Nevels P. Connected Communities: A Vision for the Future of Electric Utilities. IEEE Engineering Management Review 2020;48:18–20.
[193] He L, Zhang J. Distributed Solar Energy Sharing within Connected Communities: A Coalition Game Approach. 2019 IEEE Power & Energy Society General Meeting (PESGM), IEEE; 2019, p. 1–5.
[194] American Society of Heating R and A-CE. 2017 ASHRAE handbook. 2017. https://app.knovel.com/hotlink/toc/id:kpASHRAEP2/ashrae-handbook-fundamentals/ashrae-handbook-fundamentals.
[195] de Dear R. Validation of the predicted mean vote model of thermal comfort in six Australian field studies. ASHRAE Transactions, B 1985;91:452–68.
[196] Busch J. Thermal responses to the Thai office environment. ASHRAE Transactions 1992;96.
[197] Nicol JF, Humphreys M. Understanding the adaptive approach to thermal comfort. ASHRAE Transactions 1998;104:991–1004.
[198] Nicol JF, Humphreys MA. Adaptive thermal comfort and sustainable thermal standards for buildings. Energy and Buildings 2002;34:563–72. https://doi.org/10.1016/S0378-7788(02)00006-3.
[199] Wagner A, Gossauer E, Moosmann C, Gropp T, Leonhart R. Thermal comfort and workplace occupant satisfaction—Results of field studies in German low energy office buildings. Energy and Buildings 2007;39:758–69.
[200] Farhan AA, Pattipati K, Wang B, Luh P. Predicting individual thermal comfort using machine learning algorithms. 2015 IEEE International Conference on Automation Science and Engineering (CASE), Gothenburg, Sweden: IEEE; 2015, p. 708–13. https://doi.org/10.1109/CoASE.2015.7294164.
[201] Li D, Menassa CC, Kamat VR. Personalized human comfort in indoor building environments under diverse conditioning modes. Building and Environment 2017;126:304–17. https://doi.org/10.1016/j.buildenv.2017.10.004.
[202] Merabet GH, Essaaidi M, Benhaddou D, Khalil N, Chilela J. Measuring Human Comfort for Smart Building Application: Experimental Set-Up using WSN. Proceedings of the 2nd International Conference on Smart Digital Environment, ACM; 2018, p. 56–63.
[203] Daum D, Haldi F, Morel N. A personalized measure of thermal comfort for building controls. Building and Environment 2011;46:3–11. https://doi.org/10.1016/j.buildenv.2010.06.011.
[204] Albadra D, Vellei M, Coley D, Hart J. Thermal comfort in desert refugee camps: An interdisciplinary approach. Building and Environment 2017;124:460–77.
[205] Najem N, Haddou DB, Abid MR, Darhmaoui H, Krami N, Zytoune O. Context-aware wireless sensors for IoT-centeric energy-efficient campuses. 2017 IEEE International Conference on Smart Computing (SMARTCOMP), IEEE; 2017, p. 1–6.
[206] Abid RM, Saad B. An interference-aware routing metric for Wireless Mesh Networks. International Journal of Mobile Communications 2011;9:619–41. https://doi.org/10.1504/IJMC.2011.042780.
[207] Naji N, Abid MR, Krami N, Benhaddou D. An Energy-Aware Wireless Sensor Network for Data Acquisition in Smart Energy Efficient Building. 2019 IEEE 5th World Forum on Internet of Things (WF-IoT), IEEE; 2019, p. 7–12.
[208] Hunt G, Letey G, Nightingale E. The seven properties of highly secure devices. Tech Report MSR-TR-2017-16 2017.
[209] Griffor ER, Greer C, Wollman DA, Burns MJ. Framework for cyber-physical systems: Volume 2, working group reports. 2017.
[210] Fernandes E, Jung J, Prakash A. Security analysis of emerging smart home applications. 2016 IEEE Symposium on Security and Privacy (SP), IEEE; 2016, p. 636–54.
[211] Fernandes E, Rahmati A, Eykholt K, Prakash A. Internet of things security research: A rehash of old ideas or new intellectual challenges? IEEE Security & Privacy 2017;15:79–84.
[212] Schneier B. Secrets and lies: digital security in a networked world. John Wiley & Sons; 2015.
[213] Latif S, Qayyum A, Usama M, Qadir J, Zwitter A, Shahzad M. Caveat emptor: the risks of using big data for human development. IEEE Technology and Society Magazine 2019;38:82–90.
[214] Braun T, Fung BC, Iqbal F, Shah B. Security and privacy challenges in smart cities. Sustainable Cities and Society 2018;39:499–507.